\documentclass[10pt,twocolumn,letterpaper]{article}

\usepackage[pagenumbers]{cvpr} %
\usepackage{minitoc}

\usepackage[dvipsnames]{xcolor}

\definecolor{cvprblue}{rgb}{0.21,0.49,0.74}
\usepackage[pagebackref,breaklinks,colorlinks,citecolor=cvprblue]{hyperref}
\usepackage{tikz,lipsum}
\usepackage[most]{tcolorbox}
\usepackage[capitalize]{cleveref}
\usepackage{multirow}
\usepackage{listings}
\usepackage{xcolor}
\usepackage{tabularx}
\usepackage{makecell}

\def\confName{CVPR}
\def\confYear{2025}

\title{\vspace{-0.8cm}SwiftSketch: A Diffusion Model for Image-to-Vector Sketch Generation\vspace{-0.3cm}}

\author{\hspace{-1cm} Ellie Arar$^{1}$ \hspace{-0.8cm} 
\and Yarden Frenkel$^{1}$ \hspace{-0.8cm}
\and Daniel Cohen-Or$^{1}$ \hspace{-0.8cm} 
\and Ariel Shamir$^{2}$ \hspace{-0.8cm}
\and Yael Vinker$^{1,3}$ \hspace{-1cm}
\and \hspace{0.9\linewidth} \and
$^{1}$Tel Aviv University \\ {\tt\small \{elliearar, Yf2, dcor\}@mail.tau.ac.il}\and $^{2}$Reichman University  \\ {\tt\small arik@runi.ac.il} \and $^{3}$MIT \\ {\tt\small yaelvink@mit.edu} \vspace{-0.5cm}\\ \and {\tt\small \href{https://swiftsketch.github.io/}{https://swiftsketch.github.io/} } \vspace{-0.6cm}
}

\newcommand{\methodname}{{SwiftSketch\xspace}}

\begin{document}
\doparttoc %
\faketableofcontents %

\twocolumn[{%
\vspace{-0.65cm}
\maketitle
\renewcommand\twocolumn[1][]{#1}%
\vspace{-0.8cm}
\begin{center}
    \centering
    \includegraphics[width=0.95\linewidth]{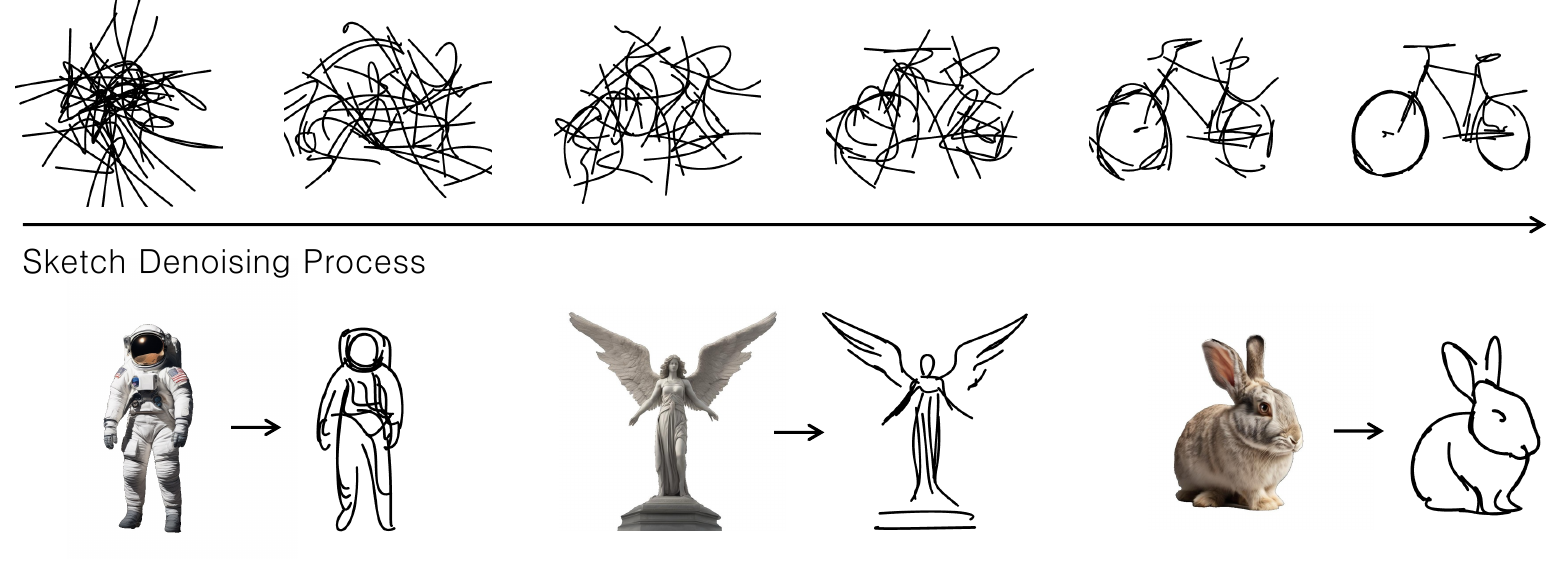}
    \captionsetup{type=figure}
    \vspace{-0.3cm}
    \caption{SwiftSketch is a diffusion model that generates vector sketches by denoising a Gaussian in stroke coordinate space (top). It generalizes effectively across diverse classes and takes under a second to produce a single high-quality sketch (bottom).}
    \label{fig:teaser}
\end{center}
}]
\begin{abstract}
\vspace{-0.4cm}
Recent advancements in large vision-language models have enabled highly expressive and diverse vector sketch generation. However, state-of-the-art methods rely on a time-consuming optimization process involving repeated feedback from a pretrained model to determine stroke placement. Consequently, despite producing impressive sketches, these methods are limited in practical applications.
In this work, we introduce \emph{SwiftSketch}, a diffusion model for image-conditioned vector sketch generation that can produce high-quality sketches in less than a second.
SwiftSketch operates by progressively denoising stroke control points sampled from a Gaussian distribution. 
Its transformer-decoder architecture is designed to effectively handle the discrete nature of vector representation and capture the inherent global dependencies between strokes.
To train SwiftSketch, we construct a \emph{synthetic} dataset of image-sketch pairs, addressing the limitations of existing sketch datasets, which are often created by non-artists and lack professional quality. For generating these synthetic sketches, we introduce ControlSketch, a method that enhances SDS-based techniques by incorporating precise spatial control through a depth-aware ControlNet.
We demonstrate that SwiftSketch generalizes across diverse concepts, efficiently producing sketches that combine high fidelity with a natural and visually appealing style.
\end{abstract}

\section{Introduction}
\begin{figure}
    \centering
    \includegraphics[width=1\linewidth]{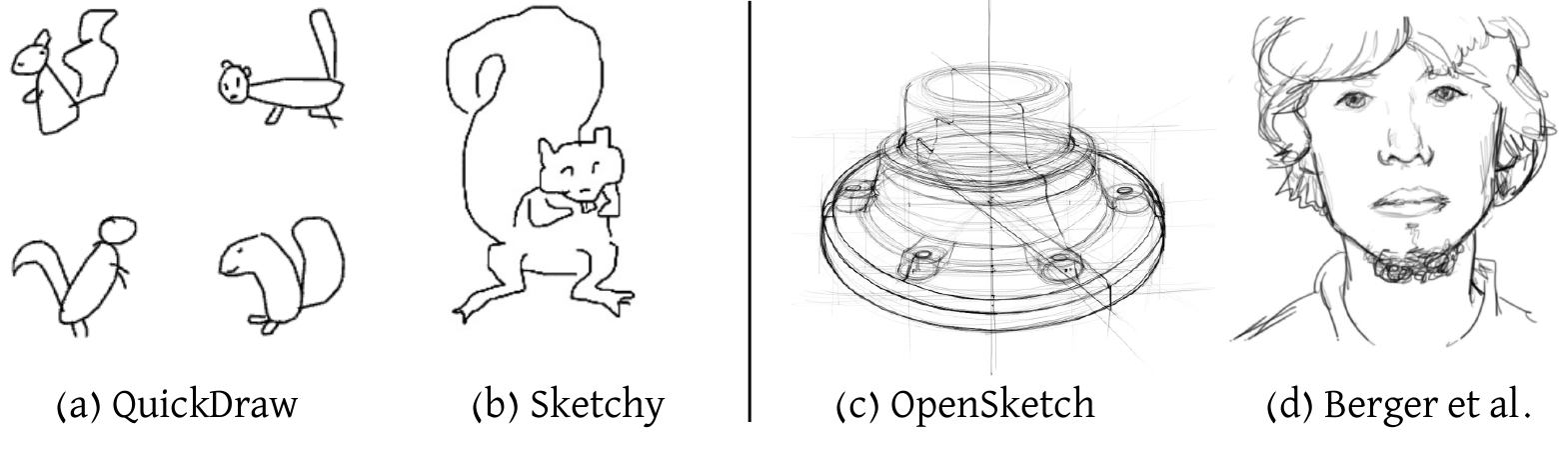}
    \vspace{-0.5cm}
    \caption{Amateur vs. Professional Sketches. (a) QuickDraw \cite{SketchRNN} and (b) Sketchy \cite{Sangkloy2016TheSD} are large-scale datasets, with Sketchy offering more fine-grained sketches, though both exhibit non-professional style. (c) OpenSketch \cite{Gryaditskaya2019OpenSketch} and (d) Berger \etal \cite{Berger2013StyleAA} contain professional sketches but are limited in scale and focus on specific domains.}
    \vspace{-0.5cm}
    \label{fig:sketch_data_example}
\end{figure}

In recent years, several works have explored the task of generating sketches from images, tackling both scene-level and object-level sketching \cite{Chan2022LearningTG, vinker2022clipasso, Vinker2022CLIPasceneSS, svgdreamer_xing_2023}. This task involves transforming an input image into a line drawing that captures its key features, such as structure, contours, and overall visual essence.
Sketches can be represented as pixels or vector graphics, with the latter often preferred for their resolution independence, enhanced editability, and ability to capture sketches' sequential and abstract nature.
Existing vector sketch generation methods often involve training a network to learn the distribution of human-drawn sketches~\cite{SurveySketchxu2020deep}.
However, collecting human-drawn sketch datasets is labor-intensive, and crowd-sourced contributors often lack artistic expertise, resulting in datasets that primarily feature amateur-style sketches (\cref{fig:sketch_data_example}, left). 
On the other hand, sketch datasets created by professional designers or artists are typically limited in scale, comprising only a few hundred samples, and are often restricted to specific domains, such as portraits or product design (\cref{fig:sketch_data_example}, right). Therefore, existing data-driven sketch generation methods are often restricted to specific domains or reflect a non-professional style present in the training data.

With recent advancements in Vision-Language Models (VLMs) \cite{zhang2024visionlanguagemodelsvisiontasks}, new approaches have emerged in the sketch domain, shifting sketch generation from reliance on human-drawn datasets to leveraging the priors of pretrained models \cite{Frans2021CLIPDrawET,vinker2022clipasso,Vinker2022CLIPasceneSS,Xing2023DiffSketcherTG}.
These methods generate professional-looking sketches by optimizing parametric curves to represent an input concept, guided by the pretrained VLM.
However, they have a significant drawback: The generation process depends on repeated feedback (backpropagation) from the pretrained model, which is inherently time-consuming -- often requiring from several minutes to over an hour to produce a single sketch.
This makes these approaches impractical for interactive applications or for tasks that require large-scale sketch data generation.

In this work, we introduce \textit{\methodname}, a diffusion-based object sketching method capable of generating high-quality vector sketches in under a second per sketch. SwiftSketch can generalize across a wide range of concepts and produce sketches with high fidelity to the input image (see Figure~\ref{fig:teaser}).

Inspired by recent advancements in diffusion models for non-pixel data \cite{tevet2023human,Luo2021DiffusionPM,Thamizharasan_2024_CVPR}, we train a diffusion model that learns to map a Gaussian distribution in the space of stroke coordinates to the data distribution (see Figure~\ref{fig:teaser}, top).
To address the discrete nature of vector graphics and the complex global topological relationships between shapes, we employ a transformer-decoder architecture with self- and cross-attention layers, trained to reconstruct ground truth sketches in both vector and pixel spaces.
The image condition is integrated into the generation process through the cross-attention mechanism, where meaningful features are first extracted from the input image using a pretrained CLIP image encoder \cite{Radfordclip}.

With the lack of available professional-quality paired vector sketch datasets, we construct a \emph{synthetic} dataset to train our network.
The input images are generated with SDXL \cite{podell2023sdxlimprovinglatentdiffusion}, and their corresponding vector sketches are produced with a novel optimization-based technique we introduce called \emph{ControlSketch}. 
ControlSketch enhances the SDS loss \cite{Poole2022DreamFusionTU}, commonly used for text-conditioned generation, by integrating a depth ControlNet \cite{controlnet2023} into the loss, enabling object sketch generation with spatial control.
Our dataset comprises over 35,000 high-quality vector sketches across 100 classes and is designed for scalability. 
We demonstrate SwiftSketch's capability to generate high-quality vector sketches of diverse concepts, balancing fidelity to input images and the abstract appearance of natural sketches.

\section{Related Work}
\paragraph{\textbf{Sketch Datasets}}
Existing sketch datasets are primarily composed of human-drawn sketches, and are designed to accomplish different sketching tasks. Class-conditioned datasets \cite{Eitz2012HowDH,SketchRNN} are particularly common, with the largest being the QuickDraw dataset \cite{SketchRNN}, containing 50 million sketches spanning 345 categories. Datasets of image-referenced sketches cover a spectrum of styles, including image trace and contours \cite{Wang2021Tracing,Li2019PhotoSketchingIC,ArbelaezBSDS500,Eitz2012HowDH}, or more abstract but still fine-grained depictions \cite{Sangkloy2016TheSD,SketchyCOCO2020}, and very abstract sketches \cite{Mukherjee2023SEVALS}. These large-scale datasets are often created by non-artists.
Efforts have been made to collect sketches from professionals \cite{Berger2013StyleAA, Gryaditskaya2019OpenSketch, Han2023AGF, Xiao2022DifferSketching}, but these datasets are often smaller in scale, and are limited to specific domains like portraits \cite{Berger2013StyleAA} or household items \cite{Gryaditskaya2019OpenSketch}. These constraints make them unsuitable for training generative models that can generalize broadly to diverse concepts.

\paragraph{\textbf{Data-Driven Sketch Generation}}
These datasets have facilitated data-driven approaches for various sketch-related tasks \cite{SurveySketchxu2020deep}. Multiple generative frameworks and architectures have been explored for vector sketch generation, including RNNs \cite{SketchRNN}, BERT \cite{Lin2020SketchBERTLS}, Transformers \cite{Bhunia2020EdinburghRE, Ribeiro2020SketchformerTR}, CNNs \cite{Kampelmhler2020SynthesizingHS,Chen2017Sketchpix2seqAM,Song2018LearningTS}, LSTMs \cite{Qi2021SketchLatticeLR, Song2018LearningTS}, GANs \cite{V2019TeachingGT}, reinforcement learning \cite{Zhou2018LearningTD,Muhammad2018LearningDS}, and diffusion models \cite{wang2023sketchknitter}. However, these methods are fundamentally designed to operate in a class-conditioned manner, restricting their ability to generate sketches to only the classes included in the training data. Additionally, they rely on crowd-sourced datasets which contain non-professional sketches, restricting their ability to handle more complex or artistic styles.
On the other hand, existing works for generating more professionally looking sketches are either restricted to specific domains \cite{Liu_2021_ICCV} or can only generate sketches in pixel space \cite{Li2019PhotoSketchingIC,Chan2022LearningTG}.
Note that image-to-sketch generating can be formulated as a style transfer task, with recent works that employ the text-to-image diffusion priors achieving highly artistic results with high fidelity \cite{Wang2024InstantStyleFL,frenkel2024implicit,hertz2023StyleAligned}, however, all of these works also operate only in pixel space.
In contrast, we focus on vector sketches due to their resolution independence, smooth and clean appearance, control over abstraction, and editable nature.

\paragraph{\textbf{VLMs for Vector Sketches}}
To reduce reliance on existing vector datasets, recent research leverages the rich priors of large pre-trained vision-language models (VLMs) in a zero-shot manner. Early methods \cite{vinker2022clipasso,Frans2021CLIPDrawET,Vinker2022CLIPasceneSS} utilize CLIP \cite{Radfordclip} as the backbone for image- and text-conditioned generation. These approaches iteratively optimize a randomly initialized set of strokes using a differentiable renderer \cite{Li2020DifferentiableVG} to bridge the gap between vector and pixel representations. More recently, text-to-image diffusion models \cite{rombach2022highresolution} have been employed as backbones, with the SDS loss \cite{Poole2022DreamFusionTU} used to guide the optimization process, achieving superior results \cite{jain2022vectorfusion,svgdreamer_xing_2023,Xing2023DiffSketcherTG}. However, the use of the SDS loss has so far been limited to text-conditioned generation.
While these approaches yield highly artistic results across diverse concepts, they are computationally expensive, relying on iterative backpropagation.

\paragraph{\textbf{Diffusion Models for Non-Pixel Data}}
Diffusion models have emerged as a powerful generative framework, extending their impact beyond traditional pixel-based data. Recent research demonstrates their versatility across diverse domains, including tasks such as human motion synthesis \cite{tevet2023human}, 3D point cloud generation \cite{Luo2021DiffusionPM, Huang2025SPAR3DSP}, and object detection reframed as a generative process \cite{Chen2022DiffusionDetDM}.
Some prior works have explored diffusion models for vector graphics synthesis. 
VecFusion \cite{Thamizharasan_2024_CVPR} uses a two-stage diffusion process for vector font generation but its architecture and vector representation are highly complex and specialized for fonts, limiting adaptability to other vector tasks. SketchKnitter \cite{wang2023sketchknitter} and Ashcroft~\etal~\cite{ashcroft2024modelling} generate vector sketches using a diffusion-based model trained on the QuickDraw and Anime-Vec10k dataset, but without conditioning on images or text inputs.

\section{Preliminaries}
\label{sec:preliminaries}
\paragraph{\textbf{Diffusion Models}}
\label{sec:diffusion}
Diffusion models \cite{ddpm2020Ho,song2021ddim} are a class of generative models that learn a distribution by gradually denoising a Gaussian. 
Diffusion models consist of a forward process $q(x_t|x_{t-1})$ that progressively noises data samples $x_0 \sim p_{data}$ at different timesteps $t\in [1,T]$, and a backward or reverse process $p(x_{t-1}|x_t)$ that progressively cleans the noised signal. 
The reverse process is the generative process and is approximate with a neural network $\epsilon_{\theta}(x_t,t)$.
During training, a noised signal at differnet timesteps is derived from a sample $x_0$ as follows:
\begin{equation}
\label{eq:xt}
x_t = \sqrt{\bar{\alpha}_t} x_0 + \sqrt{1 - \bar{\alpha}_t} \epsilon,
\end{equation}
where $\epsilon \sim \mathcal{N}(0, \mathbf{I})$, and $\bar{\alpha}_t = \prod_{s=1}^t \alpha_s$ is called the noise scheduler.
The common approach for training the model is with the following simplified objective:
\begin{equation}
\label{eq:lsimple}
L_\text{simple} = \mathbb{E}_{x_0\sim q(x_0), \epsilon \sim \mathcal{N}(0, \mathbf{I}), t \sim \mathcal{U}(1, T)} \big\| \epsilon - \epsilon_\theta(x_t, t) \big\|^2.
\end{equation}
At inference, to generate a new sample, the process starts with a Gaussian noise $x_T \sim \mathcal{N}(0, \mathbf{I})$ and the denoising network is applied iteratively for $T$ steps, yielding a final sample $x_0$.

\paragraph{\textbf{SDS Loss}}
The Score Distillation Sampling (SDS) loss \cite{Poole2022DreamFusionTU} is used to extract signals from a pretrained text-to-image diffusion model to optimize a parametric representation. For vector graphics, the parameters $\phi$ defining an SVG can be optimized using the SDS loss to represent a desired textual concept. A differentiable rasterizer \cite{Li2020DifferentiableVG} rasterize $\phi$ into a pixel image $x$, which is then noised to produce $x_t$ at a sampled timestep $t$. This noised image, conditioned on a text prompt $c$, is passed through the pretrained diffusion model, $\epsilon_\theta(x_t, t, c)$.
The deviation of the diffusion loss in \cref{eq:lsimple} is used to approximate the gradients of the initial image synthesis model's parameters, $\phi$, to better align its outputs with the conditioning prompt. Specifically, the gradient of the SDS loss is defined as:
\begin{equation}\label{eq:sds_loss}
    \nabla_\phi \mathcal{L}_{SDS} = \left[ w(t)(\epsilon_\theta(x_t,t,y) - \epsilon) \frac{\partial x}{\partial \phi} \right] ,
\end{equation}
where $w(t)$ is a constant that depends on $\alpha_t$. This optimization process iteratively adjusts the parametric model.

\section{Method}
Our method consists of three key components: (1) ControlSketch, an optimization-based technique for generating high-quality vector sketches of input objects; (2) a synthetic paired image-sketch dataset, created using ControlSketch; and (3) SwiftSketch, a diffusion model trained on our dataset for efficient sketch generation.

\subsection{ControlSketch}
\label{sec:controlsketch}
\begin{figure}
    \centering
    \includegraphics[width=1\linewidth]{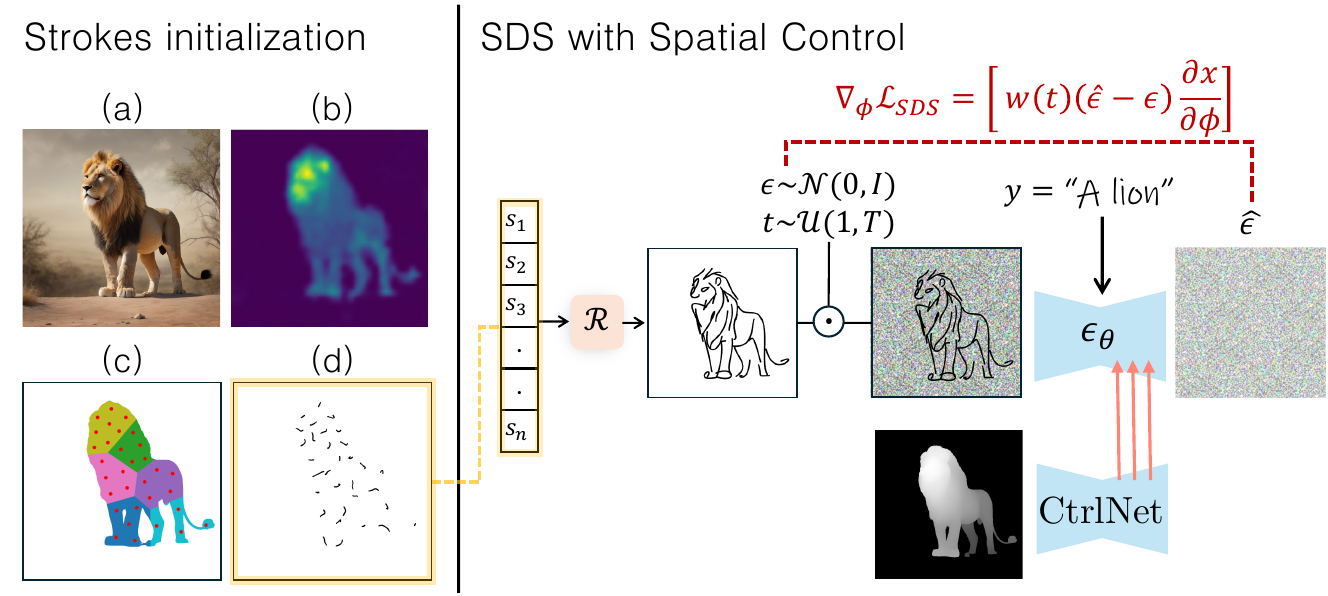}
    \caption{ControlSketch Pipeline. Left: The object area is divided into $k$ regions (c), with $n$ points distributed based on attention values from (b) while ensuring a minimum allocation per region. (d) The initial strokes are derived from these points. Right: The initial strokes are iteratively optimized to form the sketch. At each iteration, the rasterized sketch is noised based on $t$ and $\epsilon$ and fed into a diffusion model with a depth ControlNet conditioned on the image's depth and caption $y$. The predicted noise $\hat{\epsilon}$ is used for the SDS loss.}
    \label{fig:ControlSketch_pipeline}
\end{figure}

Given an input image $I$ depicting an object, our goal is to generate a corresponding sketch $S$ that maintains high fidelity to the input while preserving a natural sketch-like appearance. Following common practice in the field, we define $S$ as a set of $n$ strokes $\{s_i\}_{i=1}^n$, where each stroke is a two-dimensional cubic Bézier curve: $s_i = \{p_j^i\}_{j=1}^4 = \{(x_j, y_j)^i\}_{j=1}^4$. 
We optimize the set of strokes using the standard SDS-based optimization pipeline, as described in \Cref{sec:preliminaries}, with two key enhancements: an improved stroke initialization process and the introduction of spatial control. Our process rely on the image's attention map $I_{attn}$, depth map $I_{depth}$, and caption $y$, extracted using DDIM inversion \cite{song2021ddim}, MiDaS \cite{birkl2023midasv31model}, and BLIP2 \cite{li2023blip2bootstrappinglanguageimagepretraining} respectively.
While previous approaches \cite{vinker2022clipasso,Xing2023DiffSketcherTG} sample initial stroke locations based on the image's attention map, we observe that this method often results in missing areas in the output sketch, especially when spatial control is applied. To address this, we propose an enhanced initialization method (see \cref{fig:ControlSketch_pipeline}, left) that ensures better coverage. We divide the object area into $k=6$ equal-area regions (\cref{fig:ControlSketch_pipeline}c), using a weighted K-Means method that accounts for both attention weights and pixel locations. We distribute $\frac{n}{2}$ points equally across the regions, while the remaining $\frac{n}{2}$ points are allocated proportionally to the average attention value in each region. This means that more points are assigned to regions with higher attention. Within each region, the points are evenly spaced to further ensure good coverage.
This process determines the location of the initial set of strokes' control points to be optimized, as demonstrated in \Cref{fig:ControlSketch_pipeline}d.

The stroke optimization process is depicted in \Cref{fig:ControlSketch_pipeline}, right.
At each optimization step, the rasterized sketch $\mathcal{R}(\{s_i\}_{i=1}^n)$ is noised based on $t$ and $\epsilon$, then fed into a depth ControlNet text-to-image diffusion model \cite{controlnet2023}. The model predicts the noise $\hat{\epsilon}$ conditioned on the caption $y$ and the depth map $I_{depth}$.
We balance the weighting between the spatial and textual conditions to achieve an optimal trade-off between ``semantic'' fidelity, derived from $y$ (ensuring the sketch is recognizable), and ``geometric'' fidelity, derived from $I_{depth}$, which governs the accuracy of the spatial structure.

\begin{figure}
    \centering
    \includegraphics[width=0.9\linewidth]{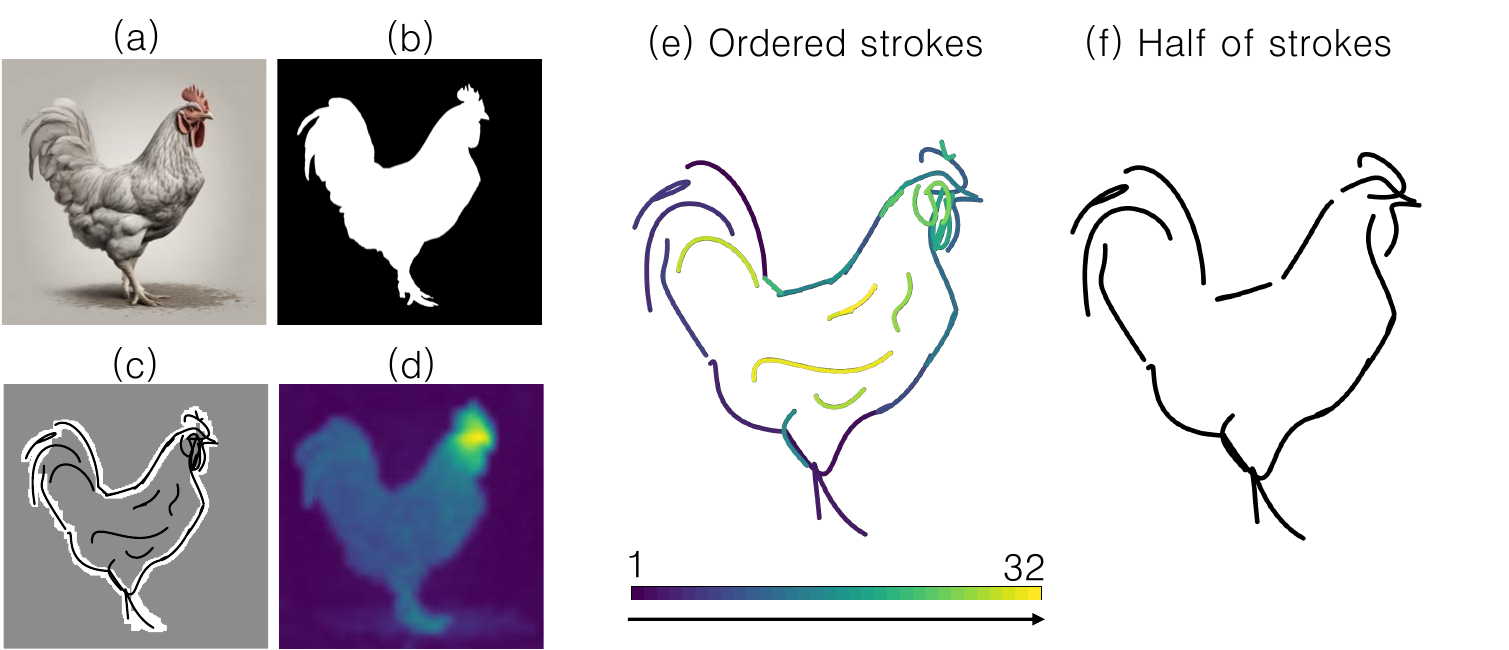}
    \caption{(a) Input image. (b) Object mask. (c) The object's contour is extracted from the mask using morphological operations, and sketch pixels that intersect with the contour are given higher weight. (d) Attention map. (e) We sort the strokes based on a combination of contour intersection count and attention score. (f) A visualization of the first 16 strokes in the ordered sketch, demonstrating the effectiveness of our sorting scheme.}
    \label{fig:sketchsort}
\end{figure}

\begin{figure*}
    \centering
    \includegraphics[width=1\linewidth]{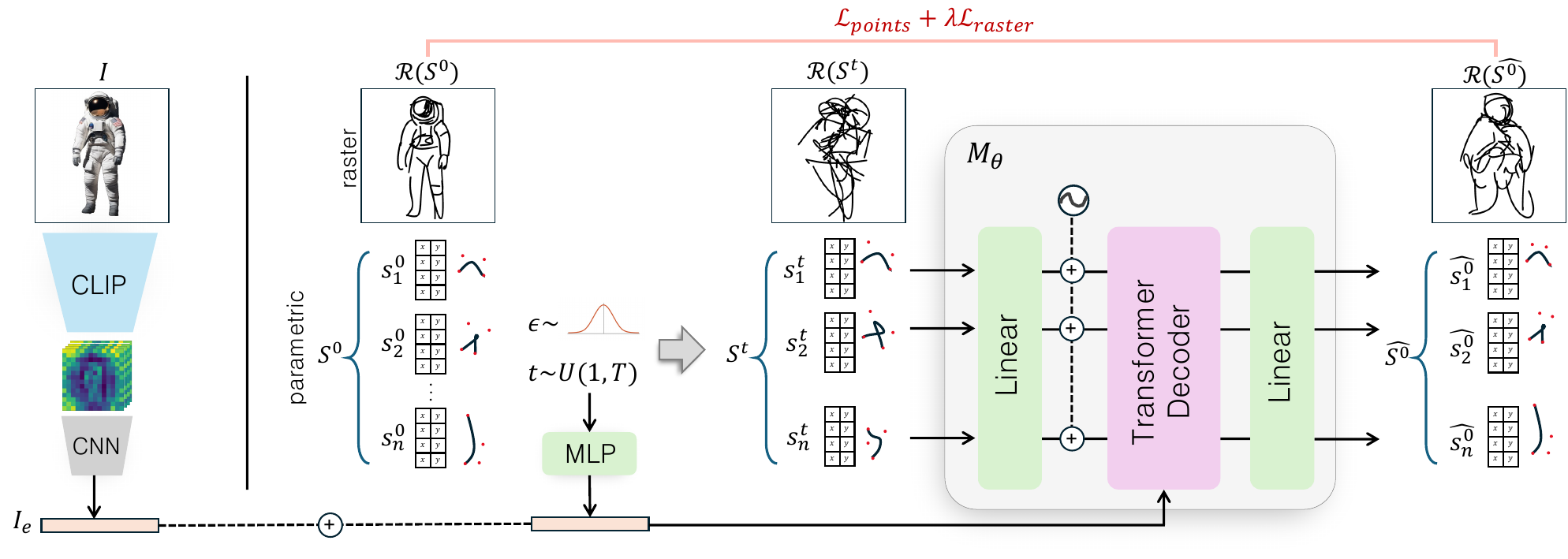}
    \caption{SwiftSketch Training Pipeline. At each training iteration, an image $I$ is passed through a frozen CLIP image encoder, followed by a lightweight CNN, to produce the image embedding $I_e$. The corresponding vector sketch $S^0$ is noised based on the sampled timestep $t$ and noise $\epsilon$, forming $S^t$ (with $\mathcal{R}(S^t)$ illustrating the rasterized noised sketch, which is not used in training). The network $M_\theta$, a transformer decoder, receives the noised signal  $S^t$ and is tasked with predicting the clean signal $\hat{S^0}$, conditioned on the image embedding $I_e$ and the timestep $t$ (fed through the cross-attention mechanism). The network is trained with two loss functions: one based on the distance between the control points and the other on the similarity of the rasterized sketches.}\vspace{-0.2cm}
    \label{fig:diffusion_pipe}
\end{figure*}

\subsection{The ControlSketch Dataset}
\label{sec:data}
We utilize ControlSketch to generate a paired image-vector sketch dataset. Each data sample comprises the set $\{I$, $I_{attn}$, $I_{depth}$, $I_{mask}$, $S$, $c$, $y\}$, which includes, respectively, the image, its attention map, depth map, and object mask, along with the corresponding vector sketch of the object, class label, and caption. 
To generate the images, we utilize SDXL \cite{podell2023sdxlimprovinglatentdiffusion}, along with a prompt template designed to produce images for each desired class $c$ (an example of a generated image for the class ``lion'' is shown in \Cref{fig:ControlSketch_pipeline}a).
We then apply ControlSketch on the masked images to generate the corresponding vector sketches. Additional details are provided in the supplementary.
Optimization-based methods, such as ControlSketch, do not impose an inherent stroke ordering. Learning an internal stroke order enables the generation of sketches with varying levels of abstraction by controlling the number of strokes generated.
Thus, we propose a heuristic stroke-sorting scheme that prioritizes contour strokes and those depicting salient regions (illustrated in \Cref{fig:sketchsort}). Consequently, each vector sketch $S$ is represented as an ordered sequence of strokes $(s_1, \dots, s_n)$.

\begin{figure}
    \centering
    \includegraphics[width=1\linewidth]{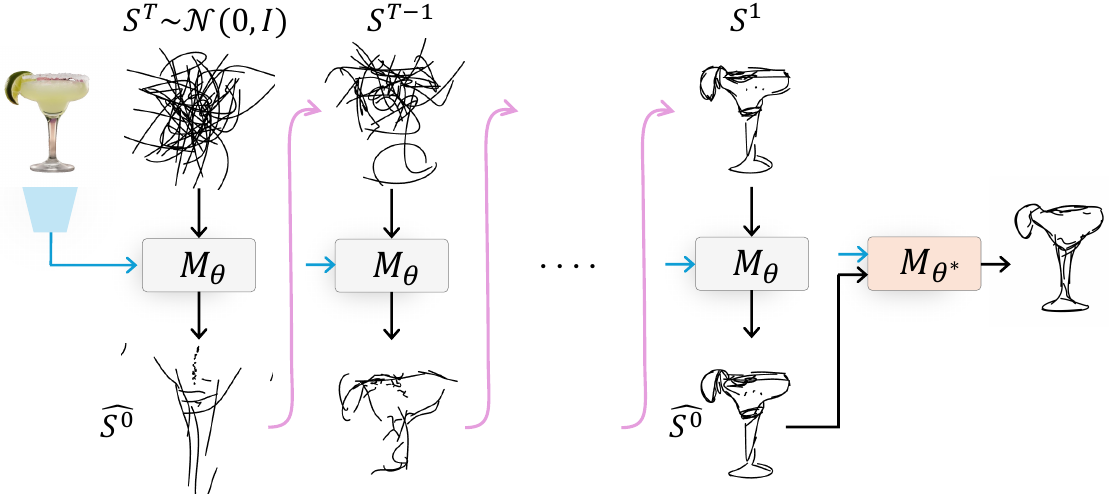}
    \vspace{-0.4cm}
    \caption{Inference Process. Starting with randomly sampled Gaussian noise $S^T \sim \mathcal{N}(0, \mathbf{I})$, the model $M_{\theta}$ predicts the clean sketch $\hat{S}^0 = M_{\theta}(S^t, t, I_e)$ at each step $t$, which is then re-noised to $S^{t-1}$. This iterative process is repeated for $T$ steps and is followed by a final feed-forward pass through a refinement network, $M_{\theta^*}$, which is a trainable copy of $M_{\theta}$, specifically trained to correct very small residual noise.}
    \label{fig:Inference}
\end{figure}

\begin{figure*}[h!]
    \centering
    \setlength{\tabcolsep}{0pt}
    {\small
    \begin{tabular}{c c | c c}

        \raisebox{0.1\height}{\includegraphics[height=0.055\linewidth]{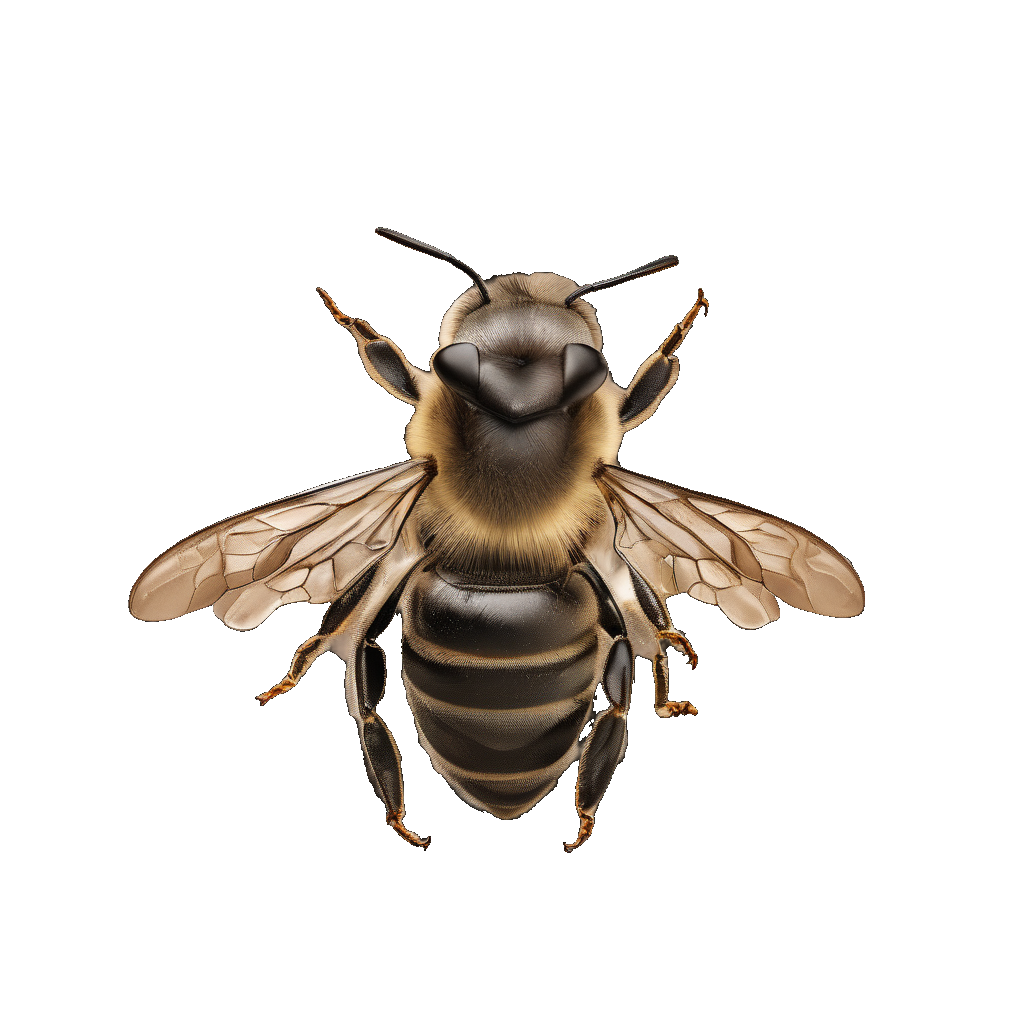}} &
        \includegraphics[height=0.065\linewidth]{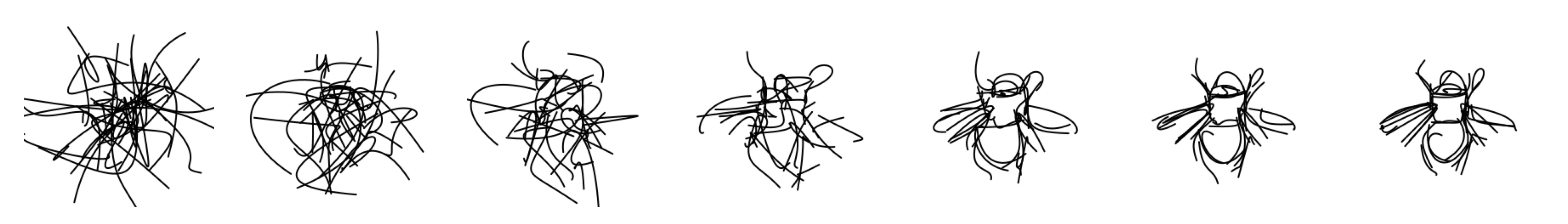} &
        \raisebox{0.1\height}{\includegraphics[width=0.055\linewidth]{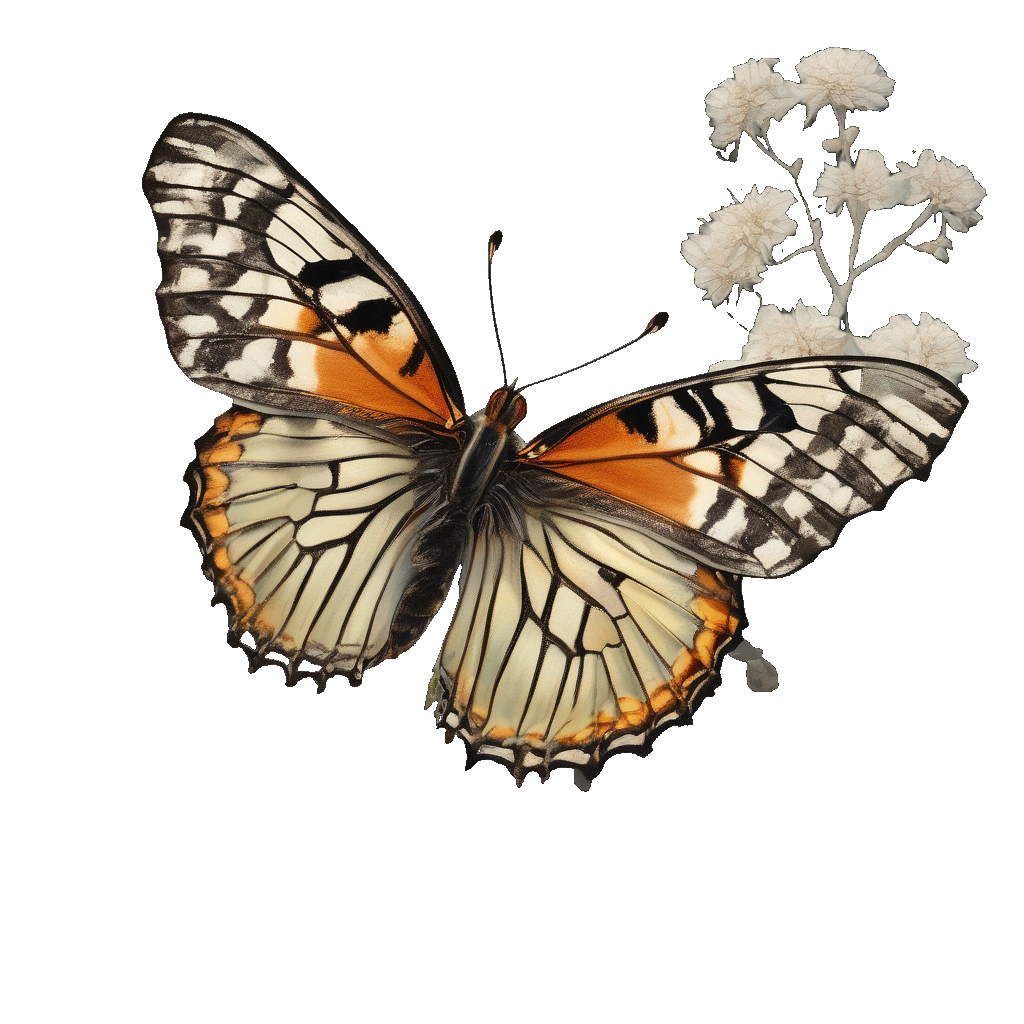}} &
        \includegraphics[height=0.065\linewidth]{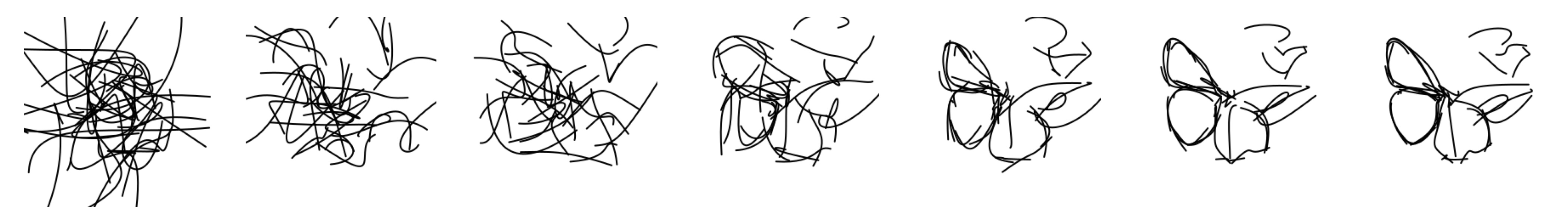} \\
    
        \raisebox{0.12\height}{\includegraphics[height=0.055\linewidth]{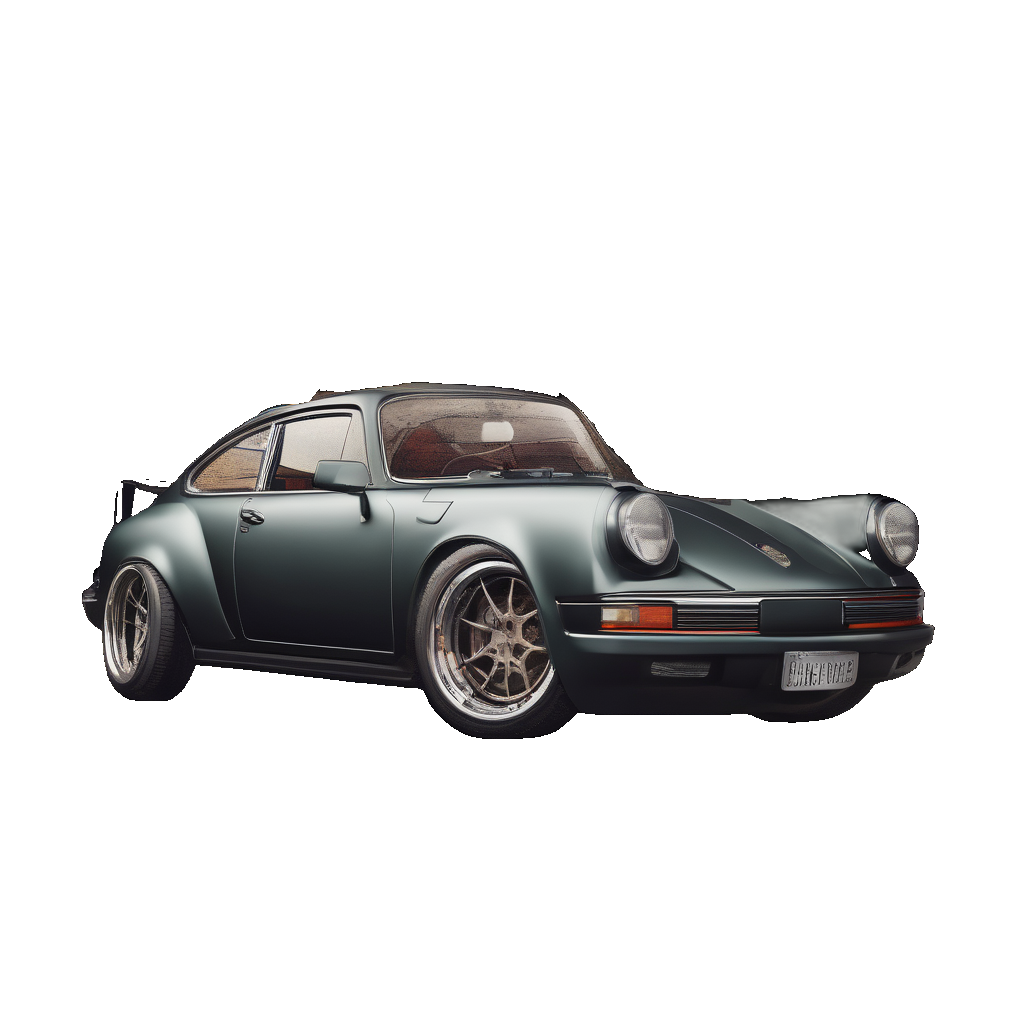}} &
        \includegraphics[height=0.065\linewidth]{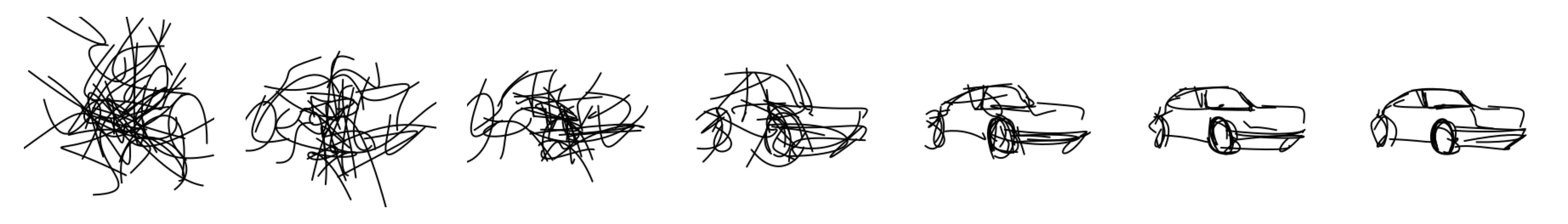} &    
        \raisebox{0.1\height}{\includegraphics[width=0.06\linewidth]{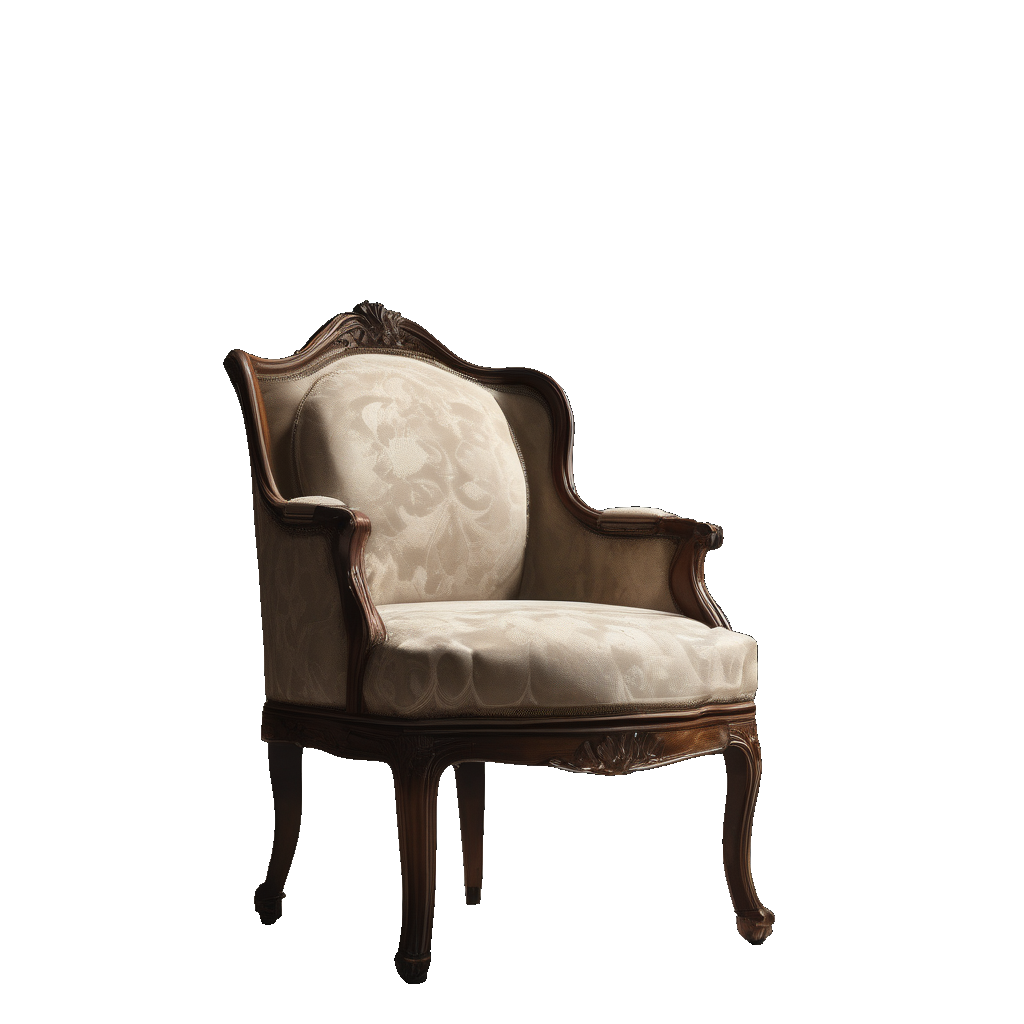}} &
        \includegraphics[height=0.065\linewidth]{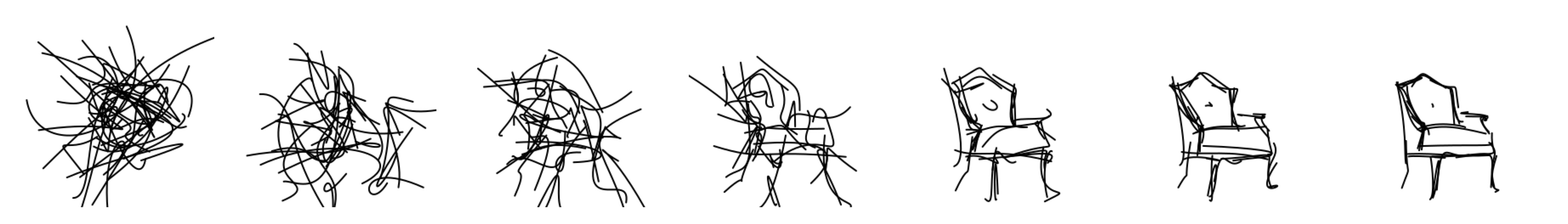} \\
    
        \raisebox{0.12\height}{\includegraphics[height=0.055\linewidth]{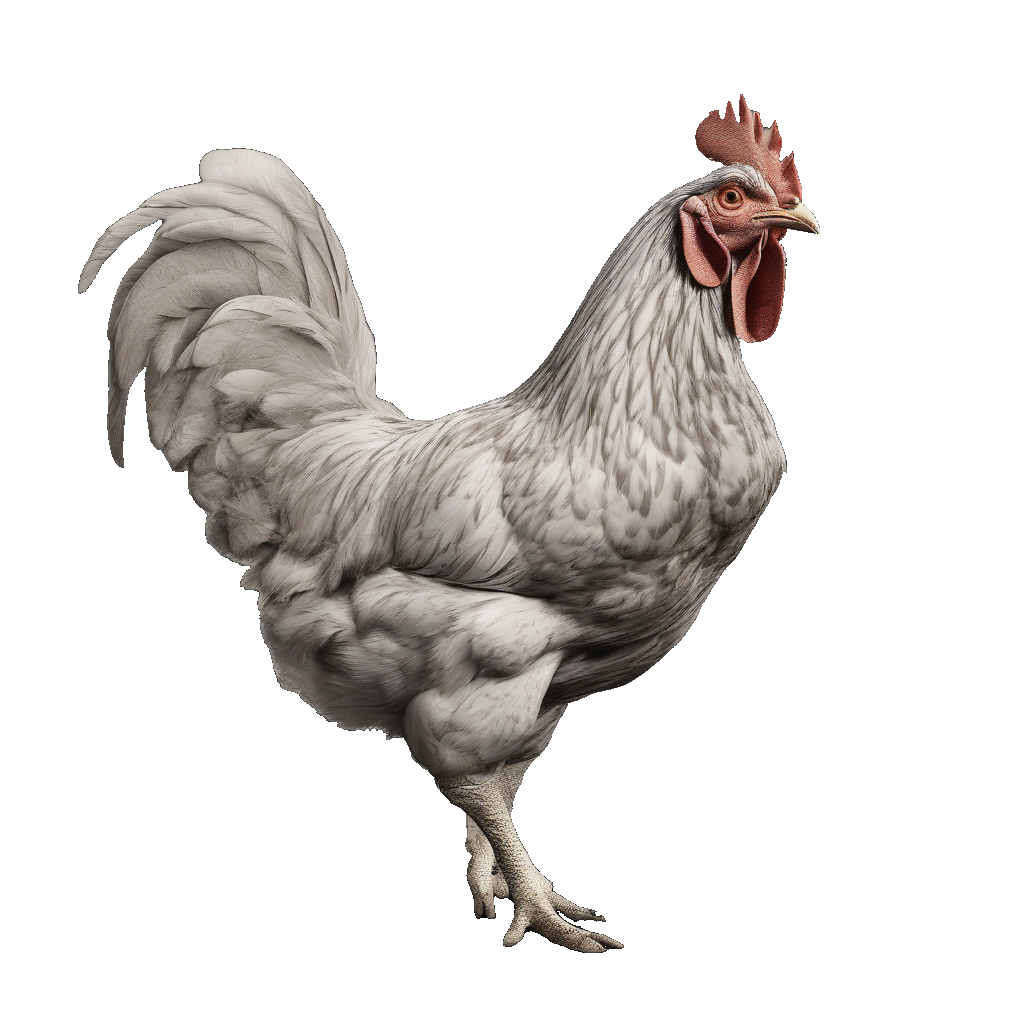}} &
        \includegraphics[height=0.065\linewidth]{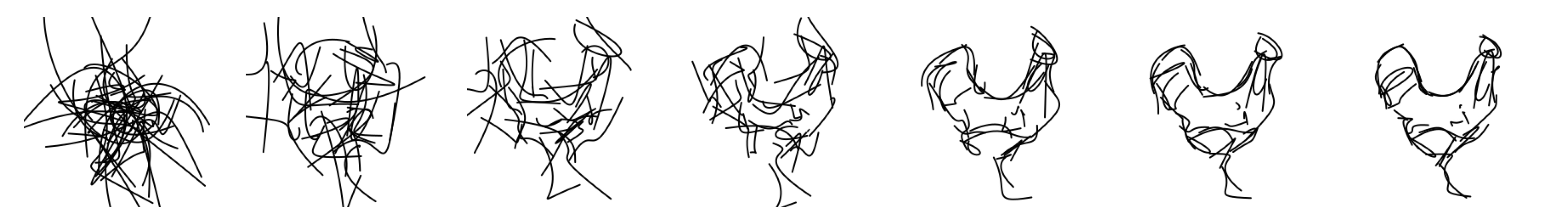} &
         \raisebox{0.15\height}{\includegraphics[height=0.05\linewidth]{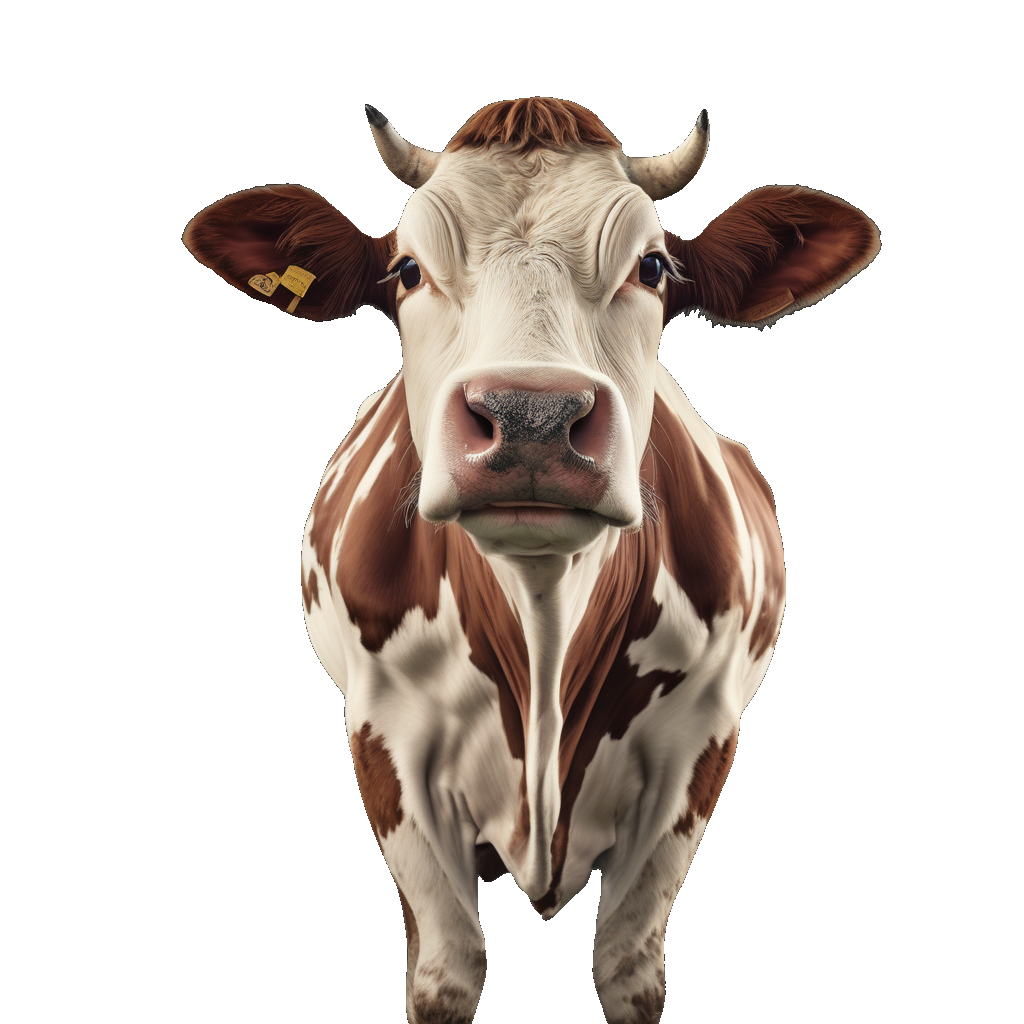}}&
        \includegraphics[height=0.065\linewidth]{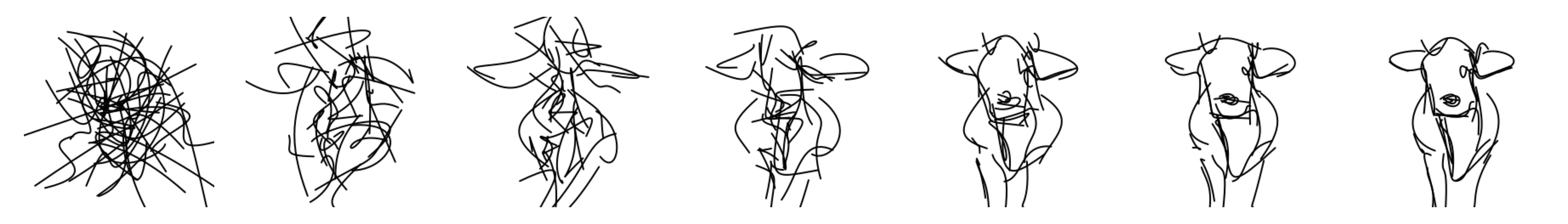} \\
    
        \raisebox{0.1\height}{\includegraphics[height=0.05\linewidth]{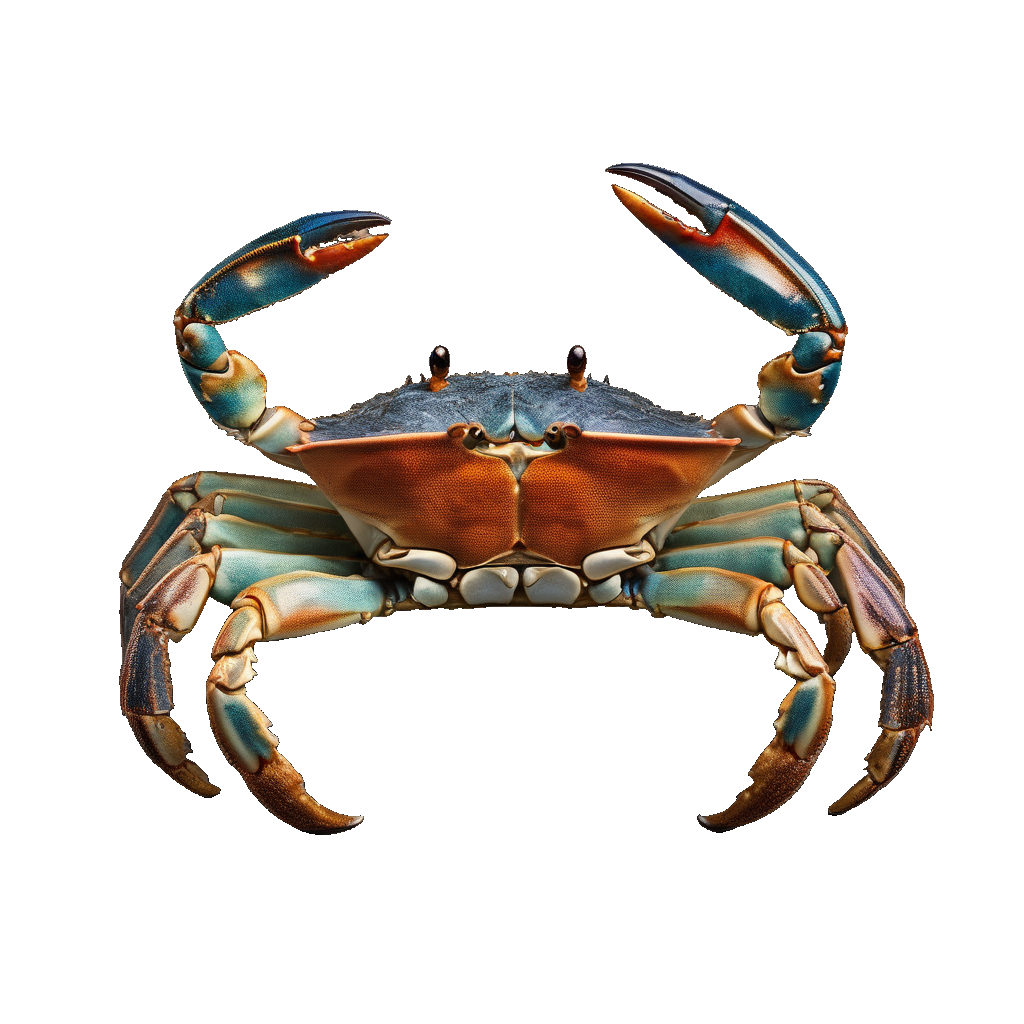}} &
        \includegraphics[height=0.065\linewidth]{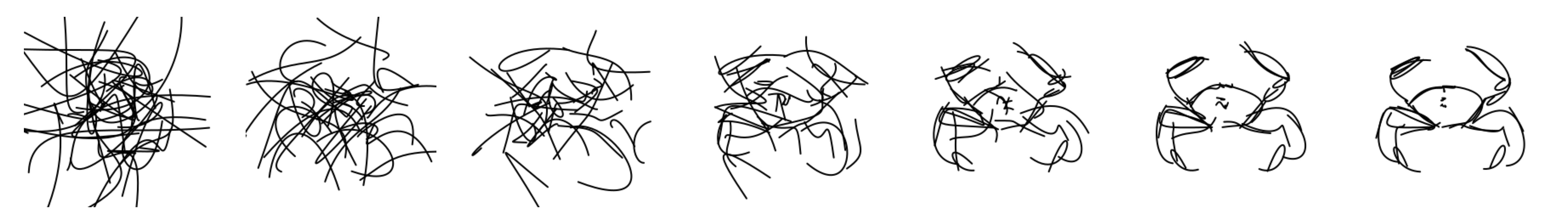} &
   
        \raisebox{0.1\height}{\includegraphics[height=0.055\linewidth]{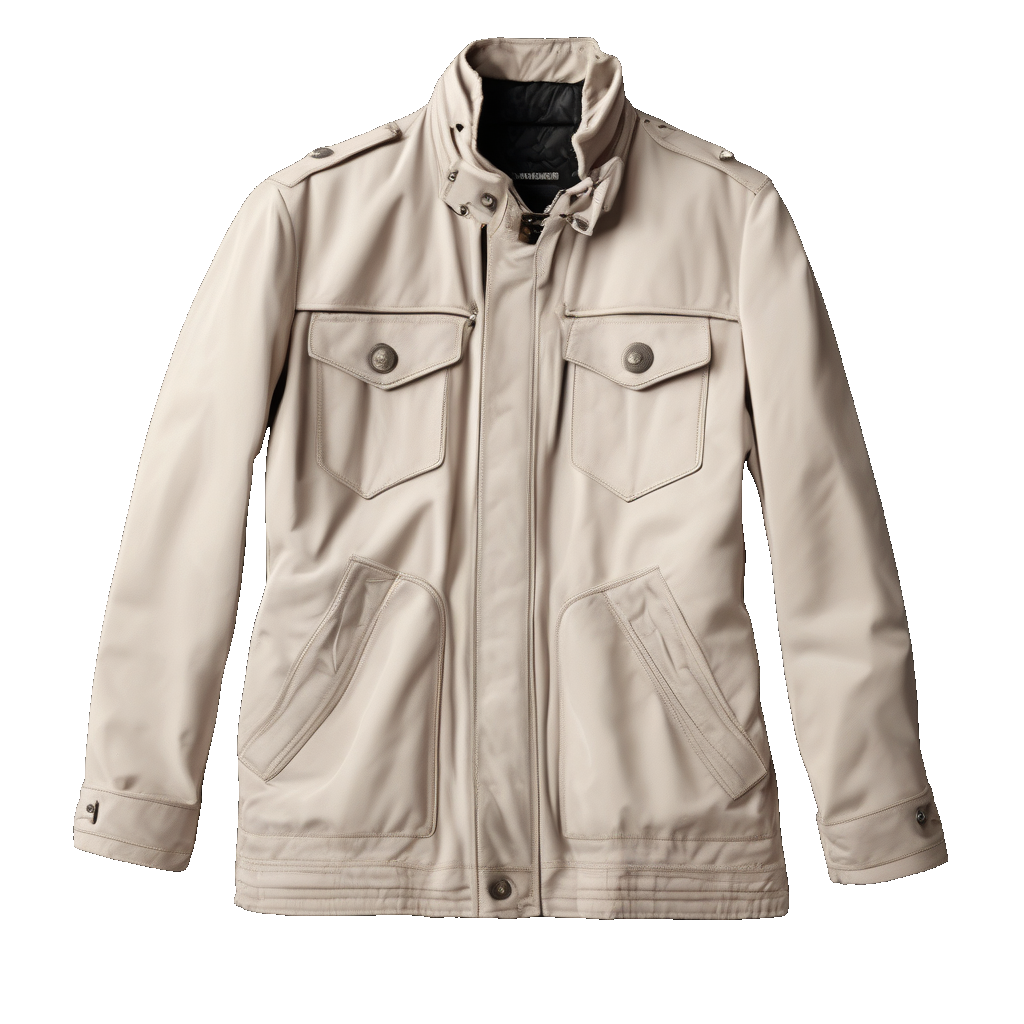}} &
        \includegraphics[height=0.065\linewidth]{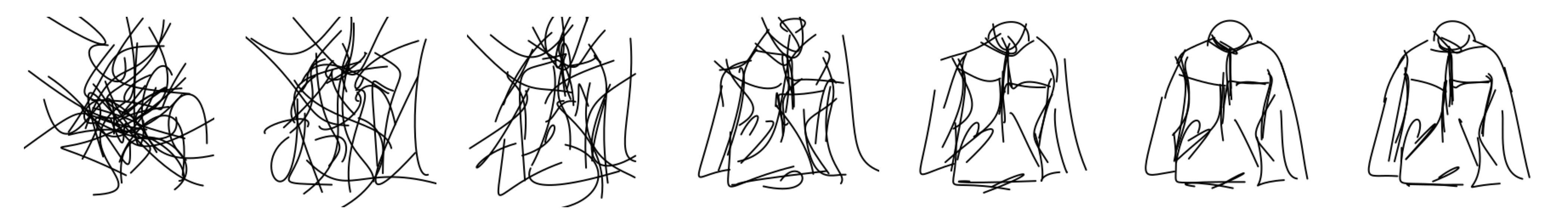} \\

    \end{tabular}
    }
    \vspace{-0.4cm}
    \caption{Examples of the denoising process. From left to right: strokes' control points are sampled from a Gaussian distribution, and our network progressively refines the signal to generate a sketch.}
    \label{fig:denoising}
\end{figure*}

\subsection{SwiftSketch}
\label{sec:swiftsketch}
We utilize the ControlSketch dataset to train a generative model $M_{\theta}$ that learns to efficiently produce a vector sketch from an input image $I$.
We define $M_{\theta}$ as a transformer decoder to account for the discrete and long-range dependencies inherent in vector sketches.
The training of $M_{\theta}$ follows the standard conditional diffusion framework, as outlined in \Cref{sec:diffusion}, with task-specific modifications to address the characteristics of vector data and the image-to-sketch task. In our case, the model learns to denoise the set of $(x,y)$ coordinates that define the strokes in the sketch. 

The training process is depicted in \Cref{fig:diffusion_pipe}. 
At each iteration, a pair $(I, S^0)$ is sampled from the dataset, where $S^0\in R^{2\times 4 \times n}$ is the clean sketch in vector representation, and $\mathcal{R}(S^0)$ denotes the corresponding rasterized sketch in pixel space, with $\mathcal{R}$ being a differentiable rasterizer \cite{Li2020DifferentiableVG}. 
The image $I$ is processed using a pretrained CLIP ResNet model \cite{Radfordclip}, where features are extracted from its fourth layer, recognized for effectively capturing both geometric and semantic information \cite{vinker2022clipasso}. These features are then refined through a lightweight CNN to enhance learning and align dimensions for compatibility with $M_\theta$. This process yields the image embedding $I_e$.
At each iteration, we sample a timestep $t \sim \mathcal{U}(1,T)$ and noise $\epsilon \sim \mathcal{N}(0, \mathbf{I})$ to define $S^t$:
\begin{equation}
S^t = \sqrt{\bar{\alpha}_t} S^0 + \sqrt{1 - \bar{\alpha}_t} \epsilon,
\label{eq:St}
\end{equation} 

where $\bar{\alpha}_t$ is the noise scheduler as a function of $t$. As illustrated in \Cref{fig:diffusion_pipe}, $S^t$ represents a noised version of $S^0$ in vector space, with the level of noise determined by the timestep $t$. 
The control points $\{s_1^t,\dots s_n^t\}$ are fed into the network $M_{\theta}$, where they are first encoded via a linear layer (depicted in green), and combined with a standard positional embedding before being passed through the transformer decoder (in pink), which consists of 8 layers of cross-attention and self-attention. The encoded timestep $t$ and image features $I_e$ are fed into the transformer through the cross-attention mechanism.
The decoder output is projected back to the original points dimension through a linear layer, yielding the prediction $M_{\theta}(S_t,t,I_e) = \hat{S^0}$. 

We train $M_{\theta}$ with two training objectives $\mathcal{L}_{\text{points}}$ and $\mathcal{L}_{\text{raster}}$, applied on both the vector and raster representation of the sketch:
\vspace{-0.1cm}
\begin{equation}
\label{eq:1}
\begin{aligned}
\mathcal{L}_{\text{points}} = \| S^0 - \hat{S}^0 \|_1 = \sum_{i=1}^{n} \| s_i^0-\hat{s_i^0}\|_1,  \\ 
\mathcal{L}_{\text{raster}} = LPIPS \left( \mathcal{R}(S^0), \mathcal{R}(\hat{S}^0)\right), 
\end{aligned}
\end{equation}

where $\mathcal{L}_{\text{points}}$ is defined by the $L_1$ distance between the sorted control points of the ground truth sketch $S^0$ and the predicted sketch $\hat{S}^0$, and $\mathcal{L}_{\text{raster}}$ is the LPIPS distance \cite{zhang2018perceptual} between the rasterized sketches.
$\mathcal{L}_{\text{points}}$ encourages per-stroke precision, while $\mathcal{L}_{\text{raster}}$ encourages the generated sketch to align well with the overall structure of the ground truth sketch. 
Together, our training loss is: $\mathcal{L} = \mathcal{L}_{\text{points}} + \lambda \mathcal{L}_{\text{raster}}$, with $\lambda=0.2$.

As is often common, to apply classifier-free guidance \cite{Ho2022ClassifierFreeDG} at inference, we train $M_{\theta}$ to learn both the conditioned and the unconditioned distributions by randomly setting \( I = \emptyset \) for 10\% of the training steps.

The inference process is illustrated in \Cref{fig:Inference}. The model, $M_{\theta}$, generates a new sketch by progressively denoising randomly sampled Gaussian noise, $S^T \sim \mathcal{N}(0, \mathbf{I})$. At each step $t$, $M_{\theta}$ predicts the clean sketch $\hat{S^0} = M_{\theta}(S^t, t, I_e)$, conditioned on the image embedding $I_e$ and time step $t$. The next intermediate sketch, $S^{t-1}$, is derived from $\hat{S^0}$ using \Cref{eq:St}. This process is repeated for $T$ steps.
We observe that the final output sketches from the denoising process may retain slight noise. This is likely because the network prioritizes learning to clean heavily noised signals during training, while small inaccuracies in control point locations have a smaller impact on the loss function, leading to reduced precision at finer timesteps.
To address this, we introduce a refinement stage, where a learned copy of our network, $M_{\theta^*}$, is fine-tuned to perform an additional cleaning step. This refinement network is trained in a manner similar to the original model, with the objective of denoising a slightly noised sketch, conditioned on the same input image features, while the timestep condition is fixed at 0. More details are provided in the supplementary. This refinement stage is inspired by similar strategies employed in the pixel domain \cite{podell2023sdxlimprovinglatentdiffusion,saharia2022photorealistic}, where additional processing steps are used to improve the quality and resolution of generated images. As illustrated in \Cref{fig:Inference}, after the final denoising step of $M_\theta$ is applied, $\hat{S^0}$ is passed through $M_{\theta^*}$ to perform the additional refinement.

\begin{figure}
    \centering
    \setlength{\tabcolsep}{0pt}
    {\small
    \begin{tabular}{c c c c c c }
    Input & 12s & 17s & 22s& 27s & 32s\\

         \includegraphics[width=0.17\linewidth]{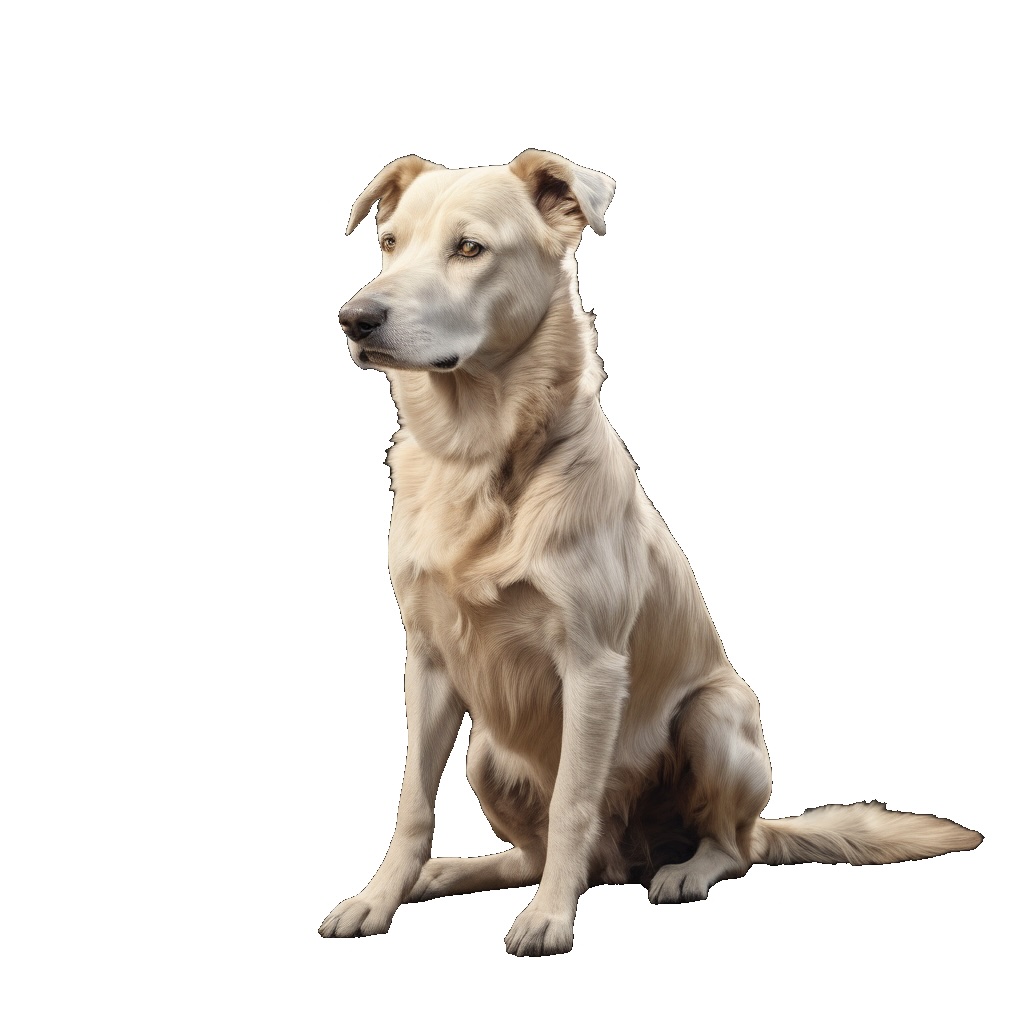} &
        \includegraphics[width=0.17\linewidth]{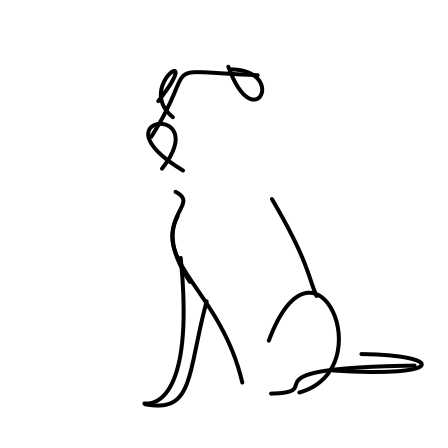} &
        \includegraphics[width=0.17\linewidth]{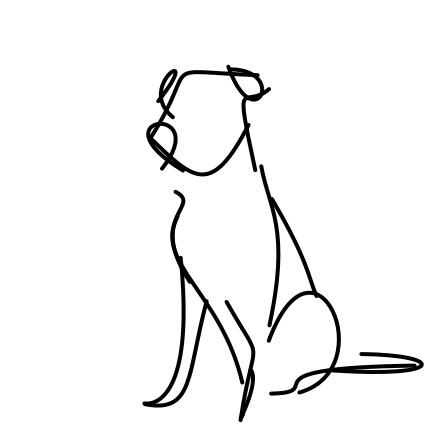} &
        \includegraphics[width=0.17\linewidth]{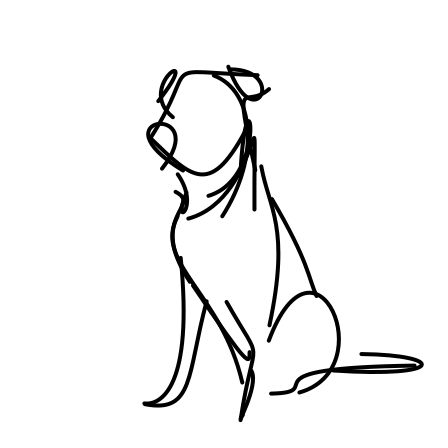} &
        \includegraphics[width=0.17\linewidth]{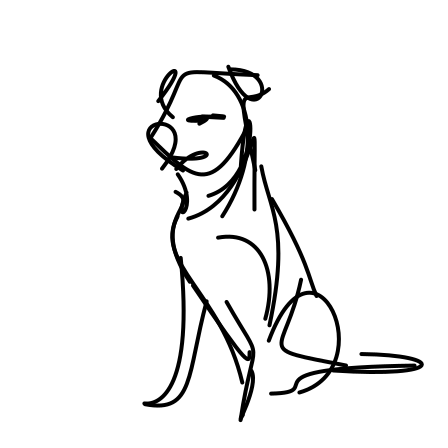} &
        \includegraphics[width=0.17\linewidth]{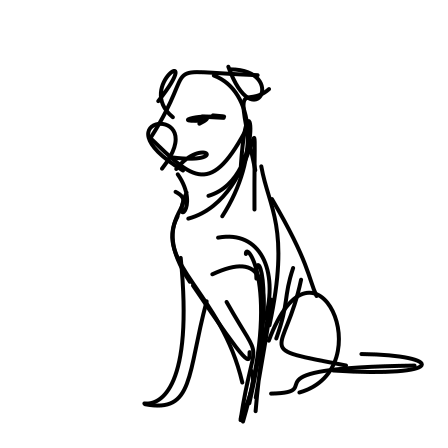} \\

        \includegraphics[width=0.17\linewidth]{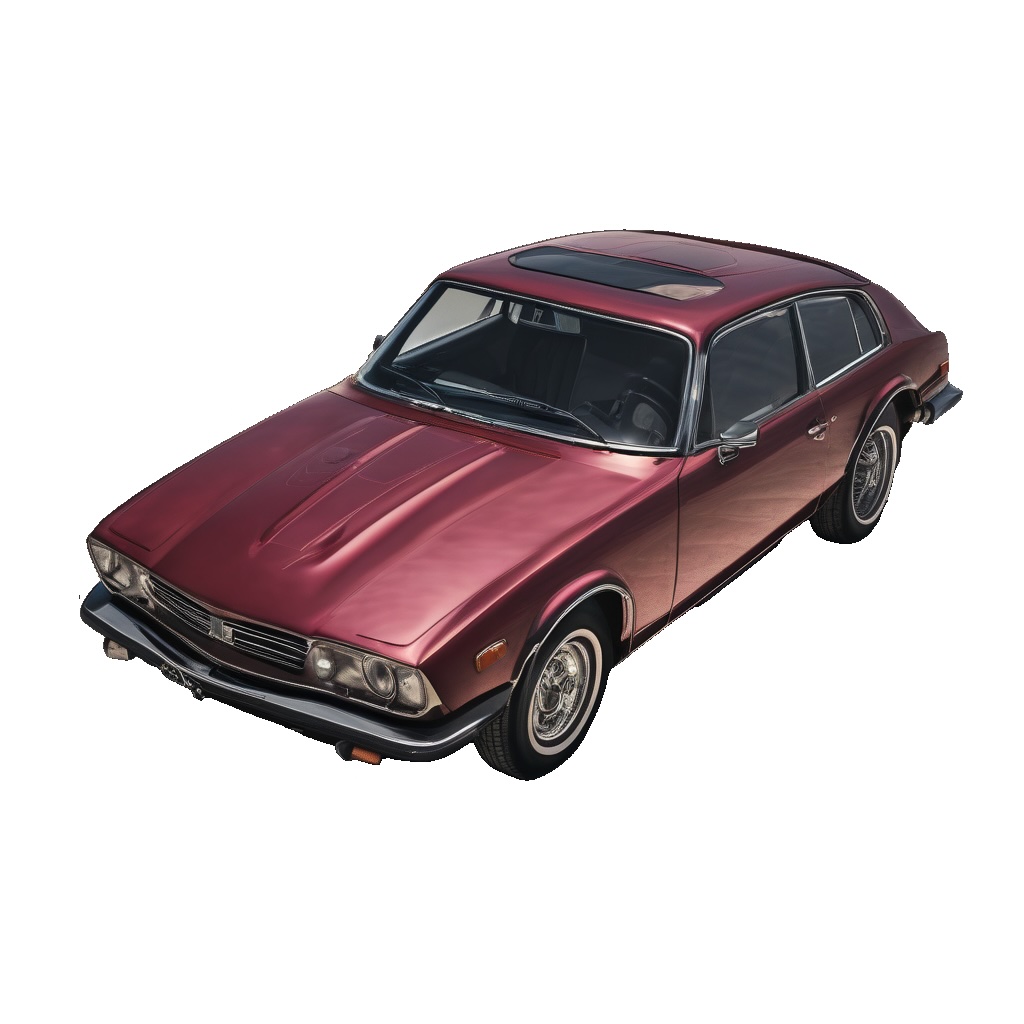} &
        \includegraphics[width=0.17\linewidth]{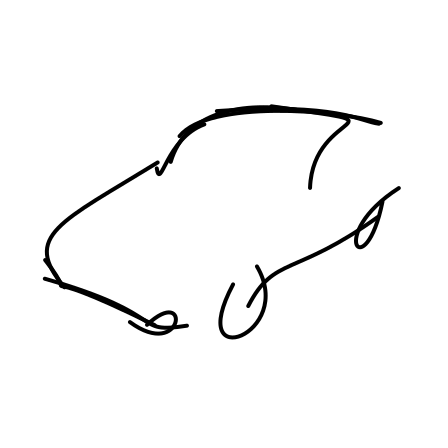} &
        \includegraphics[width=0.17\linewidth]{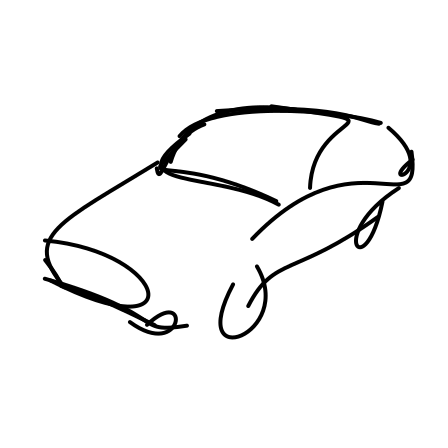} &
        \includegraphics[width=0.17\linewidth]{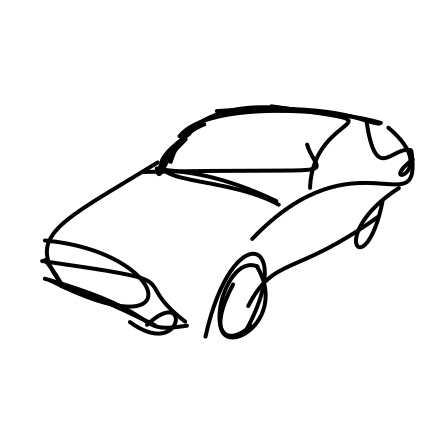} &
        \includegraphics[width=0.17\linewidth]{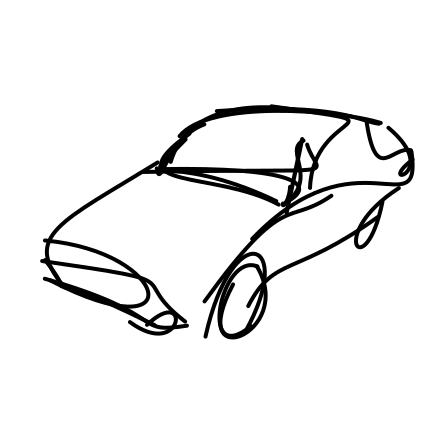} &
        \includegraphics[width=0.17\linewidth]{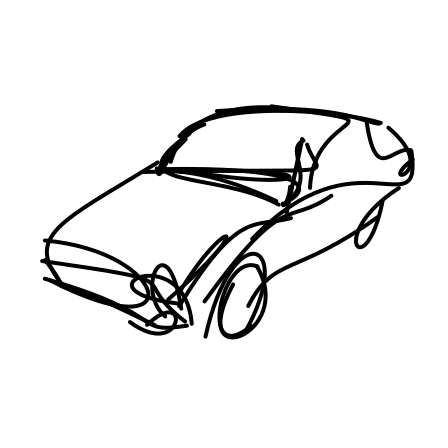}\\

    \end{tabular}
    }
    \caption{Stroke Order Visualization. Generated sketches are visualized progressively, with the stroke count shown on top. Early strokes capture the object's contour and key features, while later strokes add finer details.}
    \label{fig:order}
\end{figure}

\subsection{Implementation Details}
ControlSketch requires approximately 2,000 steps to converge, taking around $10$ minutes on a standard RTX3090 GPU.
SwiftSketch is trained with $T = 50$ noising steps, to support fast generation. 
To encourage the model to focus on fine details, we adjust the noise scheduler to perturb the signal more subtly for small timesteps compared to the cosine noise schedule proposed in \cite{Nichol2021ImprovedDD}.
The model is trained on images from 15 classes, with 1,000 samples per class. The training process spans 400K steps, requiring approximately six days on a single A100 GPU. At inference, we use a guidance scale of 2.5. Our synthetic dataset includes an additional 200 test samples for the 15 training classes, as well as 85 additional object categories, each with 200 samples. Additional implementation details, as well as detailed class labels and dataset visualizations are provided in the supplementary material.

\begin{figure*}[t]
    \centering
    \setlength{\tabcolsep}{2pt}
    {\small
    \begin{tabular}{c |c c c c c |c c}
         \midrule
         & XDoG &  Chan et al. & Instant-Style  & Photo-Sketching & CLIPasso & ControlSketch  & SwiftSketch \\
        \midrule
        Time $|$ \textcolor{orange}{\textbf{P}} / \textcolor{cyan}{\textbf{V}} &$\approx$ 0.1 sec. $|$ \textcolor{orange}{\textbf{P}} & $\approx$ 0.04 sec. $|$ \textcolor{orange}{\textbf{P}} & $\approx$ 1 min.$|$ \textcolor{orange}{\textbf{P}} & $\approx 0.6 $ sec.$|$ \textcolor{orange}{\textbf{P}} & $\approx 5$ min. $|$ \textcolor{cyan}{\textbf{V}} & $\approx 10$ min. $|$ \textcolor{cyan}{\textbf{V}} & $\approx 0.5 $ sec.$|$ \textcolor{cyan}{\textbf{V}} \\
        \midrule
        \includegraphics[trim=4cm 1.8cm 4cm 2.5cm,clip,width=0.11\textwidth]{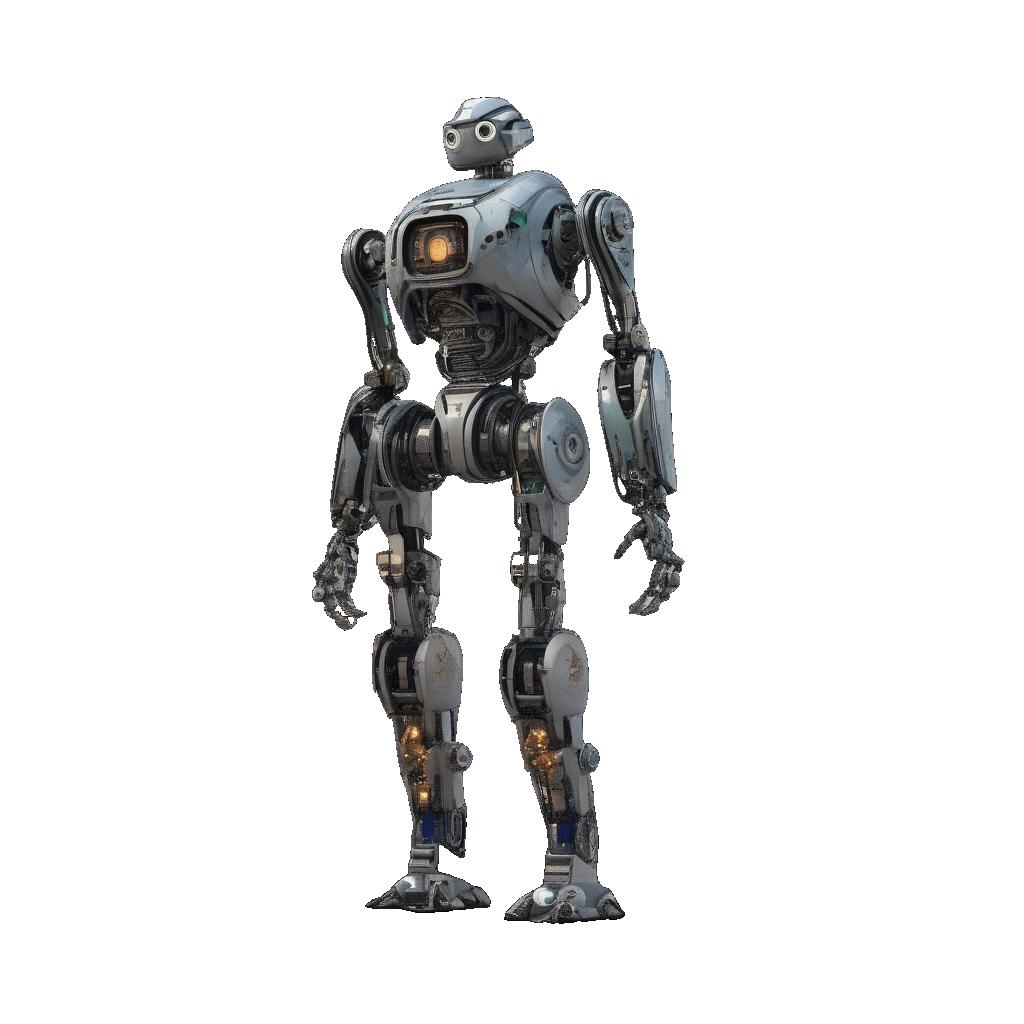} &
        \includegraphics[trim=4cm 1.8cm 4cm 2.5cm,clip,width=0.11\textwidth]{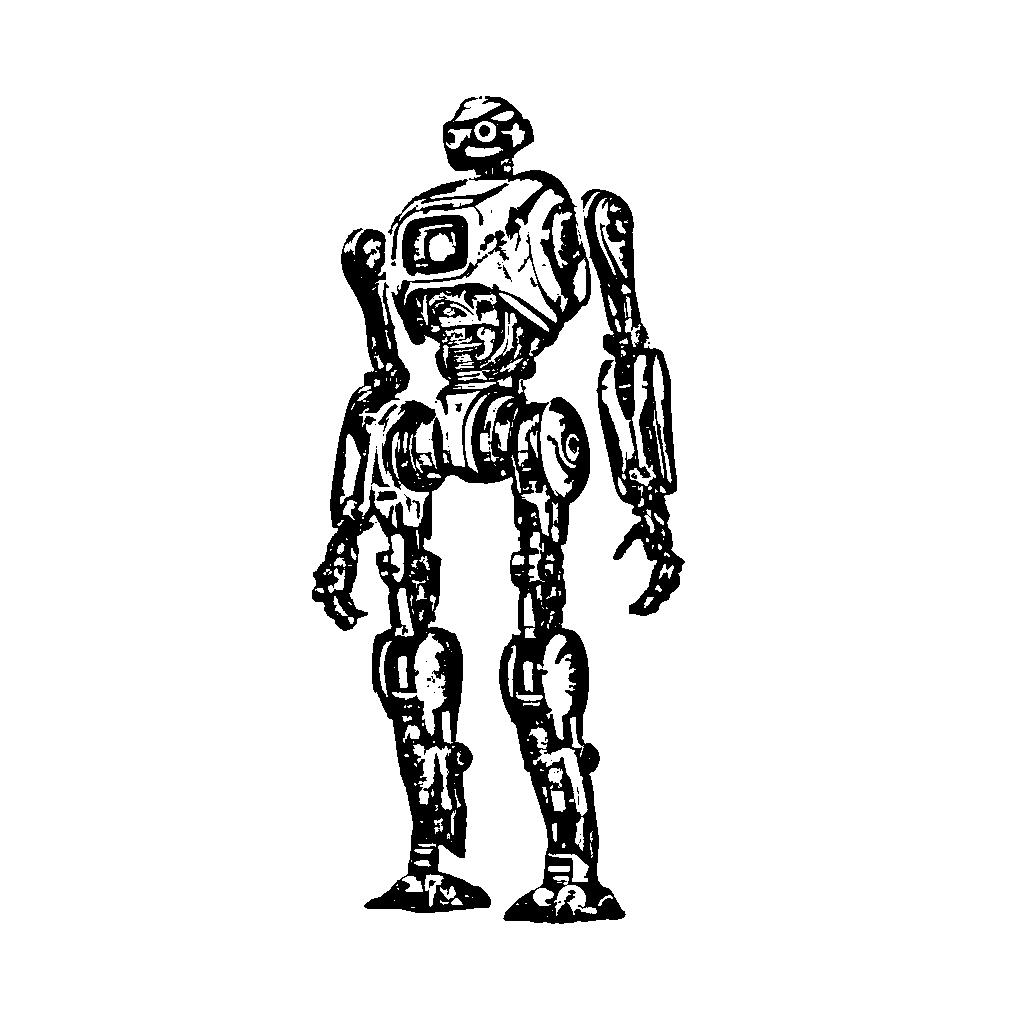} &
        \includegraphics[trim=1cm 0.45cm 1cm 0.6cm,clip,width=0.11\textwidth]{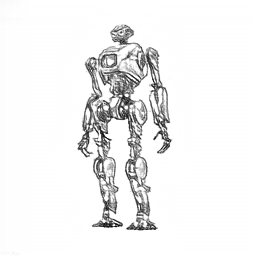} &
        \includegraphics[trim=4cm 1.9cm 4cm 2.1cm,clip,width=0.11\textwidth]{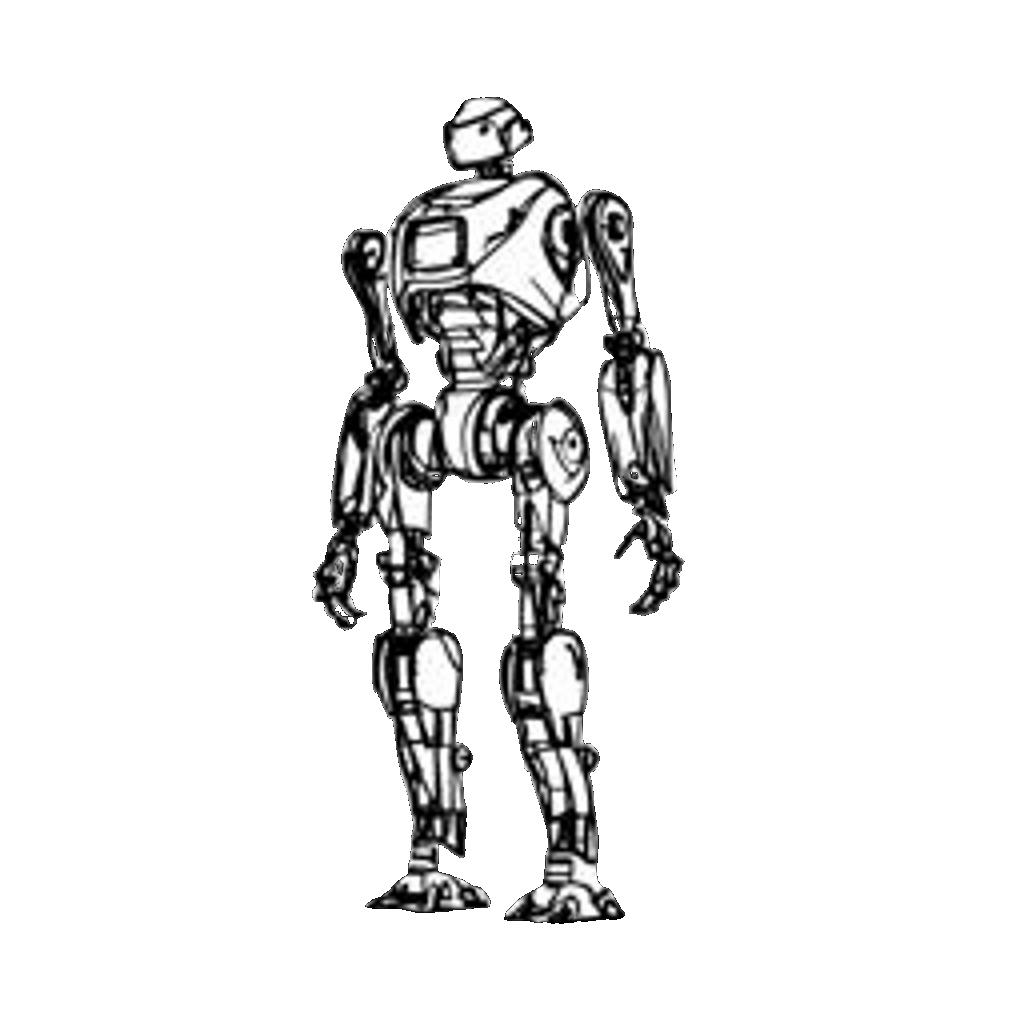} &
        \includegraphics[trim=1cm 0.45cm 1cm 0.6cm,clip,width=0.11\textwidth]{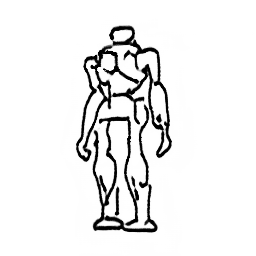} &
        \includegraphics[trim=1.6cm 0.7cm 1.6cm 0.8cm,clip,width=0.11\textwidth]{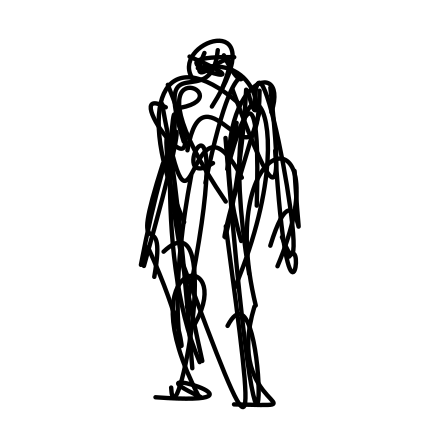} &
        \includegraphics[trim=1.6cm 0.7cm 1.6cm 0.8cm,clip,width=0.11\textwidth]{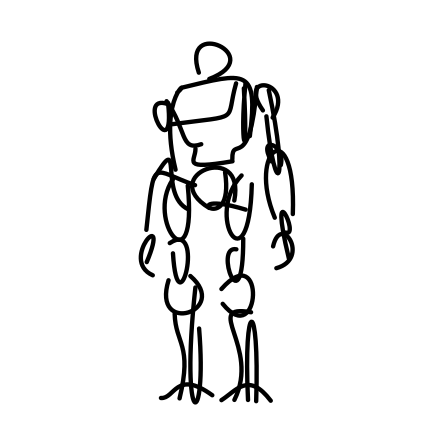}  &
        \includegraphics[trim=1.6cm 0.7cm 1.6cm 0.8cm,clip,width=0.11\textwidth]{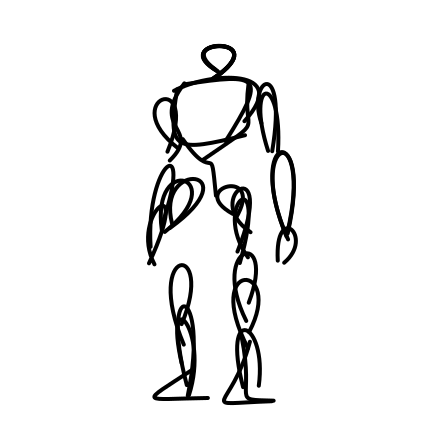} \\

        \includegraphics[trim=0 1.8cm 4cm 4cm,clip,width=0.11\textwidth]{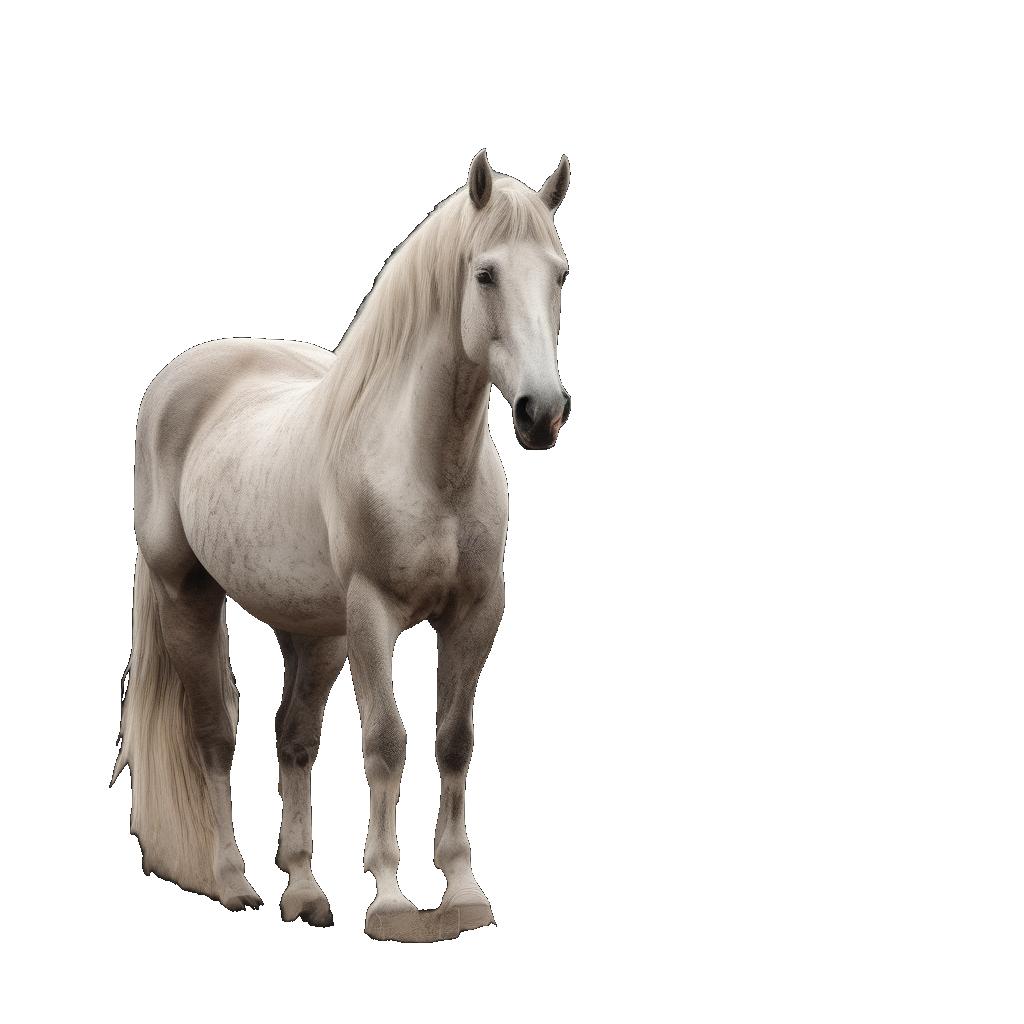} &
        \includegraphics[trim=0 1.8cm 4cm 4cm,clip,width=0.11\textwidth]{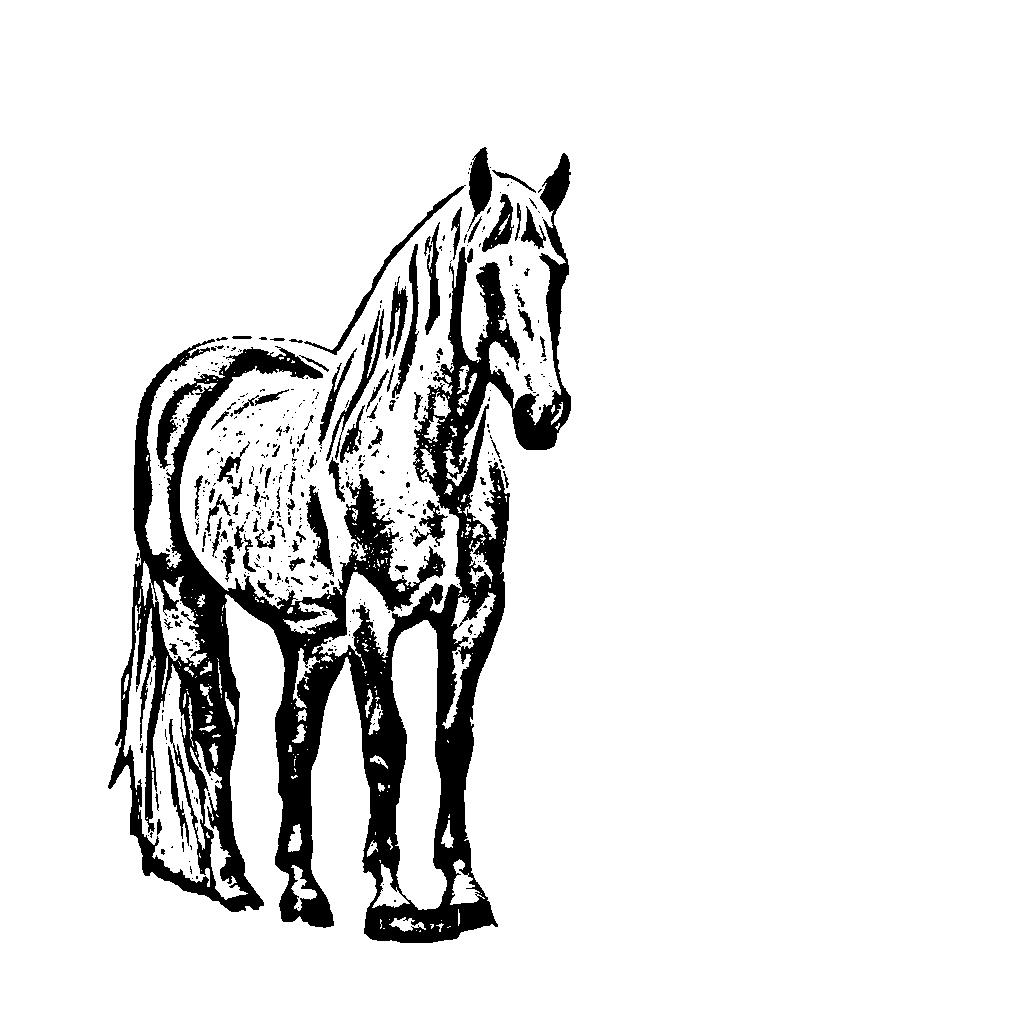} &
        
        \includegraphics[trim=0 0.5cm 1cm 1cm,clip,width=0.11\textwidth]{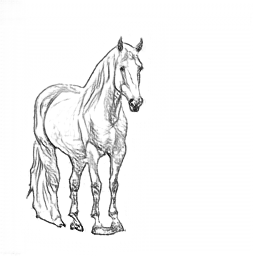} &
        \includegraphics[trim=0 1.8cm 4cm 4cm,clip,width=0.11\textwidth]{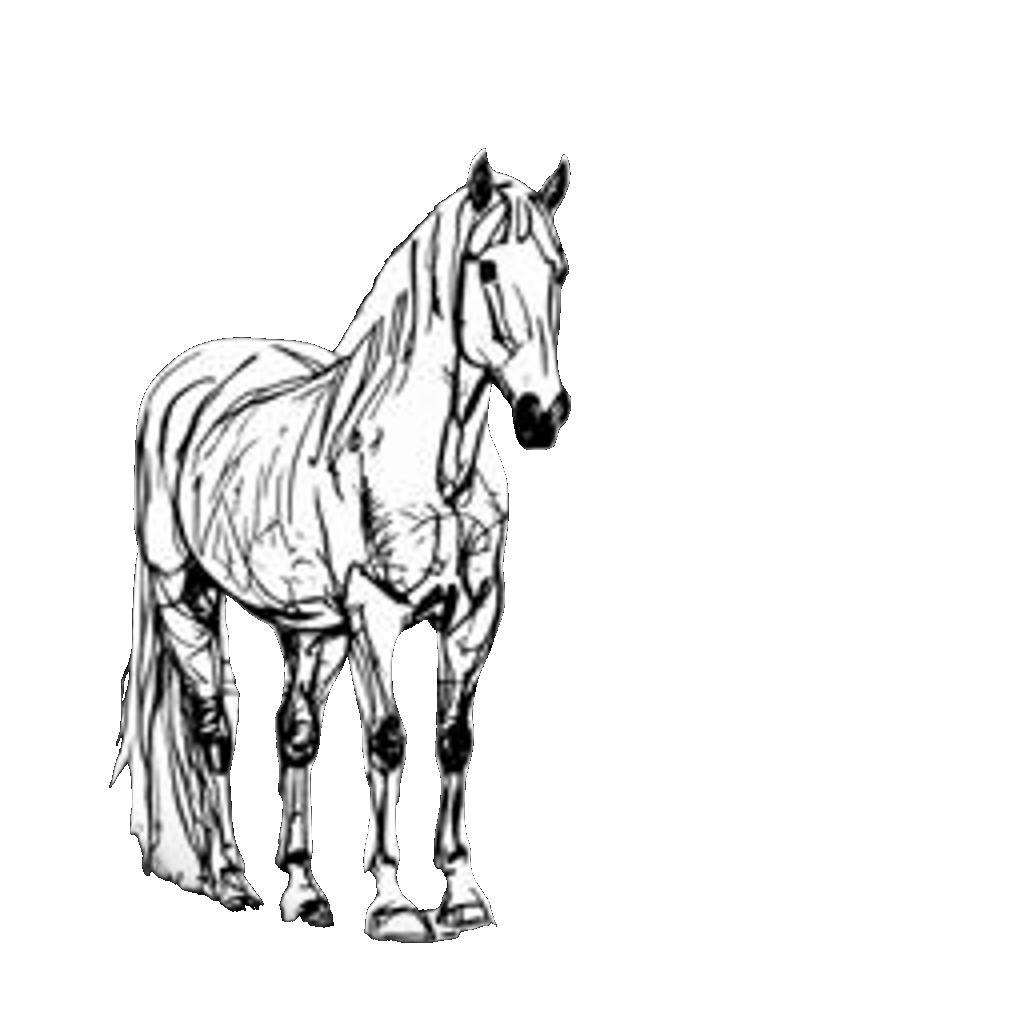} 
        &
        \includegraphics[trim=0 0.5cm 1cm 1cm,clip,width=0.11\textwidth]{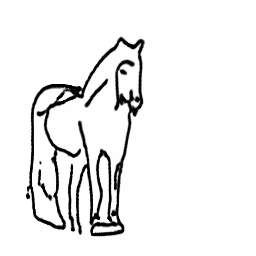} &
        \includegraphics[trim=0 0.9cm 1.7cm 1.6cm,clip,width=0.11\textwidth]{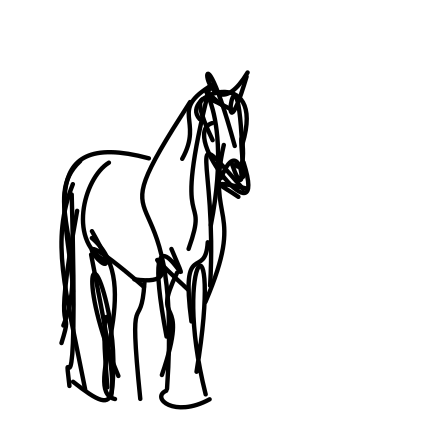} &
        \includegraphics[trim=0 0.9cm 1.6cm 1.6cm,clip,width=0.11\textwidth]{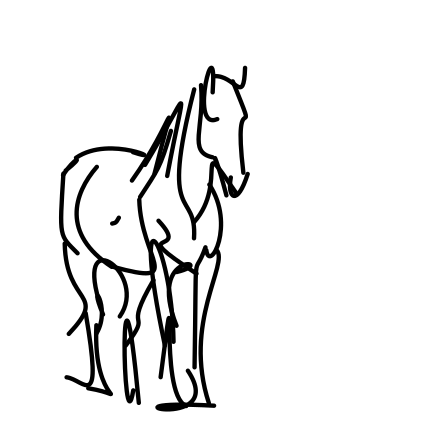}  &
        \includegraphics[trim=0 0.9cm 1.6cm 1.6cm,clip,width=0.11\textwidth]{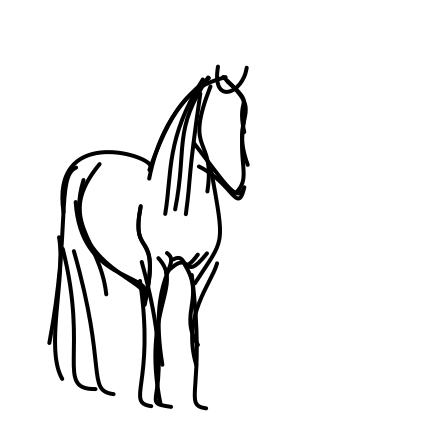} \\

        \includegraphics[trim=0 7cm 0 6cm,clip,width=0.11\textwidth]{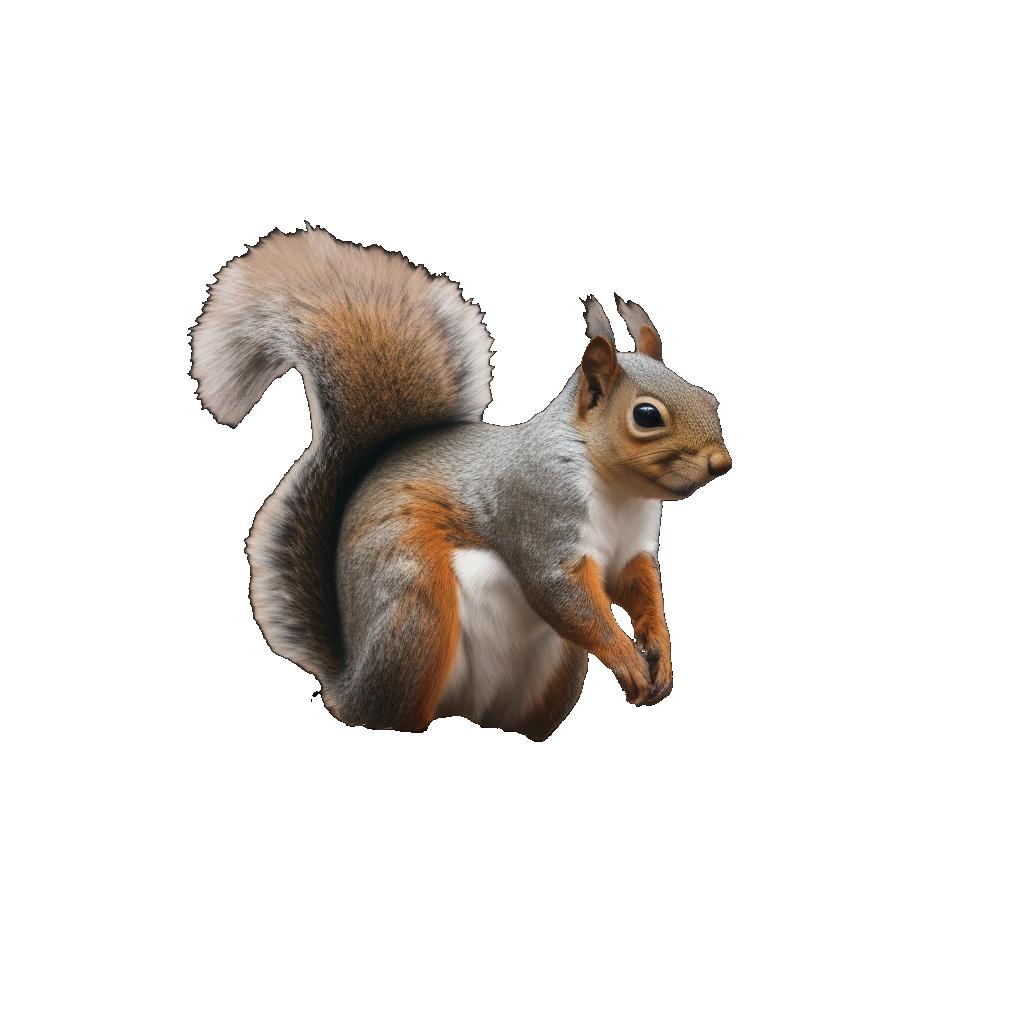} &
        \includegraphics[trim=0 7cm 0 6cm,clip,width=0.11\textwidth]{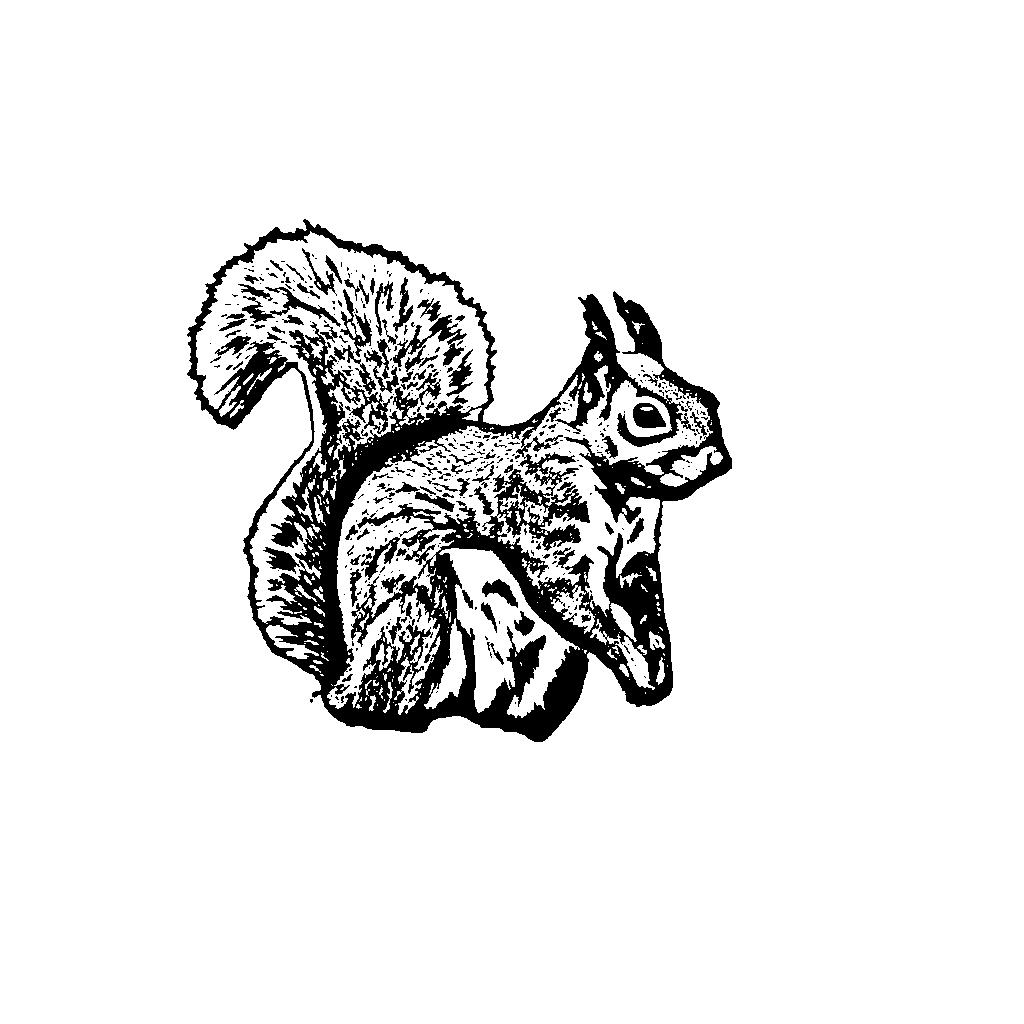} &
        
        \includegraphics[trim=0 1.8cm 0 1.5cm,clip,width=0.11\textwidth]{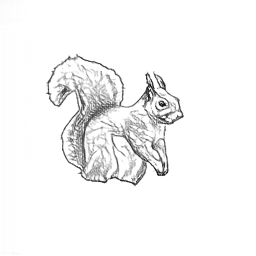} &
        \includegraphics[trim=0 7cm 0 6cm,clip,width=0.11\textwidth]{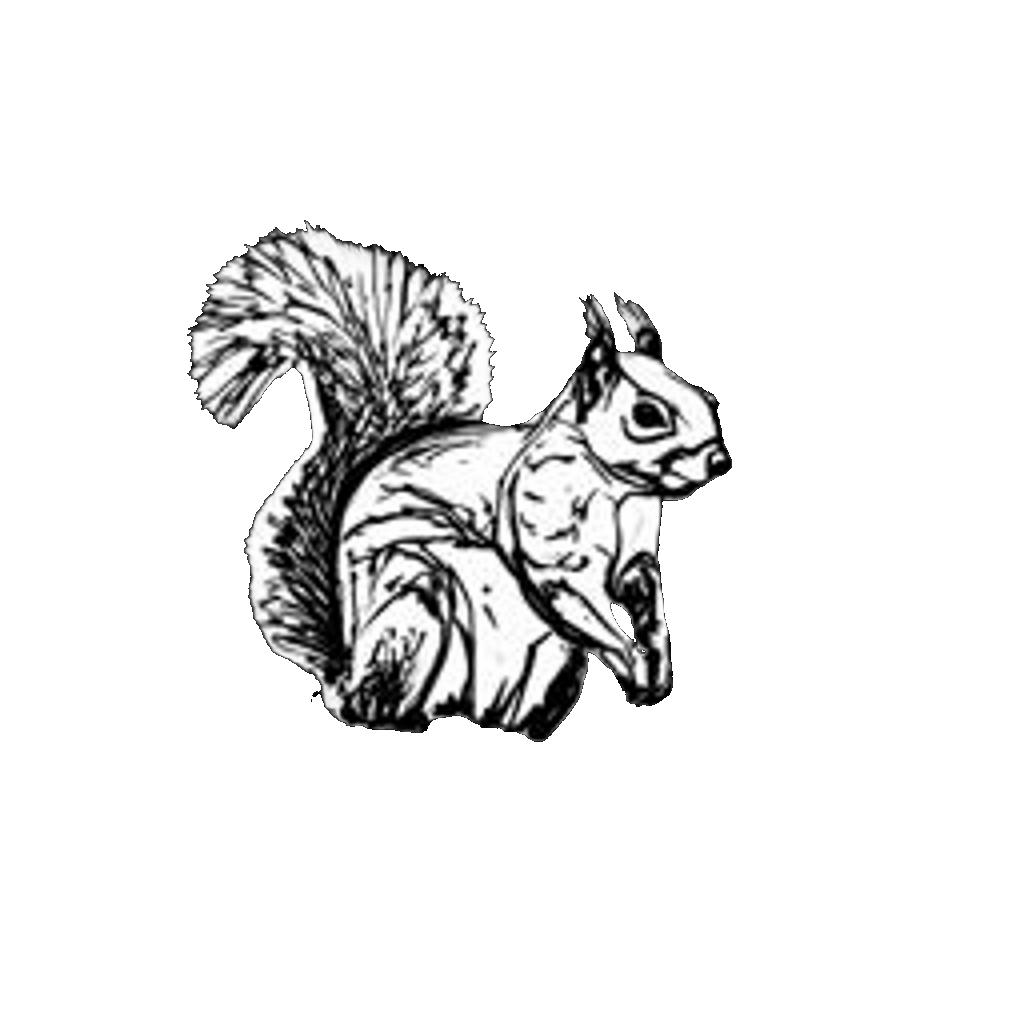}         &
        \includegraphics[trim=0 1.8cm 0 1.35cm,clip,width=0.11\textwidth]{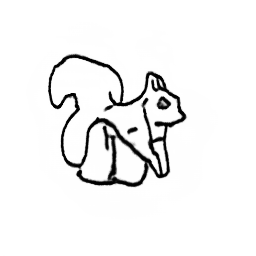} &
        \includegraphics[trim=0 3.5cm 0 2.1cm,clip,width=0.11\textwidth]{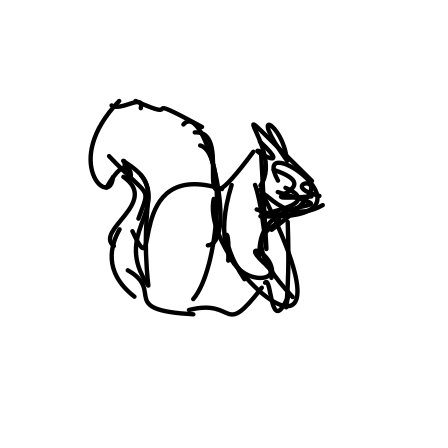} &
        \includegraphics[trim=0 3.5cm 0 2.1cm,clip,width=0.11\textwidth]{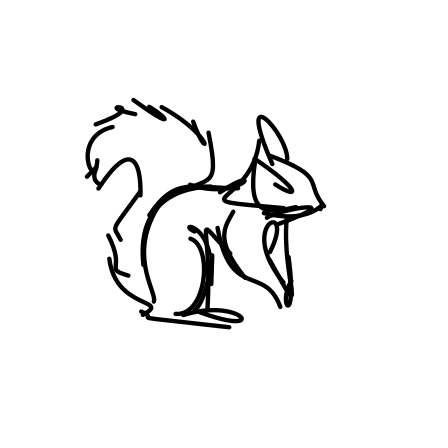}  &
        \includegraphics[trim=0 3.5cm 0 2.1cm,clip,width=0.11\textwidth]{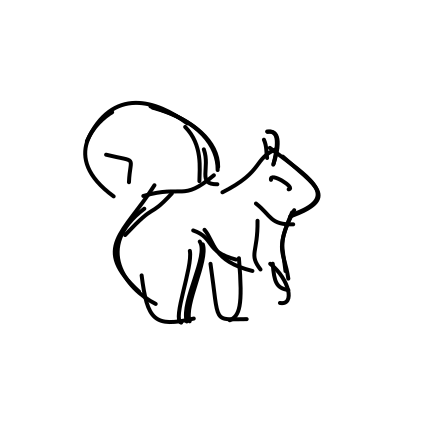}  \\

    \end{tabular}
    }
    \caption{Qualitative Comparison. Input images are shown on the left, with the time required to produce a single sketch and whether the sketches are in \textcolor{orange}{\textbf{P}}ixel or \textcolor{cyan}{\textbf{V}}ector format indicated at the top. From left to right, the sketches are generated using XDoG \cite{Winnemller2011XDoGAI}, PhotoSketching \cite{Li2019PhotoSketchingIC}, Chan et al. \cite{Chan2022LearningTG} (in anime style), InstantStyle \cite{Wang2024InstantStyleFL}, and CLIPasso \cite{vinker2022clipasso}. On the right are the resulting sketches from our proposed methods, ControlSketch and SwiftSketch.}
    \label{fig:comparison}
\end{figure*}

\section{Results}
\begin{figure*}
    \centering
    \includegraphics[width=0.87\linewidth]{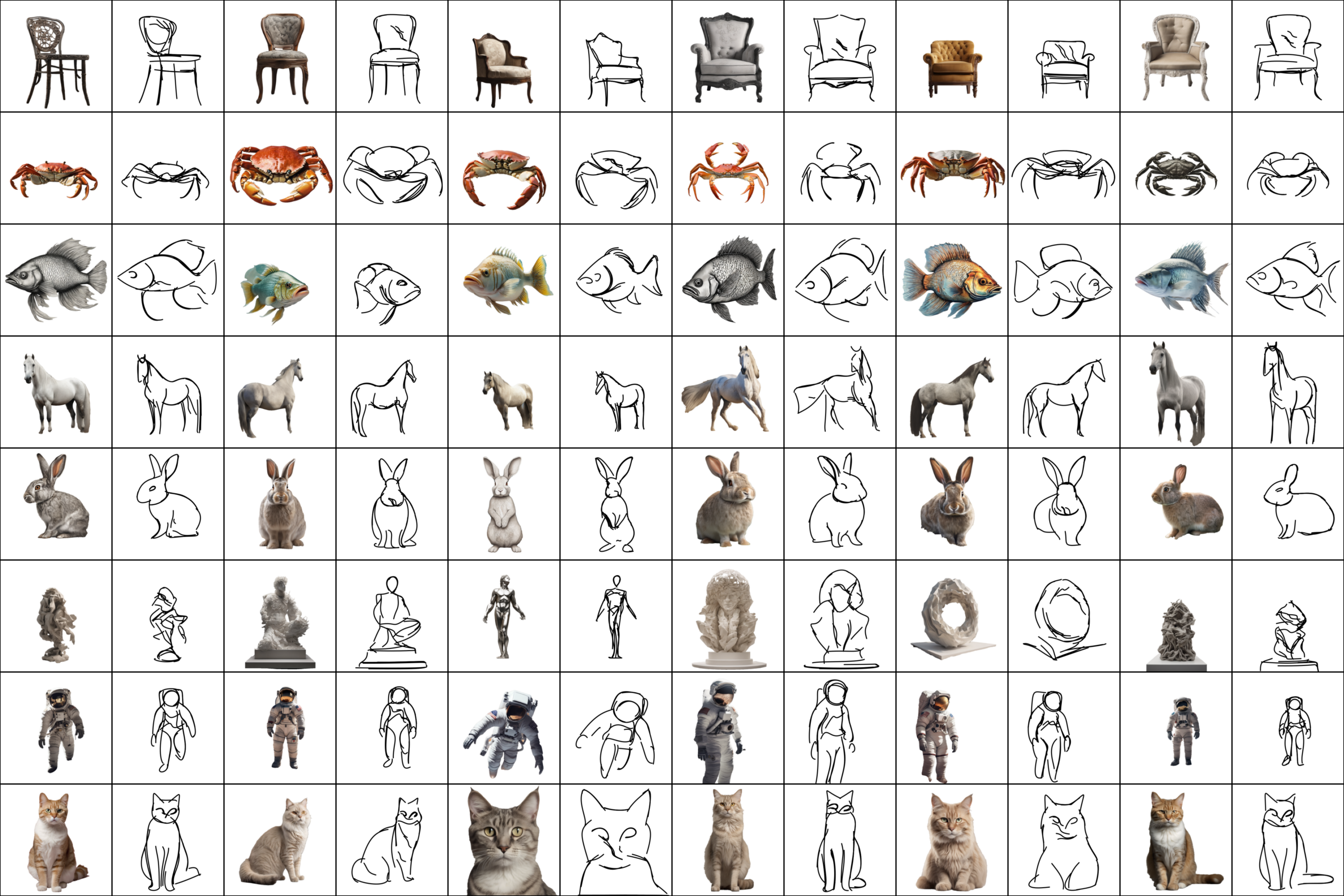}
    \vspace{-0.2cm}
    \caption{Sketches generated by SwiftSketch for seen categories, using input images not included in the training data.}
    \label{fig:qualitative-swift-seen}
\end{figure*}

\begin{figure*}
    \centering
    \includegraphics[width=0.87\linewidth]{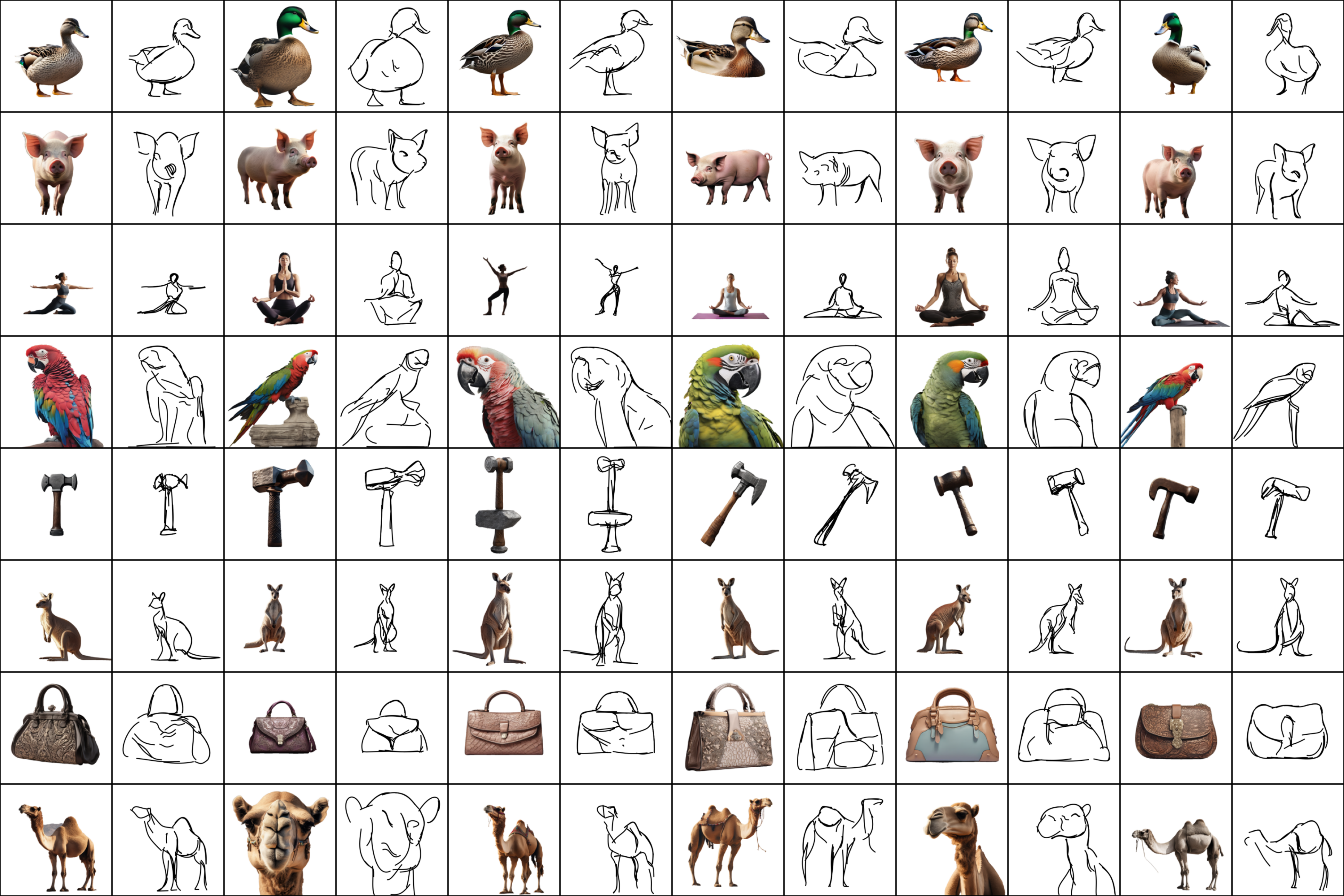}
    \vspace{-0.2cm}
    \caption{Sketches generated by SwiftSketch for unseen categories.}
    \label{fig:qualitative-swift-unseen}
\end{figure*}

We begin by showcasing \methodname's ability to generate high-quality vector sketches for a diverse set of input images.
SwiftSketch successfully generalizes to unseen images within the training categories (\Cref{fig:qualitative-swift-seen}), creating sketches that depict the input images well while demonstrating a plausible and detailed appearance. On images of unseen categories that pose greater challenges, SwiftSketch effectively captures the essential features of the input images, producing abstract yet faithful representations (\Cref{fig:qualitative-swift-unseen}). Notably, all sketches are provided in vector format, and are generated in just 50 diffusion steps, followed by a single refinement step, with the entire process taking less than one second.
In \Cref{fig:denoising}, we illustrate the denoising steps of the generation process, starting from a Gaussian distribution and progressively refining towards the data distribution. 
In \Cref{fig:order}, we demonstrate the ability of our method to create level-of-abstraction using our ordered stroke technique. We visualize the progressive addition of strokes in the sequence they appear in the output SVG file. Note how the first strokes already convey the intended concept effectively. Additional results of both SwiftSketch and ControlSketch are available in the supplementary.

\subsection{Comparisons}
We evaluate the performance of SwiftSketch and ControlSketch with respect to state-of-the-art methods for image-to-sketch generation, including Photo-Sketching \cite{Li2019PhotoSketchingIC}, Chan \etal \cite{Chan2022LearningTG}, InstantStyle \cite{Wang2024InstantStyleFL}, and CLIPasso \cite{vinker2022clipasso}. InstantStyle is applied with a sketch image as the style reference.
\Cref{fig:comparison} shows representative results from each method, with XDoG \cite{Winnemller2011XDoGAI}, a classic edge detection technique, shown on the left as a baseline.
The sketches of Chan \etal and InstantStyle are detailed and align well with the overall structure of the input images. However note that they closely follow the edge maps shown on the left. The sketches of Photo-Sketching (fifth column) are more abstract, but can fail to effectively capture the images' content in a natural way. While these approaches are efficient, producing sketches in less than a minute, they focus on generating \emph{raster} sketches. In contrast, our method produces vector sketches, which are resolution-independent, easily editable, and exhibit a smooth, clean style.
CLIPasso (sixth column) generates vector sketches that achieve a good balance between fidelity and semantics. However, it is significantly slower, requiring 5 minutes to produce a single sketch, and it may introduce artifacts, such as the noisy overlapping strokes observed in the robot example.
ControlSketch (seventh column) produces high-fidelity sketches that remain abstract, smooth, and natural, effectively depicting the input images while avoiding artifacts. However, it is even slower than CLIPasso, as SDS-based methods generally require more time to converge, making it impractical for interactive applications.
SwiftSketch, shown in the rightmost column, successfully learns the data distribution from ControlSketch samples, enabling it to produce sketches that approach the quality of optimization-based techniques but in real time.
Additional results are available in the supplamentary material.

\begin{table*}
\small
\setlength{\tabcolsep}{4pt}
\centering
\caption{Quantitative Comparison of Sketch Generation Methods. The scores for Top-1 and Top-3 CLIP recognition accuracy, MS-SSIM, and DreamSim are presented. Results are based on 6,000 random samples: 2,000 from our dataset's test set, 2,000 from unseen categories in our dataset, and 2,000 from the external SketchyCOCO dataset \cite{SketchyCOCO2020}. The best scores in each column are marked in bold.}\vspace{-0.2cm}
\begin{tabular}{l l c c c c c c c c c c c c} 
    \toprule
    & & \multicolumn{3}{c}{CLIP Top-1 $\uparrow$} & \multicolumn{3}{c}{CLIP Top-3 $\uparrow$}  & \multicolumn{3}{c}{MS-SSIM $\uparrow$} & \multicolumn{3}{c}{DreamSim $\downarrow$}  \\
    \cmidrule(lr){3-5}\cmidrule(lr){6-8}\cmidrule(lr){9-11}\cmidrule(lr){12-14}
    & Time & Seen & Unseen & Exter. & Seen & Unseen & Exter. & Seen & Unseen & Exter. & Seen & Unseen & Exter. \\
    \cmidrule(lr){1-1}\cmidrule(lr){2-2}\cmidrule(lr){3-3}\cmidrule(lr){4-4}\cmidrule(lr){5-5}\cmidrule(lr){6-6}\cmidrule(lr){7-7}\cmidrule(lr){8-8}\cmidrule(lr){9-9}\cmidrule(lr){10-10}\cmidrule(lr){11-11}\cmidrule(lr){12-12}\cmidrule(lr){13-13}\cmidrule(lr){14-14}
    Human \cite{SketchyCOCO2020} & $-$ & $-$ & $-$ & $0.85$ & $-$ & $-$ & $0.93$ & $-$ & $-$ & \textcolor{black}{$0.16$} & $-$ & $-$ & \textcolor{black}{$0.66$} \\
    Chan \etal (Anime) \cite{Chan2022LearningTG} & $\approx$ 0.04 sec. & $0.94$  & \textcolor{black}{$\mathbf{0.99}$} & $0.87$ & \textcolor{black}{$\mathbf{0.99}$} & \textcolor{black}{$\mathbf{0.99}$} & $0.96$ & $0.79$ & $0.77$ & $0.45$ & \textcolor{black}{$\mathbf{0.32}$} & \textcolor{black}{$\mathbf{0.43}$} & \textcolor{black}{$\mathbf{0.45}$} \\
    Chan \etal (Contour) \cite{Chan2022LearningTG} & $\approx$ 0.04 sec. & $0.96$  & $0.92$ & $0.80$ & $0.98$ & $0.95$ & $0.91$ & $0.77$ & $0.73$ & $0.49$ &  $0.51$ & $0.59$ & $0.54$  \\
    InstantStyle \cite{Wang2024InstantStyleFL} & $\approx$ 1 min. & $0.97$ & \textcolor{black}{$\mathbf{0.99}$} & $-$  & \textcolor{black}{$\mathbf{0.99}$} & \textcolor{black}{$\mathbf{0.99}$} & $-$ & \textcolor{black}{$\mathbf{0.89}$} & \textcolor{black}{$\mathbf{0.90}$} & $-$ & $0.36$ & $0.44$ & $-$ \\
    Photo-Sketching \cite{Li2019PhotoSketchingIC} & $\approx$ 0.6 sec. & \textcolor{black}{$0.90$} & $0.81$ &  $0.65$ & \textcolor{black}{$0.95$} & $0.87$ & $0.78$ & $0.61$ & $0.55$ & $0.40$ & $0.58$ & $0.65$ & $0.63$ \\
    CLIPasso \cite{vinker2022clipasso} & $\approx$ 5 min. & \textcolor{black}{$\mathbf{0.98}$}  & $0.97$ & $0.88$ & $\mathbf{0.99}$ & $\mathbf{0.99}$ & $0.95$ & $0.65$ & $0.60$ & $0.52$ & $0.48$ & $0.57$ & $0.57$  \\
    \cmidrule(lr){1-1}\cmidrule(lr){2-14}
    ControlSketch & $\approx$ 10 min. & $0.97$ & $0.97$ & \textcolor{black}{$\mathbf{0.91}$}  & \textcolor{black}{$\mathbf{0.99}$} & \textcolor{black}{$\mathbf{0.99}$} &  \textcolor{black}{$\mathbf{0.97}$} & $0.68$ & $0.63$ & $\mathbf{0.53}$ & $0.52$ & $0.59$  & $0.60$ \\
    SwiftSketch & $\approx$ 0.5 sec. & $0.95$ & \textcolor{black}{$0.70$} & \textcolor{black}{$0.56$} & $0.98$ & \textcolor{black}{$0.82$} &  $0.70$ & $0.62$ & $0.56$ & $0.47$ & $0.53$ & $0.66$  & $0.64$ \\
    
    \bottomrule
\end{tabular}
\label{tb:clip_metrics}
\end{table*}

\paragraph{Quantitative Evaluation} We sample 4,000 images from our dataset (2,000 from our test set of categories seen during training and 2,000 from unseen categories) and additional 2,000 images from the SketchyCOCO \cite{SketchyCOCO2020} dataset to assess generalization on external data. Each set consists of 10 randomly selected categories with 200 images per category. 
Following common practice in the field, we use the CLIP zero-shot classifier \cite{Radfordclip} to assess class-level recognition, MS-SSIM \cite{wang2003multiscale} for image-sketch fidelity following the settings proposed in CLIPascene \cite{Vinker2022CLIPasceneSS}, and DreamSim \cite{fu2023dreamsim}. 
The results are presented in \Cref{tb:clip_metrics}, where scores for each data type are reported separately, with human sketches from the SketchyCOCO dataset included as a baseline.
Chan \etal and InstantStyle achieve the highest scores across most metrics due to their highly detailed sketches, which closely resemble the image's edge map. This level of detail ensures that their sketches are both easily recognizable as depicting the correct class (as indicated by the CLIP score) and exhibit high fidelity (as reflected in other measurements).
The results show that SwiftSketch generalizes well to test set images from seen categories, as evidenced by its similar scores to ControlSketch (which serves as the ground truth in our case). However, its performances decrease for unseen categories, particularly in class-based recognition. This is especially apparent on the SketchyCOCO dataset, which is highly challenging due to its low-resolution images and difficult lighting conditions. It is important to note that SwiftSketch is trained on only 15 image categories due to limited resources, suggesting that more extensive training could improve its generalization capabilities.

The results demonstrate that ControlSketch produces sketches that are both highly recognizable and of high fidelity, outperforming alternative methods, particularly on the SketchyCOCO dataset. To further highlight the advantages of ControlSketch over CLIPasso, we conduct a two-alternative forced-choice (2AFC) perceptual study with 40 participants. Each participant was shown pairs of sketches generated by the two methods (presented in random order) alongside the input image and asked to choose the sketch they perceived to be of higher quality. The study included 24 randomly selected sketches from both our dataset and SketchyCOCO, spanning 24 object classes. Participants rated sketches generated by ControlSketch as higher quality in 89\% of cases. Examples of sketches presented in the user study are shown in \Cref{fig:comparison_opt}.

\section{Ablation}
We evaluate the contribution of SwiftSketch's main components by systematically removing each one and retraining the network. Specifically, we examine the impact of excluding the LPIPS loss, the L1 loss, and the sorting technique, as well as the effect of incorporating the refinement network.
The results are summarized in \Cref{tb:ablation_metrics}, where ``Full'' represents our complete diffusion pipeline prior to refinement, and ``+Refine'' denotes the inclusion of the refinement stage.
Notably, removing the L1 loss results in a significant drop in performance, highlighting its essential role in the training process. Excluding the LPIPS loss negatively impacts performance, particularly in unseen classes. 
The metrics indicate comparable performance in the absence of the sorting stage. While the resulting sketches may appear visually similar, the sorting stage is crucial for supporting varying levels of abstraction. Although the network can be trained without this stage and still achieve reasonable results, learning an internal stroke order provides a foundation for training across abstraction levels, where sketches implicitly encode the importance of strokes.
The refinement stage enhances recognizability, especially in unseen categories where the output sketches from the diffusion process are noisier. We further illustrate the impact of the refinement network in \Cref{fig:refine_ablation}, with additional results provided in the supplementary.

\begin{table}
\small
\setlength{\tabcolsep}{1.7pt}
\addtolength{\belowcaptionskip}{-5pt}
\centering
\caption{Ablation Study. We systematically remove each component in our pipeline and retrain our network. Scores are computed on 4000 sketches in total: 2000 from seen categories and 2000 from unseen categories.} \vspace{-0.2cm}
\begin{tabular}{l c c c c c c c c c } 
    \toprule
    & \multicolumn{2}{c}{CLIP Top-1 $\uparrow$} & \multicolumn{2}{c}{CLIP Top-3 $\uparrow$}  & \multicolumn{2}{c}{MS-SSIM $\uparrow$} & \multicolumn{2}{c}{DreamSim $\downarrow$}  \\
    \cmidrule(lr){2-3}\cmidrule(lr){4-5}\cmidrule(lr){6-7}\cmidrule(lr){8-9}
    
     & Seen & Unseen  & Seen & Unseen & Seen & Unseen  & Seen & Unseen  \\
    \cmidrule(lr){1-1}\cmidrule(lr){2-2}\cmidrule(lr){3-3}\cmidrule(lr){4-4}\cmidrule(lr){5-5}\cmidrule(lr){6-6}\cmidrule(lr){7-7}\cmidrule(lr){8-8}\cmidrule(lr){9-9}

    w/o LPIPS  & $0.88$ & $0.36$ & $0.94$ & $0.49$ & $0.58$ & $0.52$ & $0.58$ & $0.72$  \\
    w/o L1  & $0.03$ & $0.01$ & $0.06$ & $0.03$ & $0.24$ & $0.22$ & $0.80$ & $0.84$  \\
    w/o Sort  & $0.93$ & $0.63$ & $0.97$ & $0.76$ & \textcolor{black}{$\mathbf{0.62}$} &  \textcolor{black}{$\mathbf{0.57}$}  & $0.55$ & $0.68$  \\
    \cmidrule(lr){1-1}\cmidrule(lr){2-9}
    \makecell[l]{Full}  & $0.94$ & $0.61$ & $0.97$ & $0.73$ & \textcolor{black}{$\mathbf{0.62}$}  &  \textcolor{black}{$\mathbf{0.57}$}  & $0.54$ & $0.68$  \\
    
    +Refine  & \textcolor{black}{$\mathbf{0.95}$}  &  \textcolor{black}{$\mathbf{0.70}$}   & \textcolor{black}{$\mathbf{0.98}$}  &  \textcolor{black}{$\mathbf{0.82}$}  &  \textcolor{black}{$\mathbf{0.62}$}  & $0.56$  &  \textcolor{black}{$\mathbf{0.53}$}  & \textcolor{black}{$\mathbf{0.66}$}  \\
    
    \bottomrule
\end{tabular}
\label{tb:ablation_metrics}
\end{table}

\section{Limitations and Future Work}
\begin{figure}[h]
    \centering
    \includegraphics[width=1\linewidth]{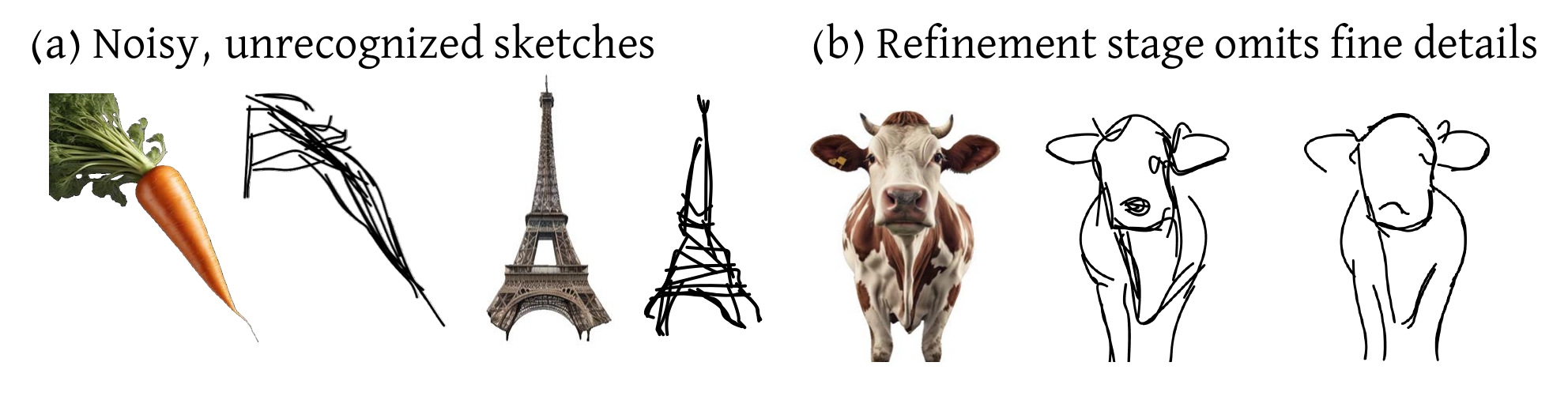}\vspace{-0.45cm}
    \caption{Limitations. (a) Sketches may appear unrecognizable (e.g., carrot) or noisy (e.g., Eiffel Tower). (b) The refinement stage can lead to the loss of fine details, such as the cow's nose and eye.}
    \label{fig:limitations}
\end{figure}

While SwiftSketch can generate vector sketches from images efficiently, it comes with limitations.
First, although SwiftSketch performs well on seen categories, as evidenced by our evaluation, its performance decreases for unseen categories. This is particularly apparent in categories that differ significantly from those seen during training (e.g., non-human or non-animal objects). Failure cases often exhibit a noisy appearance or are entirely unrecognizable, such as the carrot in \Cref{fig:limitations}. Expanding the number of training categories in future work could enhance the model's generalization.
Second, our refinement stage, which is meant to fix the noisy appearance, might over-simplify the sketches, resulting in lost details such as the nose and eyes of the cow in \cref{fig:limitations}.
Lastly, in the scope of this paper, we trained SwiftSketch on sketches with a fixed number of strokes (32). Extending the training to sketches with varying numbers of strokes, spanning multiple levels of abstraction, presents an exciting direction for future research. Our transformer-decoder architecture is inherently suited for such an extension, and our results show that the network can capture essential features from sorted sketches, highlighting its potential to effectively handle more challenging levels of abstraction.

\section{Conclusions}
We introduced SwiftSketch, a method for object sketching capable of generating plausible \emph{vector} sketches in under a second. SwiftSketch employs a diffusion model with a transformer-decoder architecture, generating sketches by progressively denoising a Gaussian distribution in the space of stroke control points.
To address the scarcity of professional-quality paired vector sketch datasets, we constructed a synthetic dataset spanning 100 classes and over 35,000 sketches. This dataset was generated using ControlSketch, an improved SDS-based sketch generation method enhanced with a depth ControlNet for better spatial control. We demonstrated both visually and numerically that ControlSketch produces high-quality, high-fidelity sketches and that SwiftSketch effectively learns the data distribution of ControlSketch, achieving high-quality sketch generation while reducing generation time from approximately 10 minutes to 0.5 seconds.
We believe this work represents a meaningful step toward real-time, high-quality vector sketch generation with the potential to enable more interactive processes. Additionally, our extensible dataset construction process will be made publicly available to support future research in this field.

\section{Acknowledgements}
We thank Guy Tevet and Oren Katzir for their valuable insights and engaging discussions. We also thank Yuval Alaluf, Elad Richardson, and Sagi Polaczek for providing feedback on early versions of our manuscript. 
This work was partially supported by Joint NSFC-ISF Research Grant no. 3077/23 and Isf 3441/21.

\begin{figure*}[t]
    \centering
    \setlength{\tabcolsep}{1pt}
    {\small
    \begin{tabular}{c c c |c c c |c c c}
         Input & ControlSk. &  CLIPasso  & Input  & ControlSk. &  CLIPasso & Input  & ControlSk. &  CLIPasso  \\
        \includegraphics[width=0.1\linewidth]{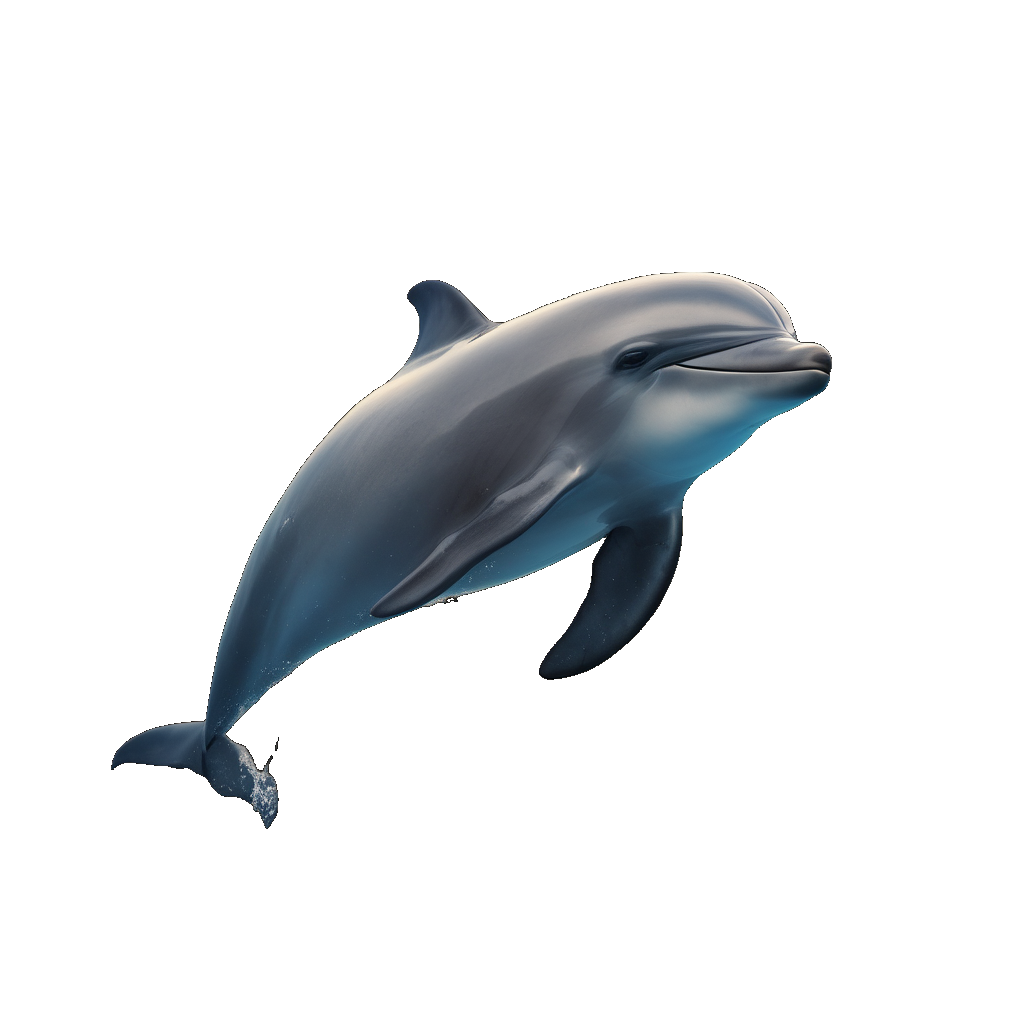} &
        \includegraphics[width=0.1\linewidth]{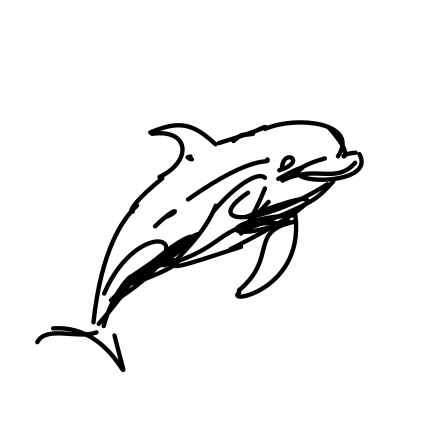} &
       \includegraphics[width=0.1\linewidth]{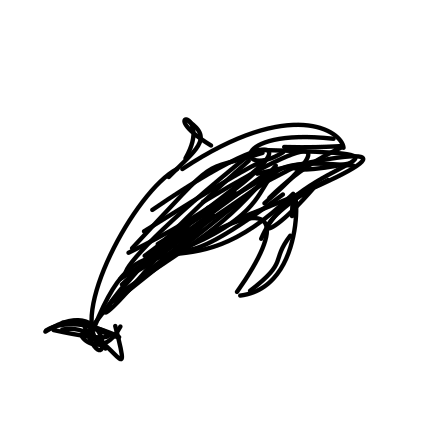} &
       \includegraphics[width=0.1\linewidth]{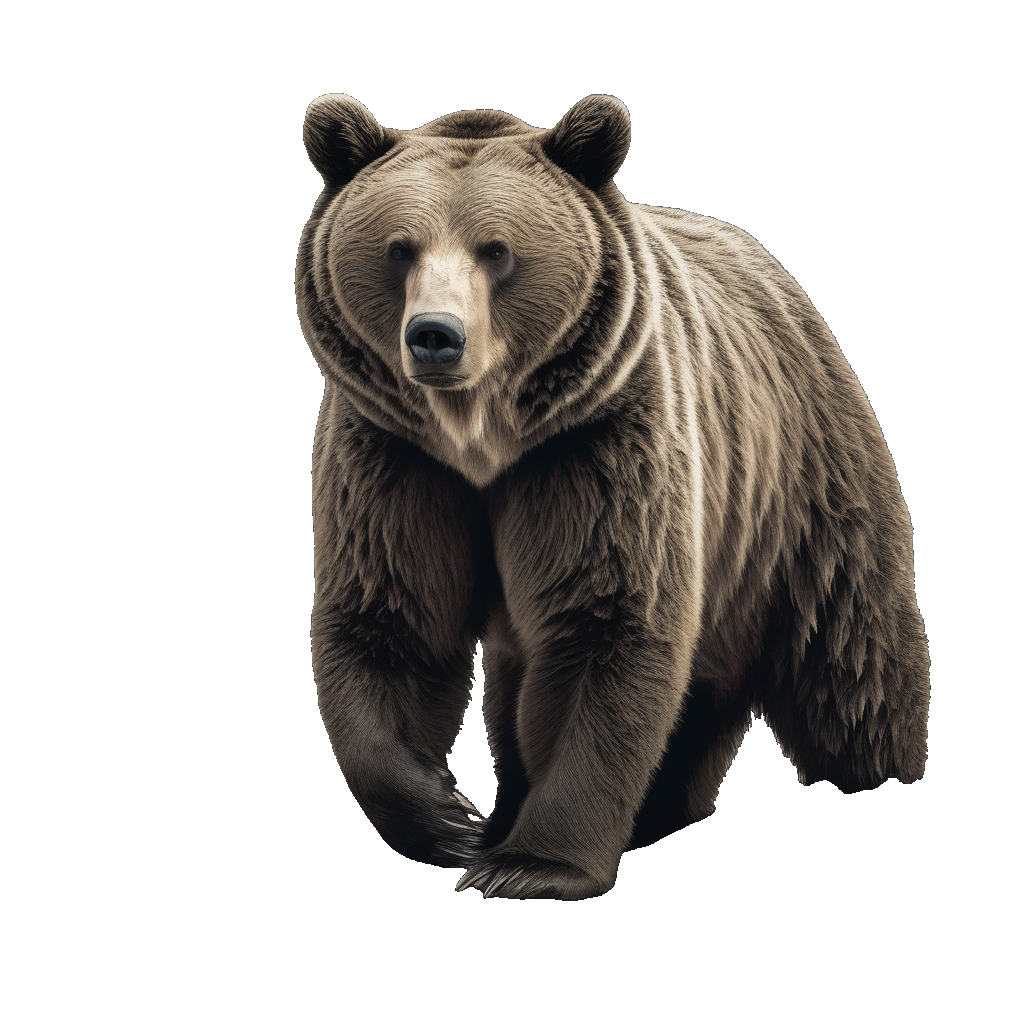} &
       \includegraphics[width=0.1\linewidth]{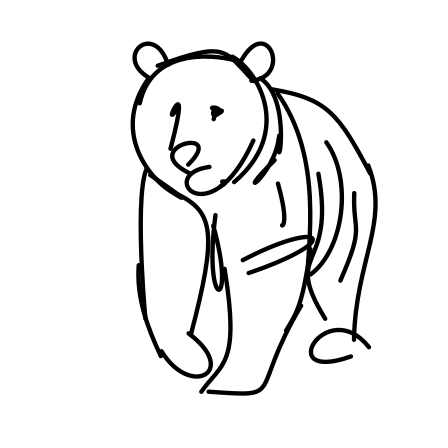} &
       \includegraphics[width=0.1\linewidth]{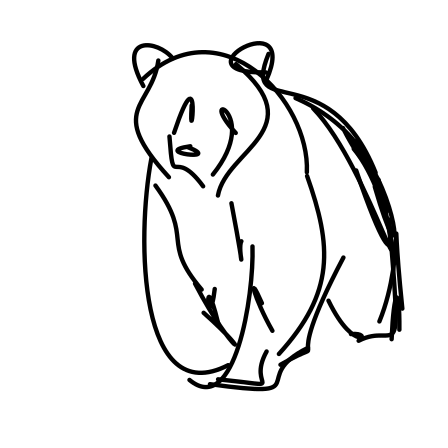} &

        \includegraphics[width=0.1\linewidth]{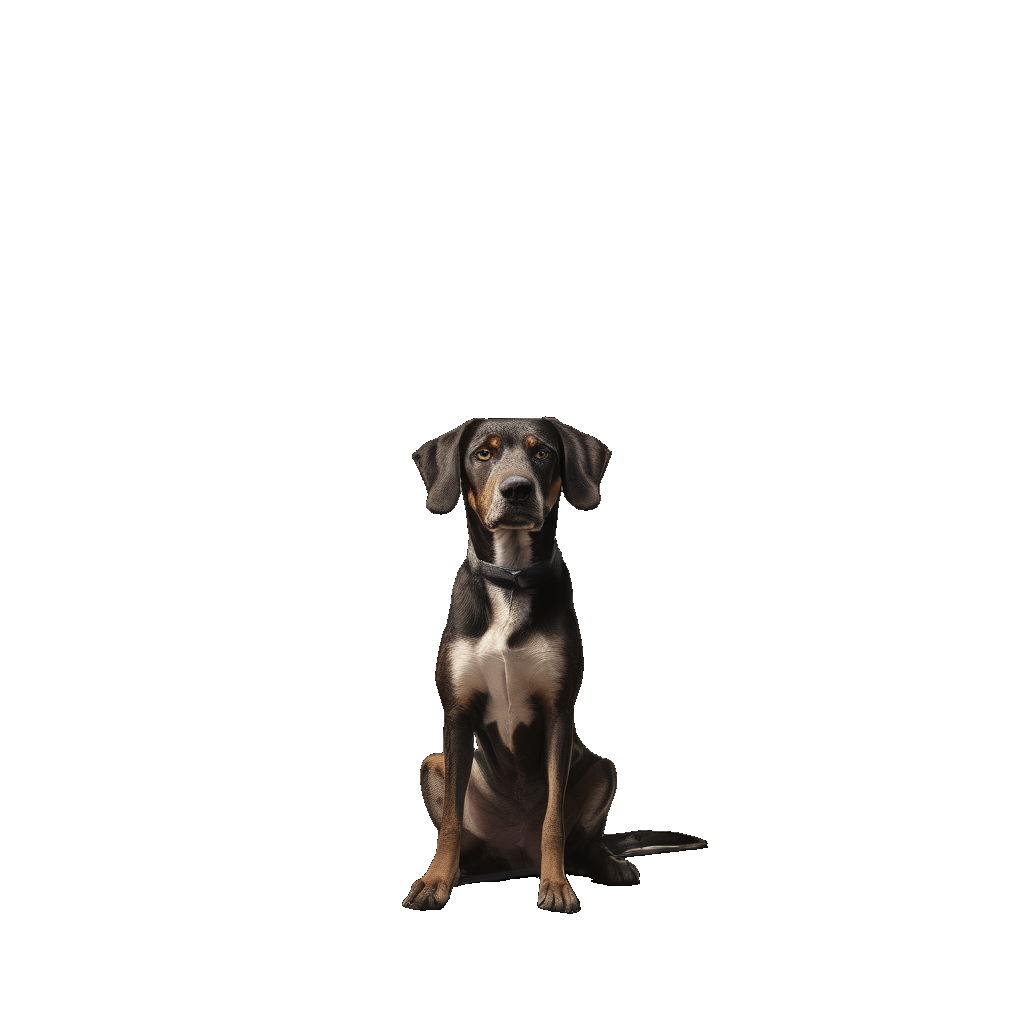} &
        \includegraphics[width=0.1\linewidth]{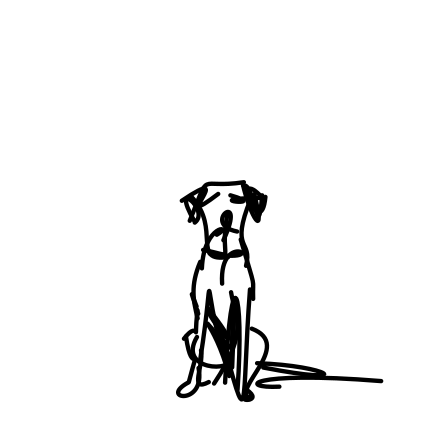} &
       \includegraphics[width=0.1\linewidth]{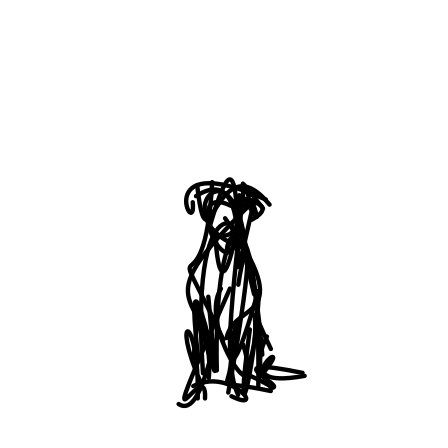}  \\
       Input & ControlSk. &  CLIPasso  & Input  & ControlSk. &  CLIPasso & Input  & ControlSk. &  CLIPasso  \\
       \includegraphics[width=0.1\linewidth]{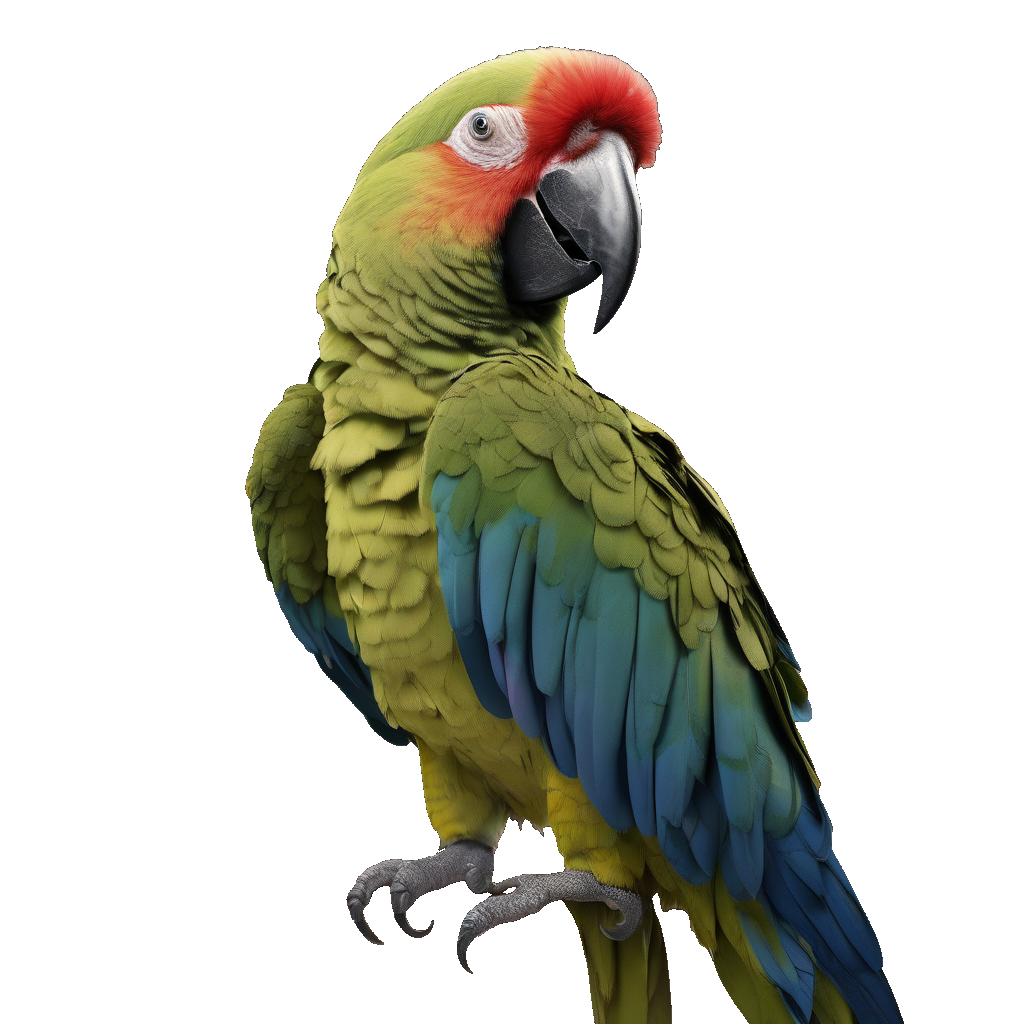} &
        \includegraphics[width=0.1\linewidth]{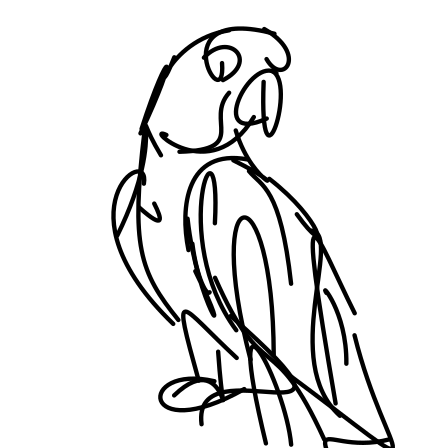} &
       \includegraphics[width=0.1\linewidth]{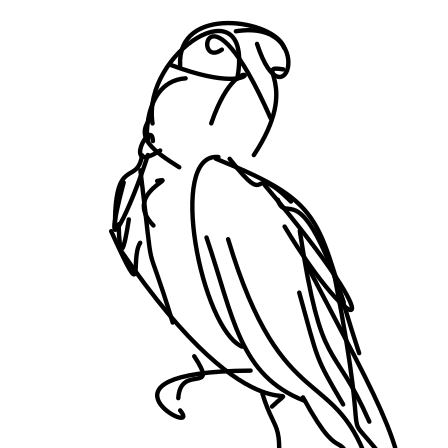} &
        \includegraphics[width=0.1\linewidth]{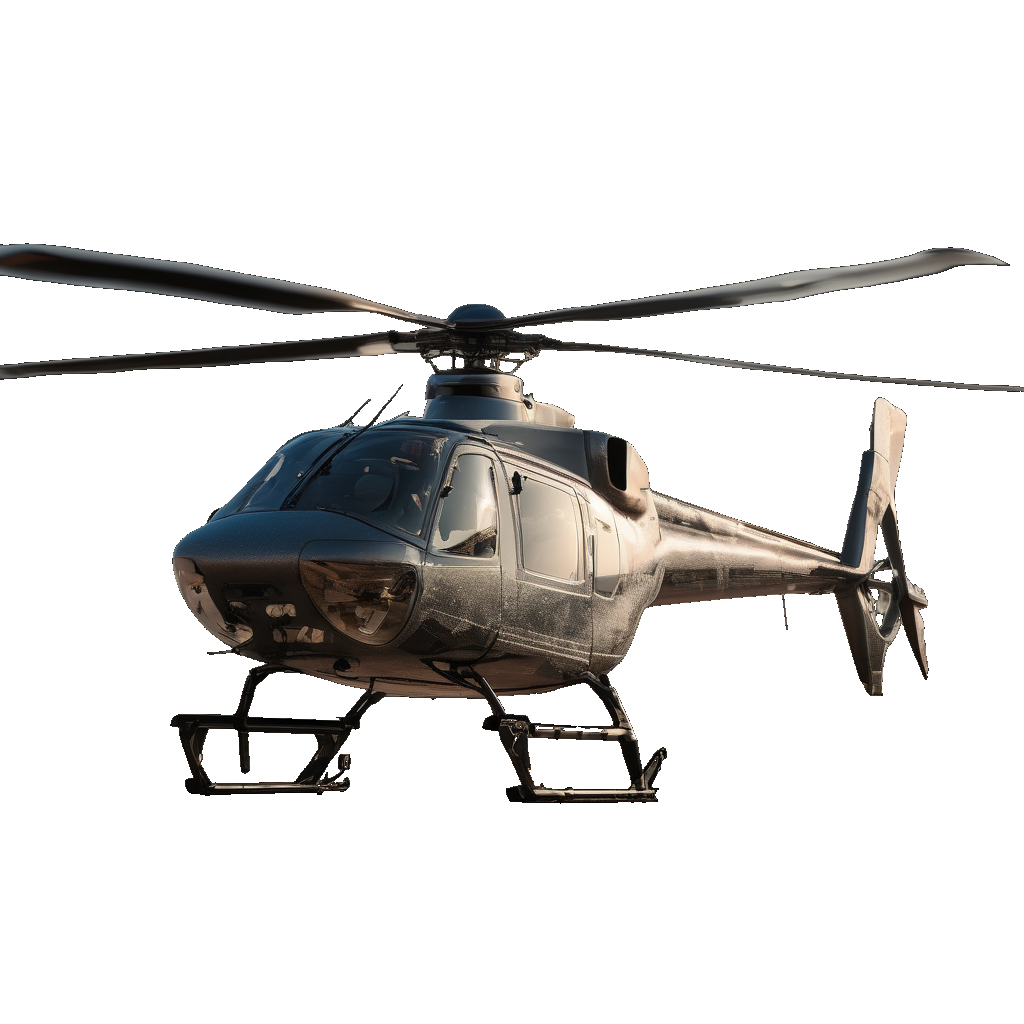} &
        \includegraphics[width=0.1\linewidth]{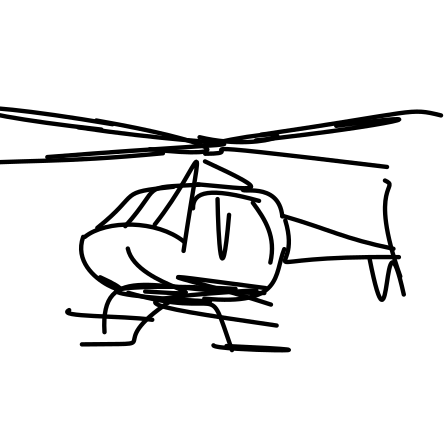} &
       \includegraphics[width=0.1\linewidth]{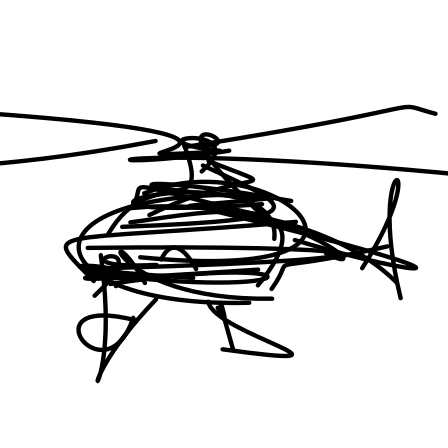}  &
       \includegraphics[width=0.1\linewidth]{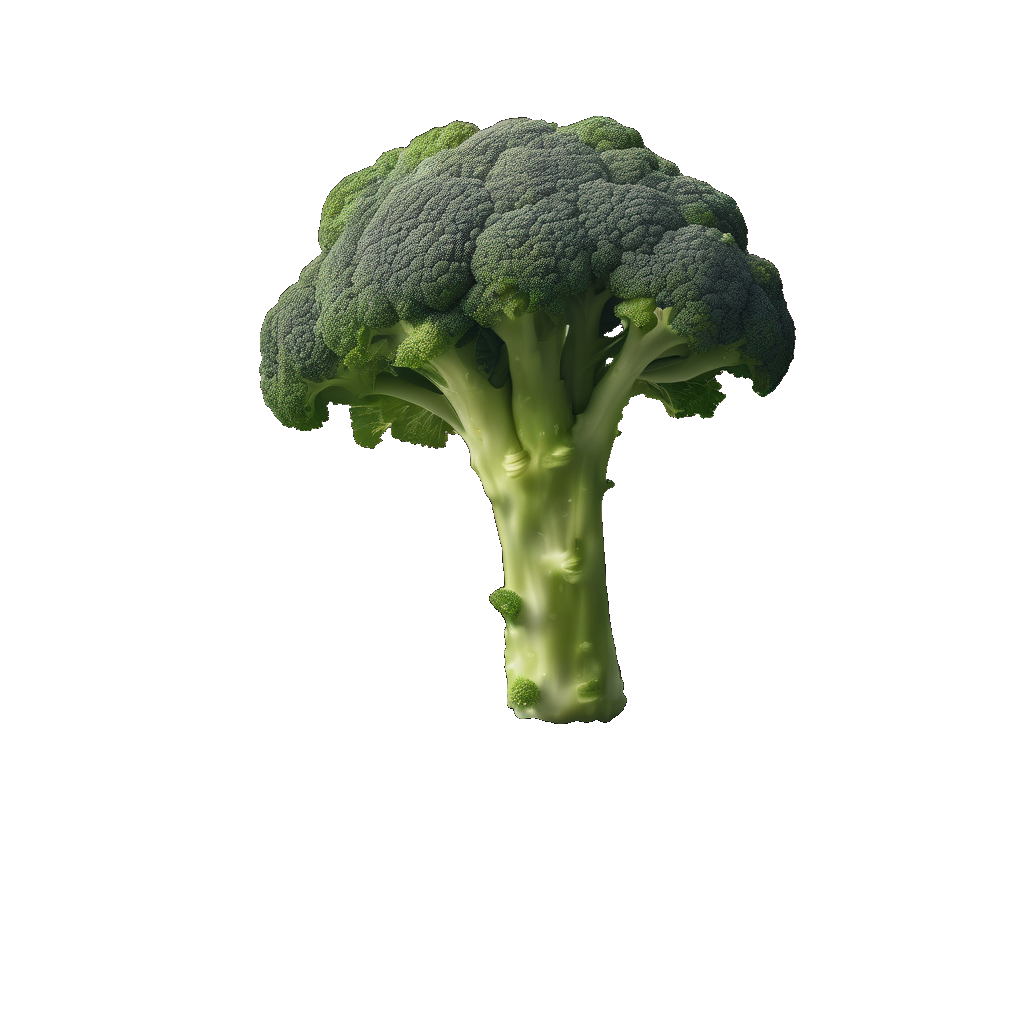} &
        \includegraphics[width=0.1\linewidth]{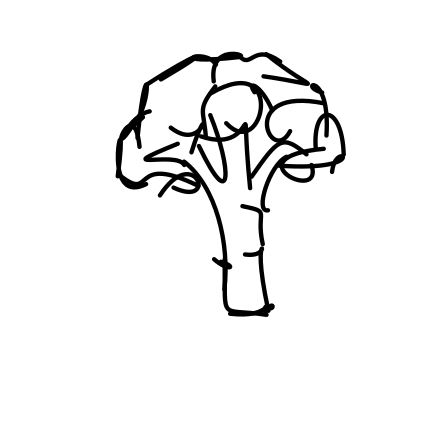} &
       \includegraphics[width=0.1\linewidth]{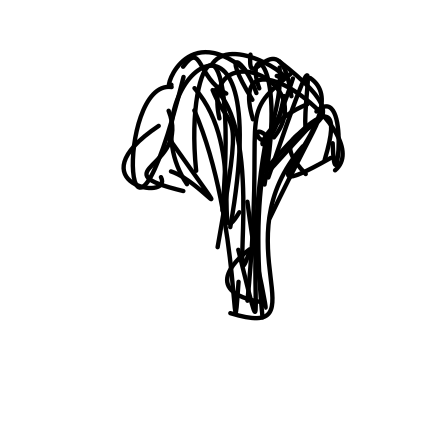} \\
        Input & ControlSk. &  CLIPasso  & Input  & ControlSk. &  CLIPasso & Input  & ControlSk. &  CLIPasso  \\

         \includegraphics[width=0.1\linewidth]{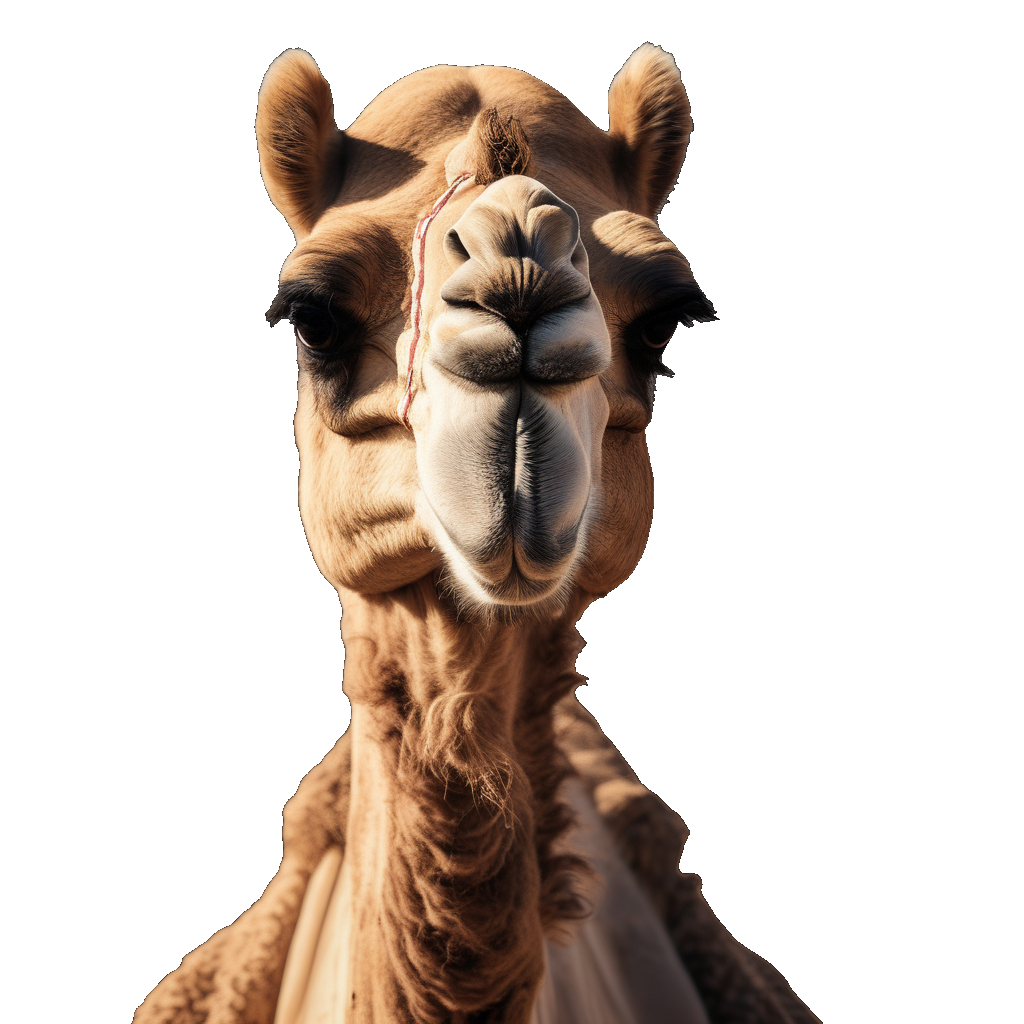} &
        \includegraphics[width=0.1\linewidth]{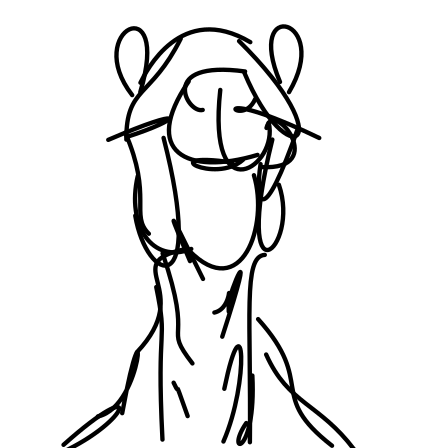} &
       \includegraphics[width=0.1\linewidth]{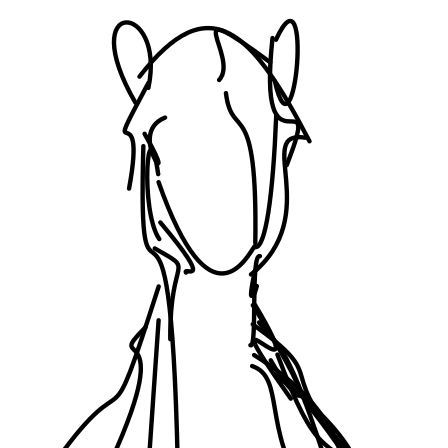}  &
        \includegraphics[width=0.1\linewidth]{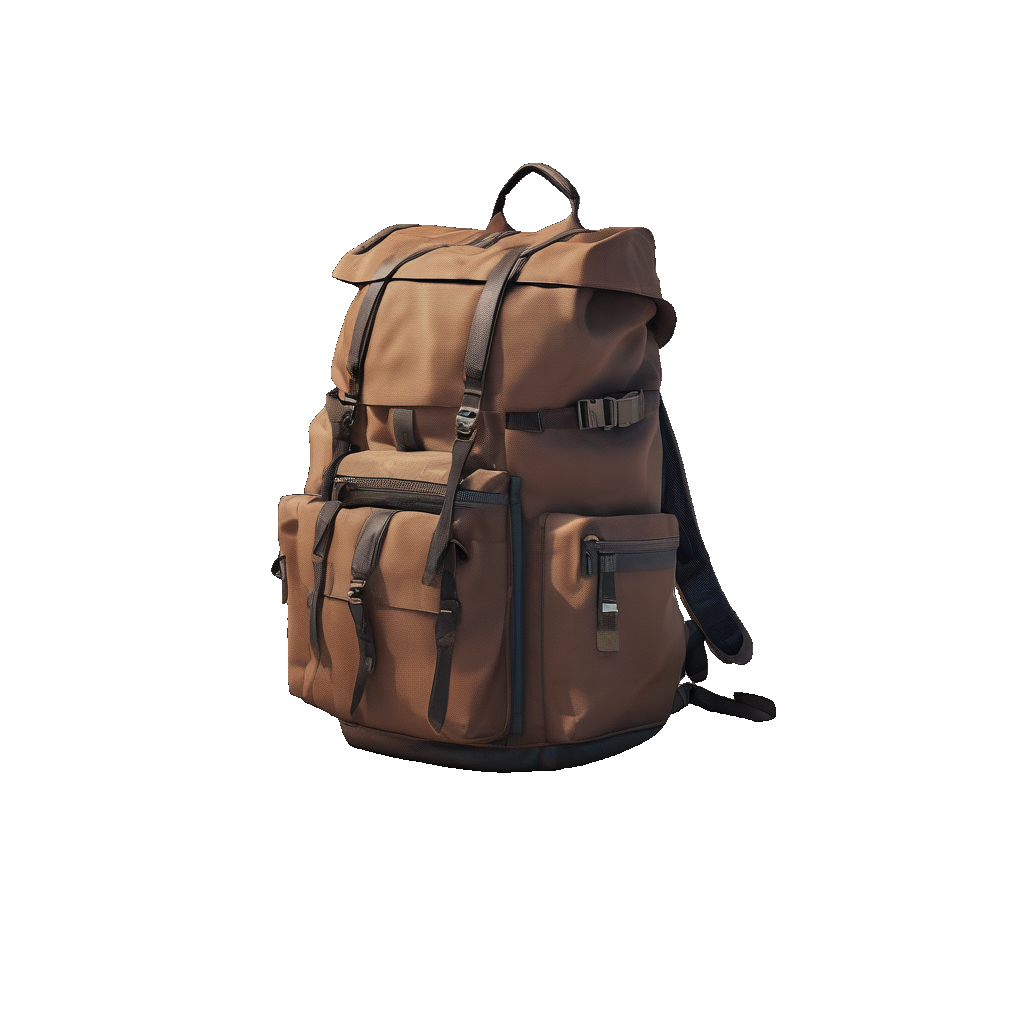} &
        \includegraphics[width=0.1\linewidth]{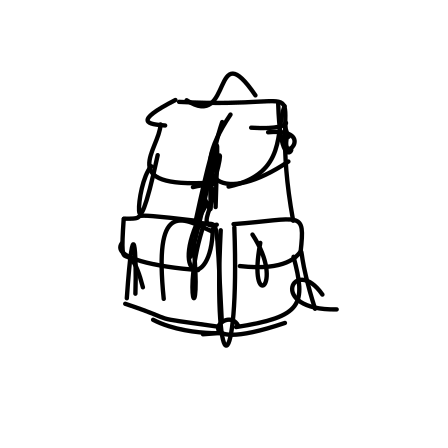} &
       \includegraphics[width=0.1\linewidth]{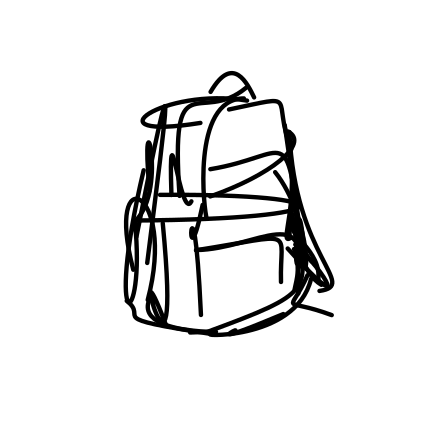}&  
       \includegraphics[width=0.1\linewidth]{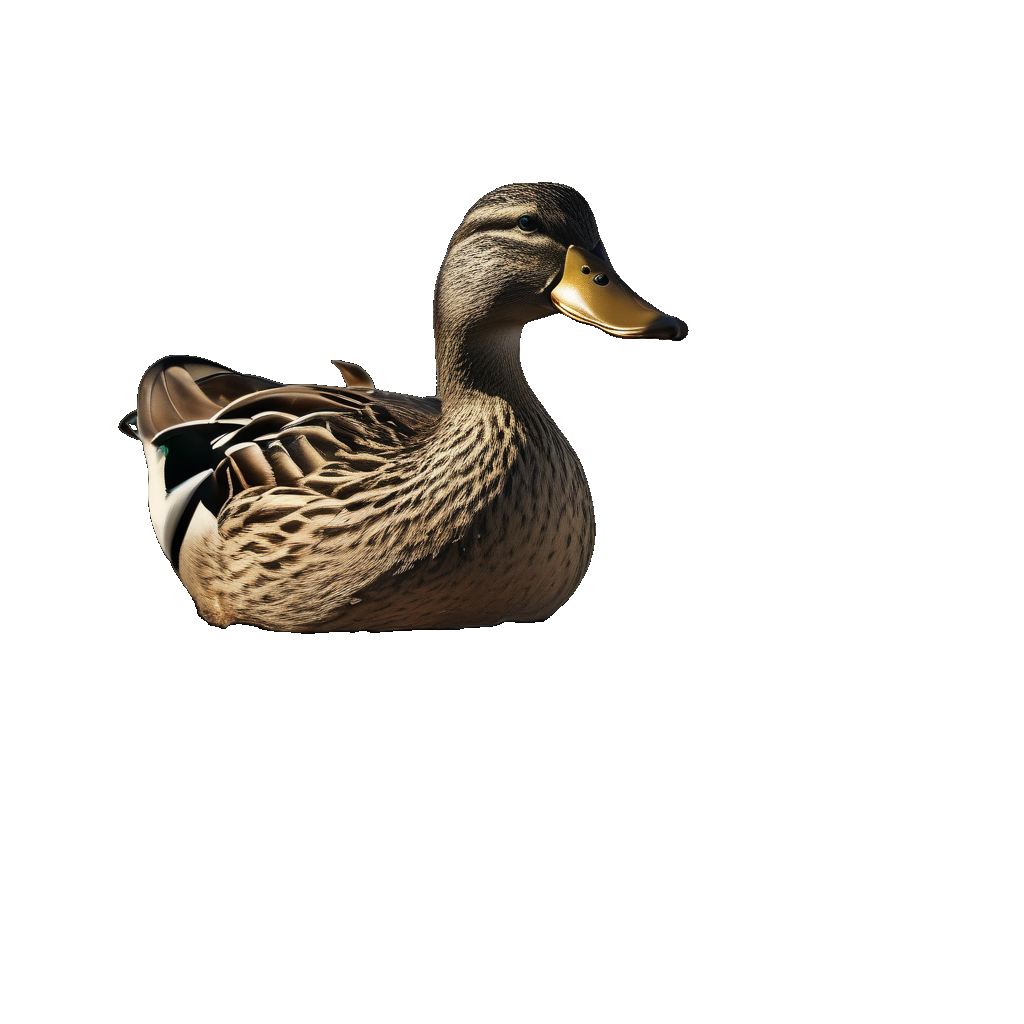} &
        \includegraphics[width=0.1\linewidth]{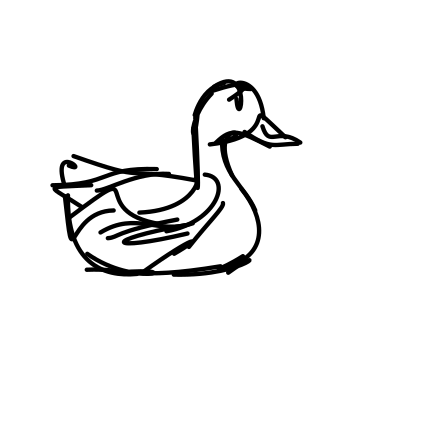} &
       \includegraphics[width=0.1\linewidth]{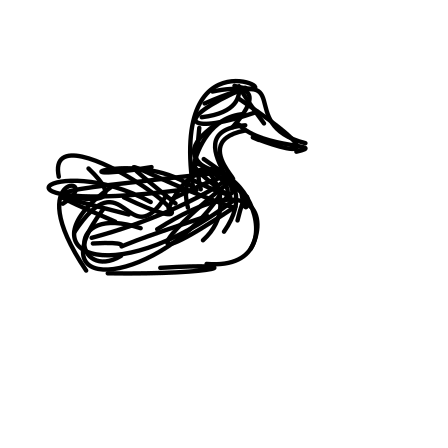} \\
        Input & ControlSk. &  CLIPasso  & Input  & ControlSk. &  CLIPasso & Input  & ControlSk. &  CLIPasso  \\

        \includegraphics[height=0.09\linewidth]{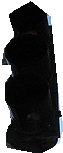} &
        \includegraphics[width=0.1\linewidth]{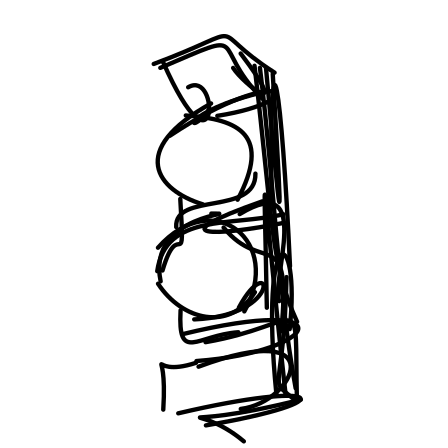} &
       \includegraphics[width=0.1\linewidth]{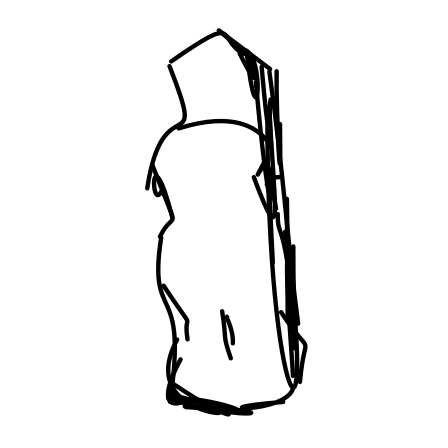}  &
       \raisebox{0.23\height}{\includegraphics[width=0.1\linewidth]{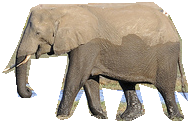}} &
        \includegraphics[width=0.1\linewidth]{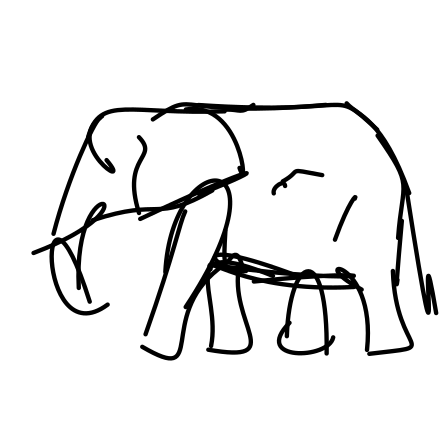} &
       \includegraphics[width=0.1\linewidth]{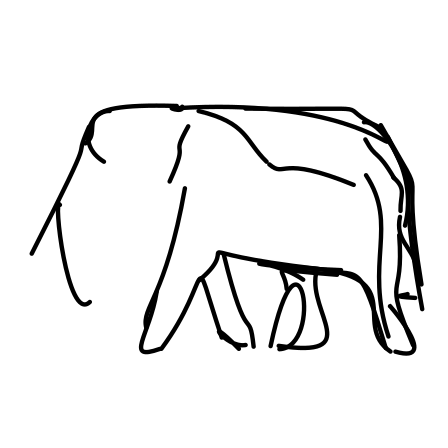} &
        \raisebox{0.15\height}{\includegraphics[height=0.08\linewidth]{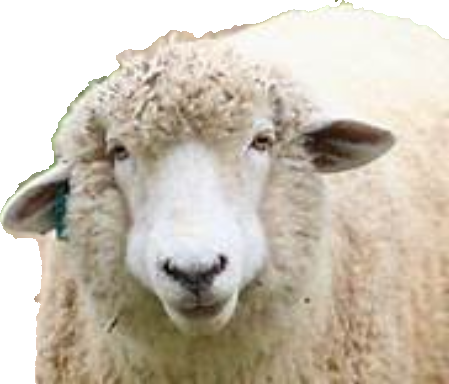}} &
        \includegraphics[width=0.1\linewidth]{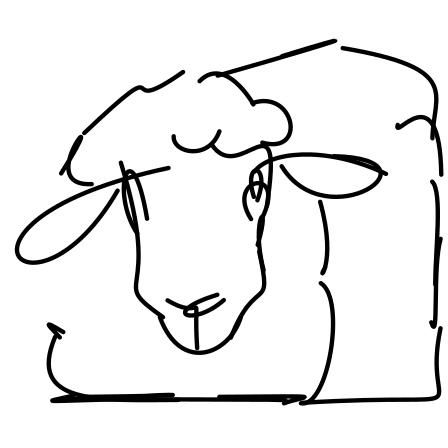} &
       \includegraphics[width=0.1\linewidth]{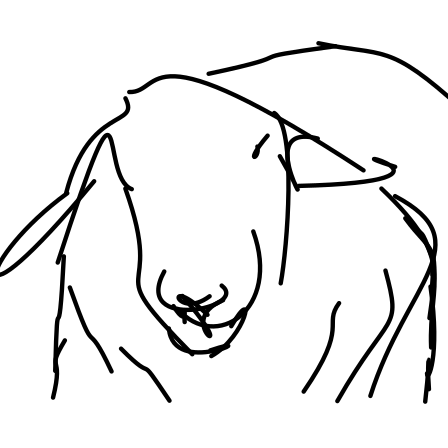}  \\

    \end{tabular}
    }
    \vspace{-0.3cm}
    \caption{Comparison of ControlSketch with CLIPasso \cite{vinker2022clipasso}. ControlSketch captures fine details (e.g., camel and bear), avoids artifacts in small objects (e.g., dog and duck), and handles challenging inputs effectively (last row).}
    \label{fig:comparison_opt}
\end{figure*}

\begin{figure*}
    \centering
    \includegraphics[width=0.95\linewidth]{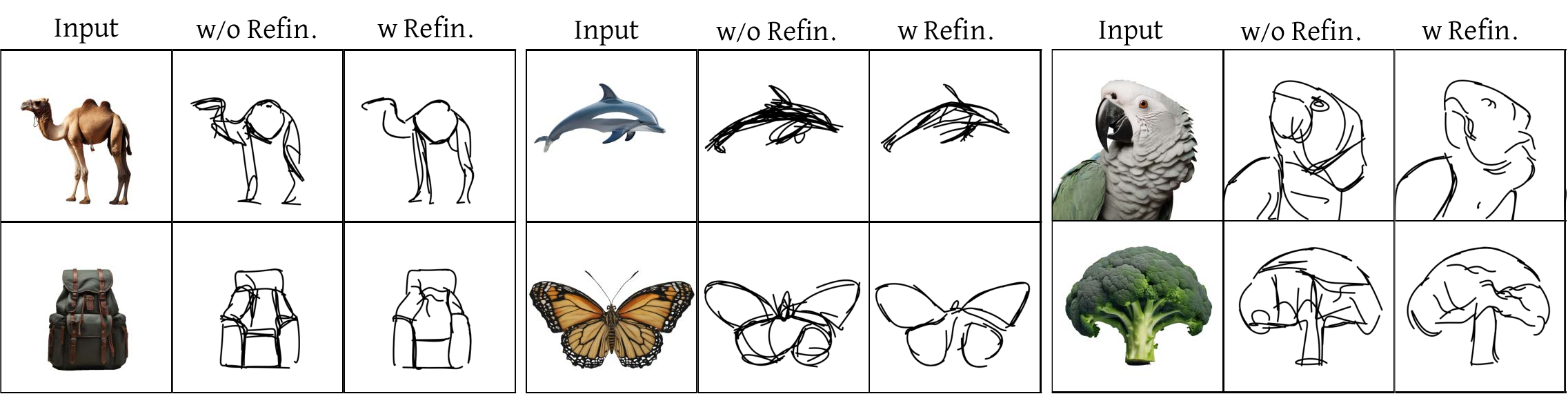}
    \vspace{-0.2cm}
    \caption{Effect of the refinement network. The output sketches from the diffusion model may contain slight noise, which the refinement network addresses by performing an additional cleaning step. However, this process can sometimes reduce details in the sketch (see Limitations).}
    \label{fig:refine_ablation}
\end{figure*}

{
    \small
    \bibliographystyle{ieee_fullname}
    \bibliography{main}
}

\clearpage
\appendix
\setcounter{page}{1}

\newpage
\twocolumn[
\centering
\Large
\textbf{\thetitle}\\
\vspace{2em}Supplementary Material \\
\vspace{1.0em}
] %

\part{}
\vspace{-20pt}
\parttoc

\vspace{-0.2cm}

\begin{figure}
    \centering
    \includegraphics[width=1\linewidth]{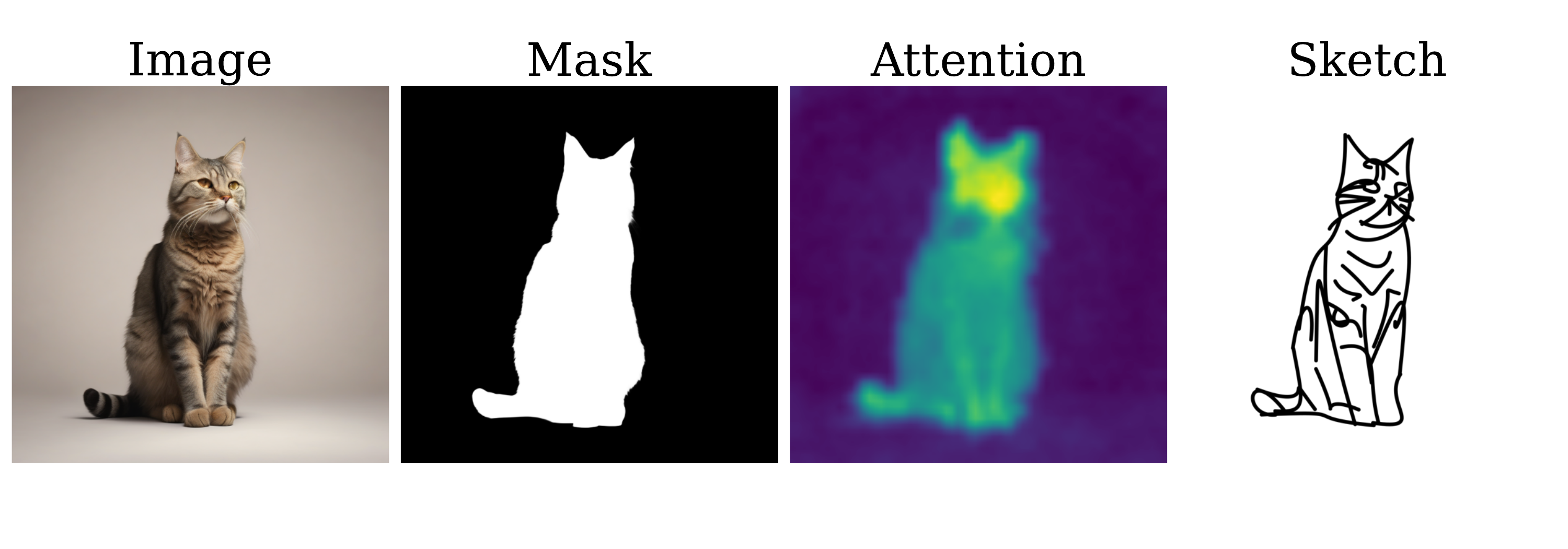}
    \caption{An example from the ControlSketch dataset, which includes the input image, object mask, attention map, and the corresponding sketch generated using ControlSketch.}
    \label{fig:data_sample}
\end{figure}

\section{ControlSketch Dataset}
The data generation process begins with generating images, followed by creating corresponding sketches using the ControlSketch framework, as described in section 4.2 in the main paper.
We generate images using the SDXL model \cite{podell2023sdxlimprovinglatentdiffusion} with the following prompt:
\textit{``A highly detailed wide-shot image of one $<c>$, set against a plain mesmerizing background. Center.''}, where $c$ is the class label.
Additionally, a negative prompt, \textit{``close up, few, multiple,''} is applied to ensure images depict a single object in a clear and high-quality pose. The generated images are of size $1024 \times 1024$. An example output image for the class ``cat'' is shown in \Cref{fig:data_sample}.

During the image generation process, we retain cross attention maps of the class label token extracted from internal layers of the model for future use. To isolate the object, we employ the BRIA Background Removal v1.4 Model  \cite{briaRMBG} to extract an object mask. After generating the image, we use BLIP2 \cite{li2023blip2bootstrappinglanguageimagepretraining} to extract the image caption that provides context beyond the object’s class. For example, for the image in \cref{fig:data_sample}, the caption  describe the cat as sitting, offering richer semantic information.
The sketches are generated using the ControlSketch method with 32 strokes. These strokes are subsequently arranged according to our stroke-sorting schema. The final SVG files contains the sorted strokes.
We use the Hugging Face implementation of SDXL version 1.0 \cite{podell2023sdxlimprovinglatentdiffusion} with its default parameters. Generating a single image with SDXL takes approximately 10 seconds, while sketch generation using the ControlSketch method on an NVIDIA RTX3090 GPU requires about 10 minutes.

Our dataset comprises 35,000 pairs of images and their corresponding sketches in SVG format, spanning 100 object categories. These categories are derived by combining common ones from existing sketch datasets {\cite{Mukherjee2023SEVALS,SketchRNN,SketchyCOCO2020,Eitz2012HowDH} with additional, unique categories such as astronaut, robot and sculpture. These unique categories are not present in prior datasets, highlighting the advantages of a synthetic data approach. The full list of categories is available in Table~\ref{tab:ControlSketch_dataset}.
All the sketches in our data were manually verified, we filtered very few generated images with artifacts that caused artifacts in the generated sketches.
The 15 categories used in training are: angel, bear, car, chair, crab, fish, rabbit, sculpture, astronaut, bicycle, cat, dog, horse, robot, woman. For each of these categories we generated 1200 image-sketch pairs, where 1000 samples are used for training and the rest for testing.
For the rest of 85 categories we created 200 samples per class. 
We show 78 random samples from each class of the training data in \Cref{fig:dog,fig:angle,fig:astronaut,fig:bicycle,fig:chair,fig:bear,fig:car,fig:cat,fig:horse,fig:chair,fig:crab,fig:fish,fig:rabbit,fig:Sculpture,fig:robot,fig:woman}, and 100 random samples from the entire dataset (one of each class) in \Cref{fig:100ControlSketch}.
Since the entire data creation pipeline is fully automated, we continuously extend the dataset and plan to release the code to enable future work in this area.

\begin{table}[h!]
\centering
\setlength{\tabcolsep}{2pt}
\begin{tabular}{|c|c|c|c|c|}
\midrule
\small
airplane & alarm clock & angel & astronaut & backpack \\ \midrule
bear & bed & bee & beer & bicycle \\ \midrule
boat & broccoli & burger & bus & butterfly \\ \midrule
cabin & cake & camel & camera & candle \\ \midrule
car & carrot & castle & cat & cell phone \\ \midrule
chair & chicken & child & cow & crab \\ \midrule
cup & deer & doctor & dog & dolphin \\ \midrule
dragon & drill & duck & elephant & fish \\ \midrule
flamingo & floor lamp & flower & fork & giraffe \\ \midrule
goat & hammer & hat & helicopter & horse \\ \midrule
house & ice cream & jacket & kangaroo & kimono \\ \midrule
laptop & lion & lobster & margarita & mermaid \\ \midrule
motorcycle & mountain & octopus & parrot & pen \\ \midrule
\begin{tabular}[c]{@{}c@{}}pickup \\ truck \end{tabular}  & pig & purse & quiche & rabbit \\ \midrule
robot & sandwich & scissors & sculpture & shark \\ \midrule
sheep & spider & squirrel & strawberry & sword \\ \midrule
t-shirt & table & teapot & television & tiger \\ \midrule
tomato & train & tree & truck & umbrella \\ \midrule
vase & waffle & watch & whale & wine bottle \\ \midrule
woman & yoga & zebra & \begin{tabular}[c]{@{}c@{}}The Eiffel \\ Tower \end{tabular} & book \\ \midrule
\end{tabular}
\caption{The 100 categories of the ControlSketch dataset.}
\label{tab:ControlSketch_dataset}
\end{table}

\begin{figure*}
    \centering
    \includegraphics[width=1\linewidth]{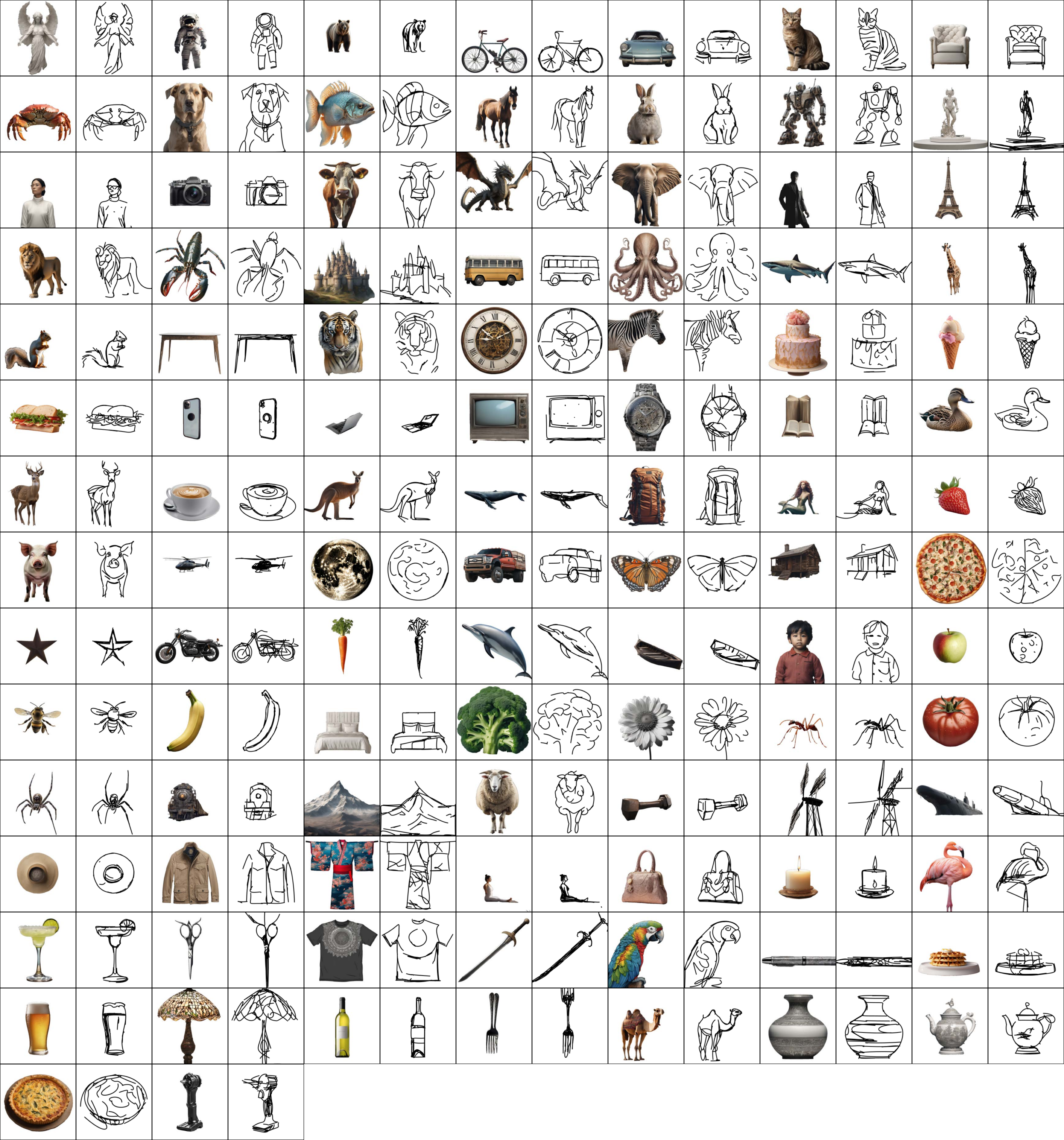}
    \caption{100 random samples of sketches generated with ControlSketch.}
    \label{fig:100ControlSketch}
\end{figure*}

\section{ControlSketch Method}
\paragraph{Technical details}
In the ControlSketch optimization, we leverage the pretrained depth ControlNet model \cite{controlnet2023} to compute the SDS loss. The Adam optimizer is employed with a learning rate of 0.8. The optimization process runs for 2000 iterations, taking approximately 10 minutes to generate a single sketch on an RTX 3090 GPU.  However, after 700 iterations most images already yield a clearly identifiable sketch.

\paragraph{Strokes initialization}
The number of areas, $k$, is defined as the rounded square root of the total number of strokes $n$ (for our default number of strokes, 32, $k$ is set to 6). 
Our initialization technique combines between saliency and full coverage of the sketch, which we find to be important when the SDS loss is applied with our spatial control. In \Cref{fig:initalization_vis} we demonstrate how the final sketches will look like when applied with and without our enhances initialization, where the default case is defined based on the attention map as was proposed in CLIPasso \cite{vinker2022clipasso}. As seen, our approach ensures comprehensive object coverage while emphasizing critical areas, resulting in visually effective and recognizable sketches without omitting essential elements. For example, in the lion image, initializing strokes based solely on saliency results in almost all strokes focusing on the lion's head. Consequently, the final sketch omits significant portions of the lion's body.

\paragraph{Spatial control}
The ControlNet model receives two inputs as conditions: the text prompt and the depth condition. The balance between these conditions which is determined  by the conditioning scale parameter influences the final sketch attributes. We found that a conditioning scale of 1.5 provides the best results, effectively maintaining both semantic and geometric attributes of the subject. 

The depth ControlNet model used in the control SDS loss can be replaced with any other ControlNet model, along with the extraction of the appropriate condition from the input image. Different ControlNet models influence the style and attributes of the final sketch. Examples of different sketches generated with different ControlNet models and conditions are shown in Figure~\ref{fig:deifferent_conditions}.

\begin{figure}[t]
    \centering
    \setlength{\tabcolsep}{0pt}
    {\small
        \begin{tabular}{c |c c c}
         \midrule
        Input &  Depth & Scribble & Segmentation  \\
         \midrule
        \includegraphics[width=0.25\linewidth]{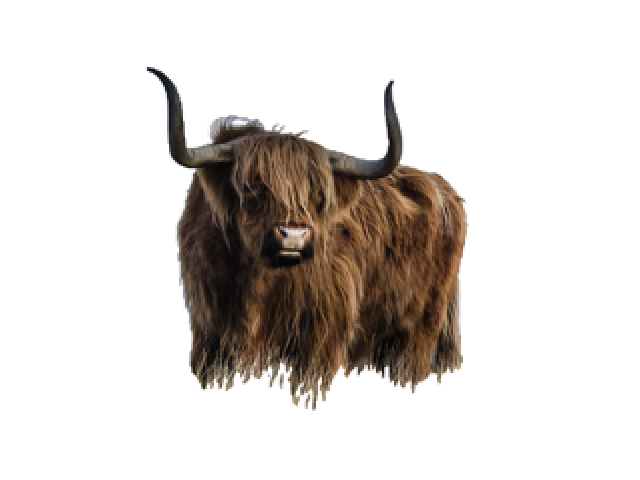} &
        \includegraphics[width=0.25\linewidth]{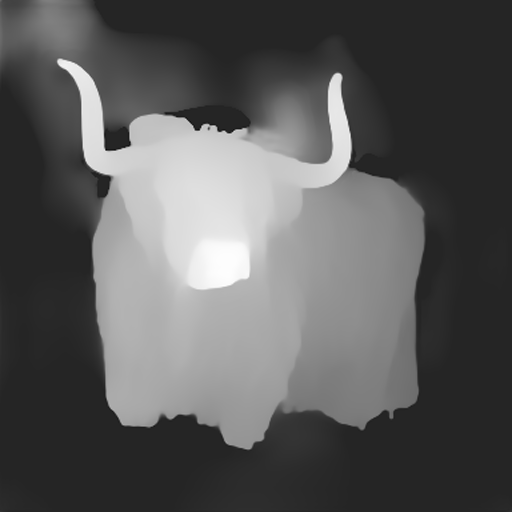} & 
        \includegraphics[width=0.25\linewidth]{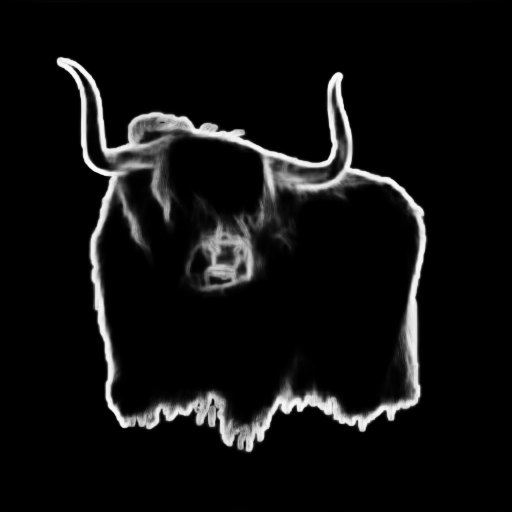} & 
        \includegraphics[width=0.25\linewidth]{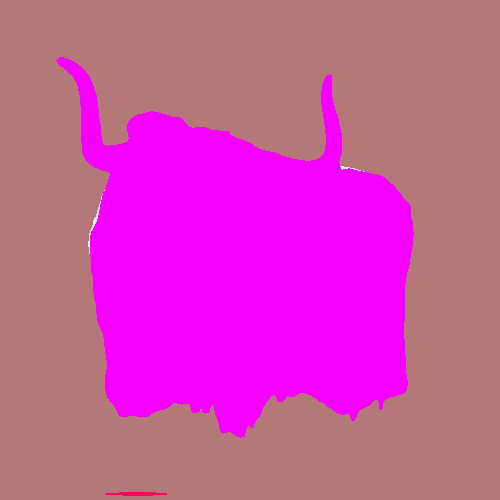} \\

        &
        \includegraphics[width=0.25\linewidth]{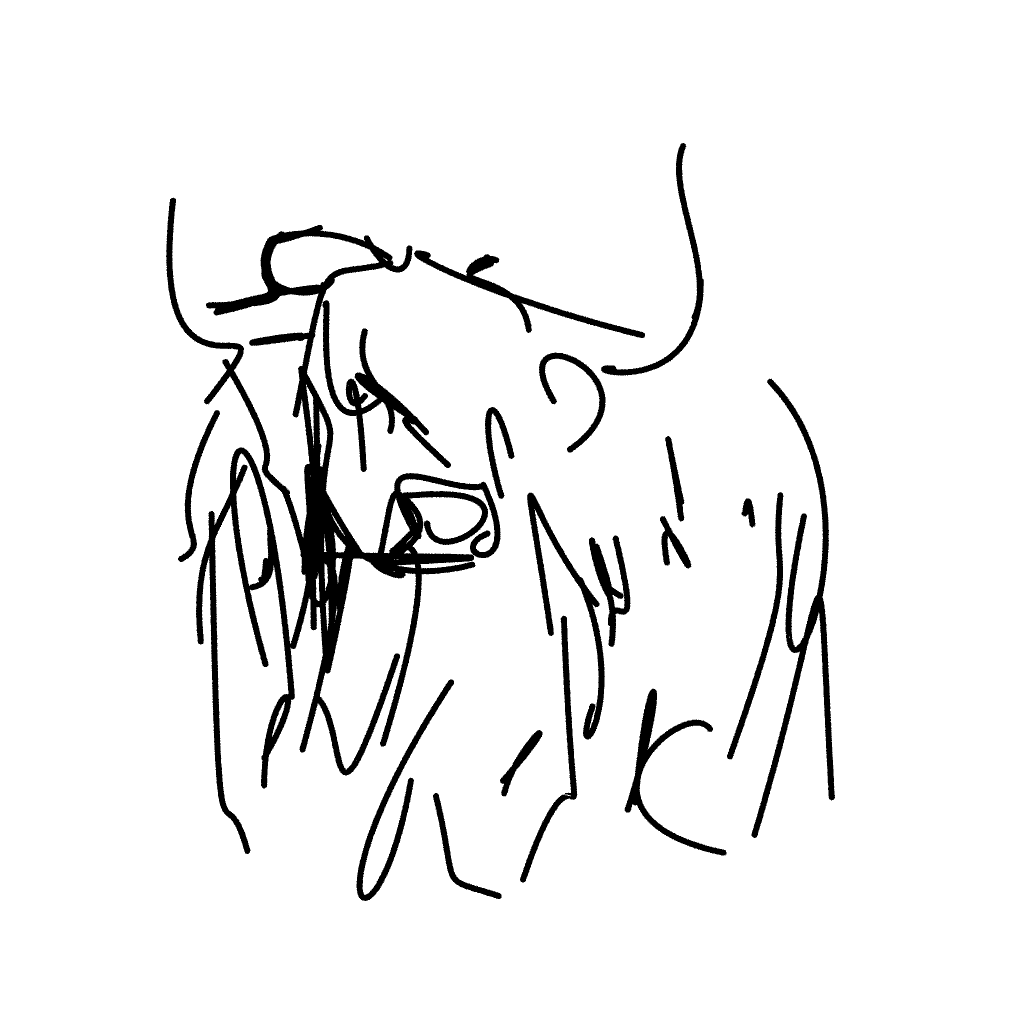} &
        \includegraphics[width=0.25\linewidth]{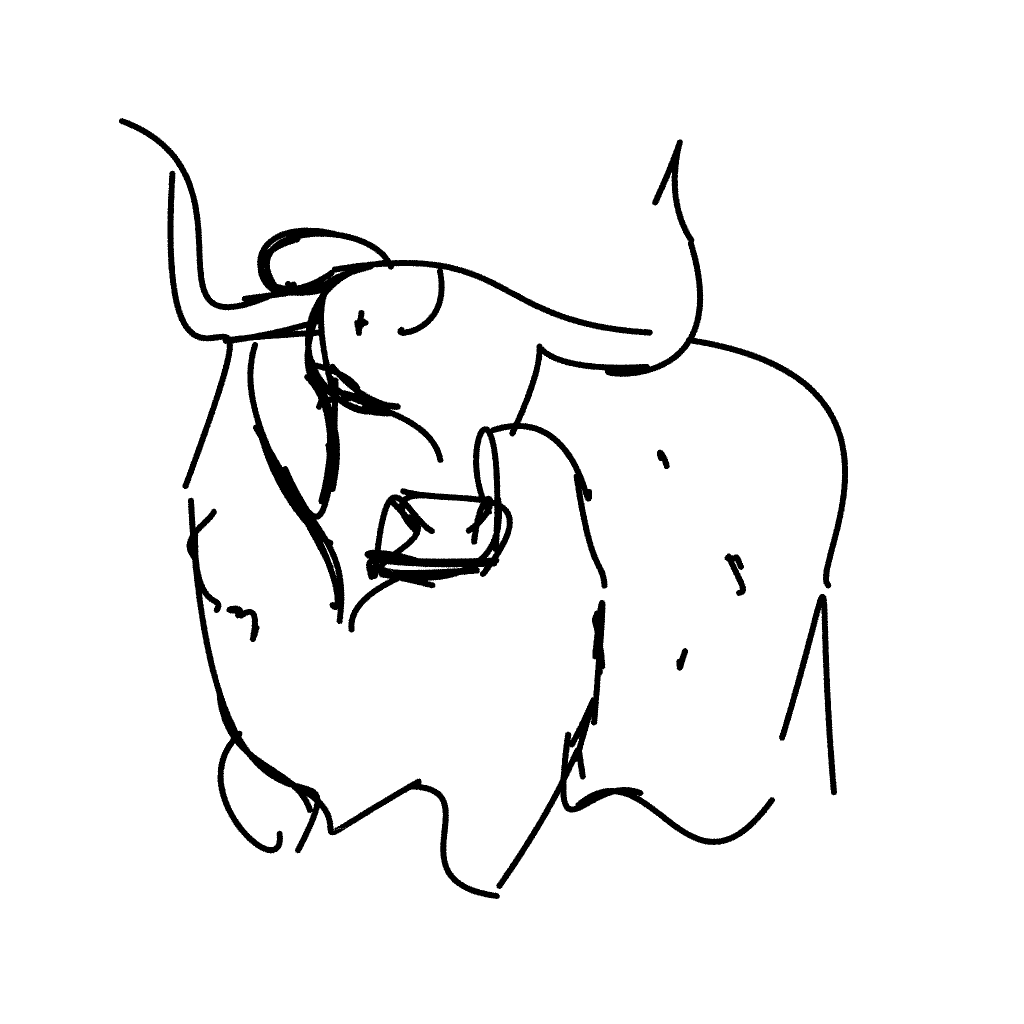} & \includegraphics[width=0.25\linewidth]{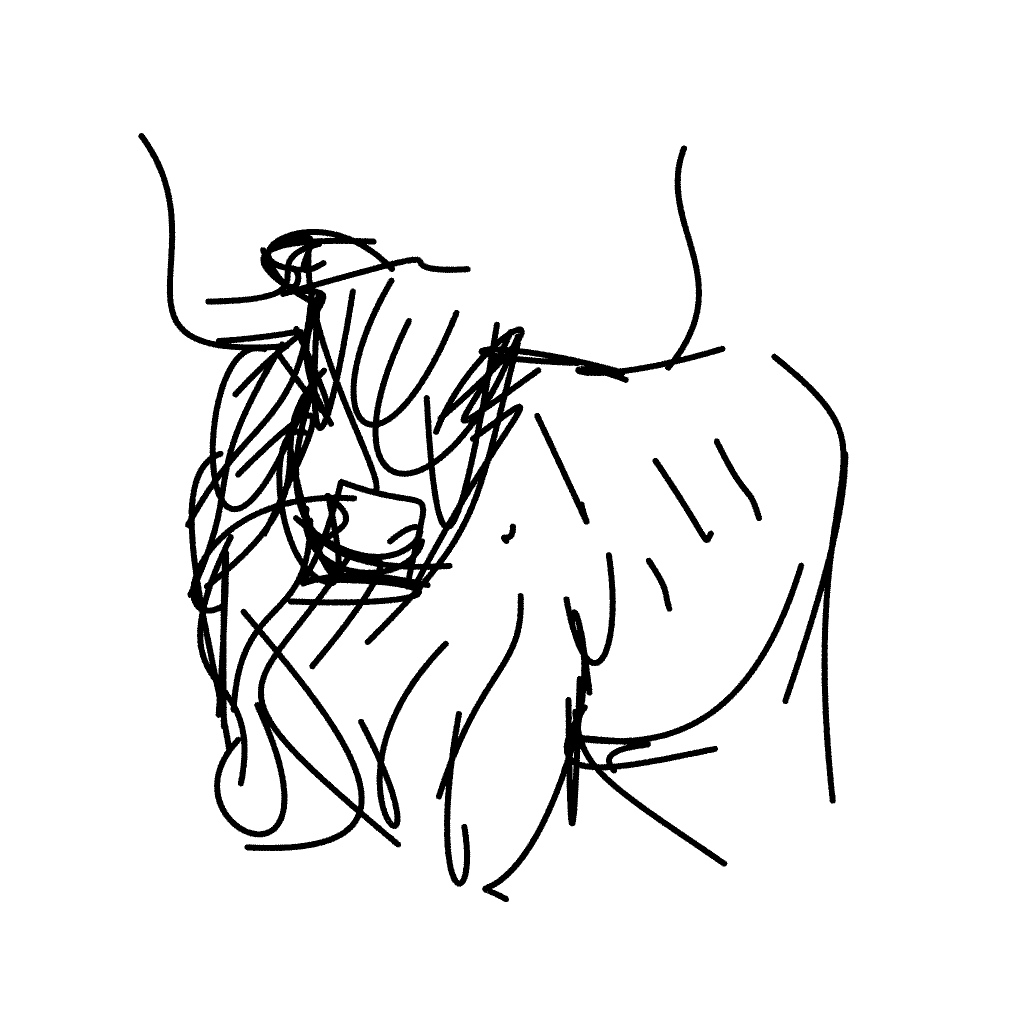} \\

         \midrule
        Input &  Depth & Scribble & Segmentation  \\
         \midrule

         \includegraphics[width=0.25\linewidth]{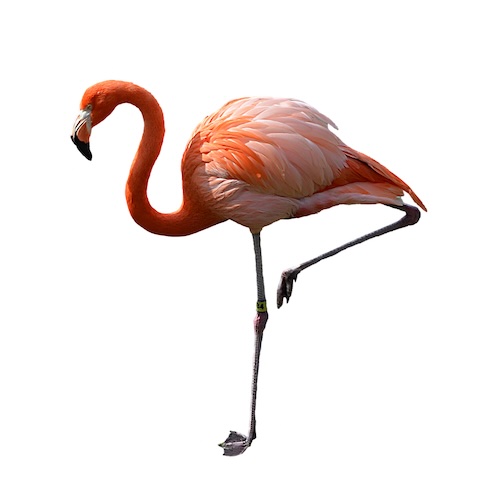} &
        \includegraphics[width=0.25\linewidth]{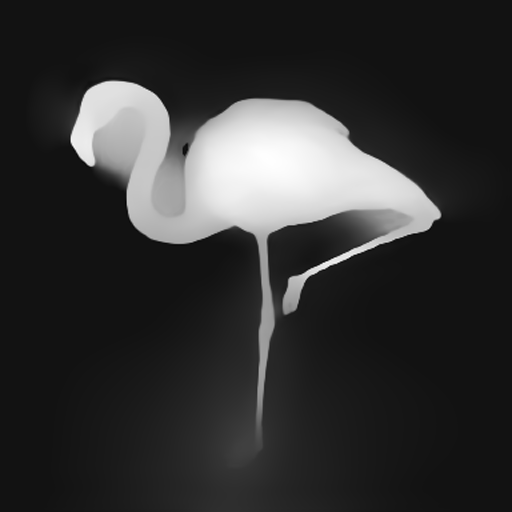} &
        \includegraphics[width=0.25\linewidth]{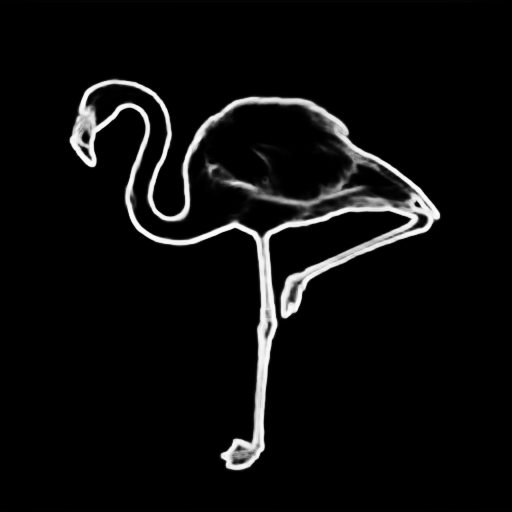}  & 
        \includegraphics[width=0.25\linewidth]{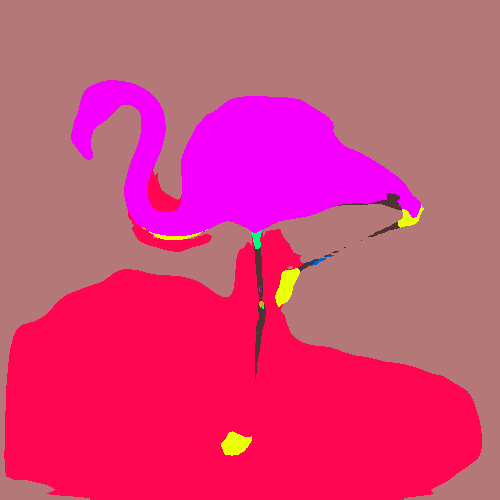} \\

         &
        \includegraphics[width=0.25\linewidth]{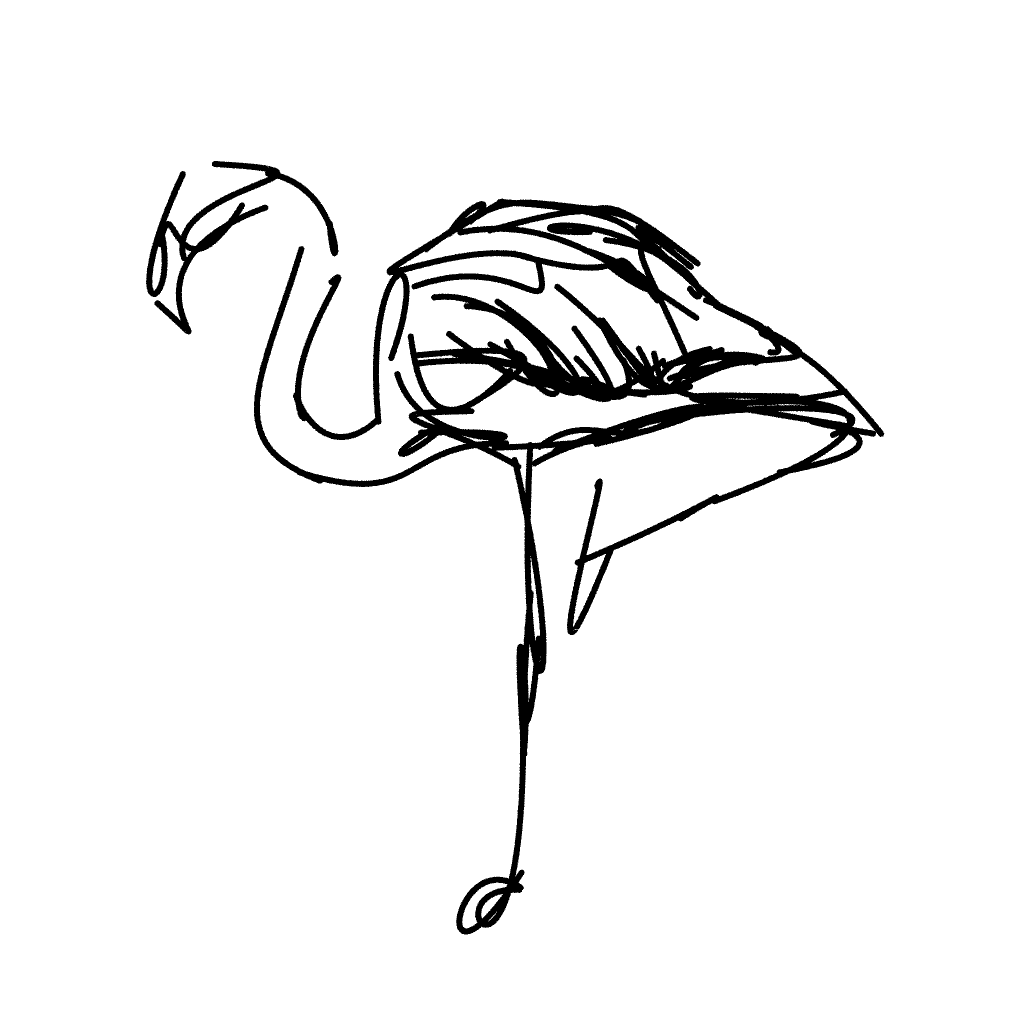} &
        \includegraphics[width=0.25\linewidth]{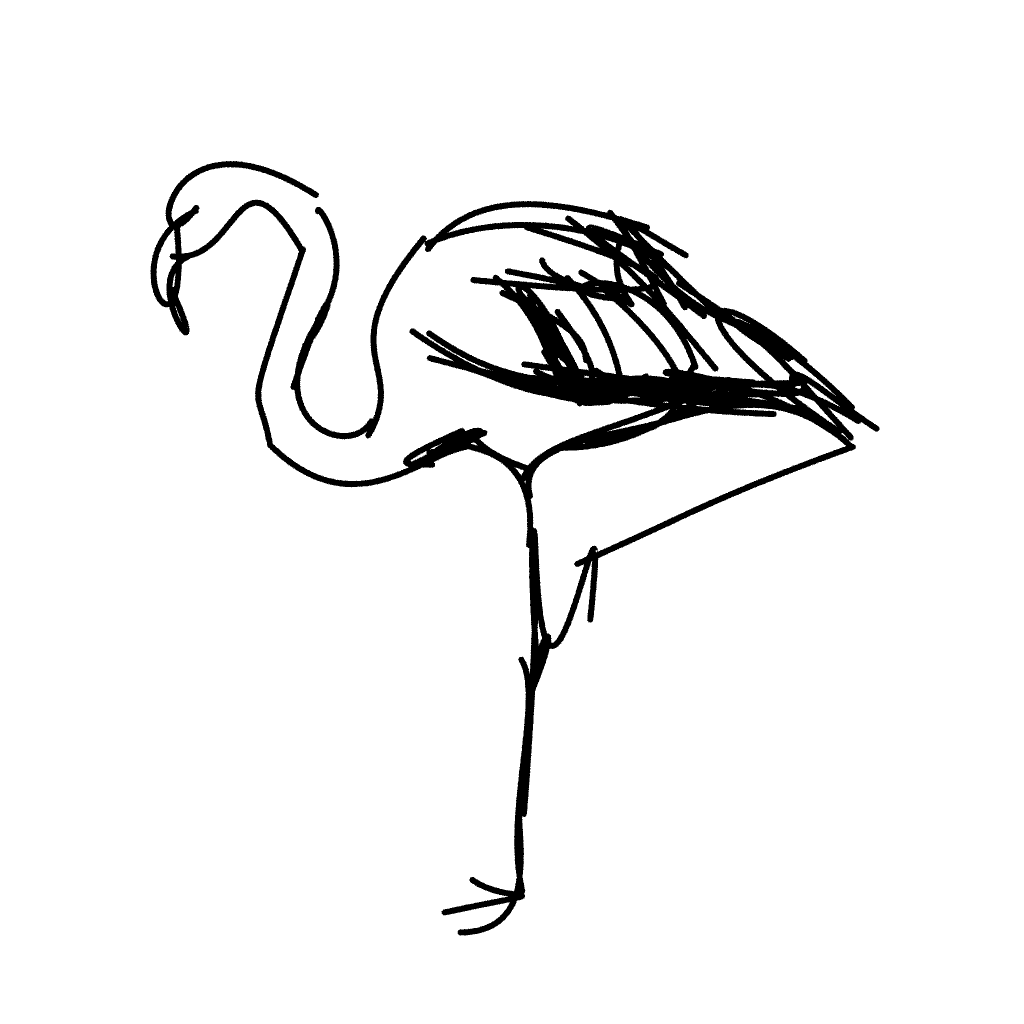}  & 
        \includegraphics[width=0.25\linewidth]{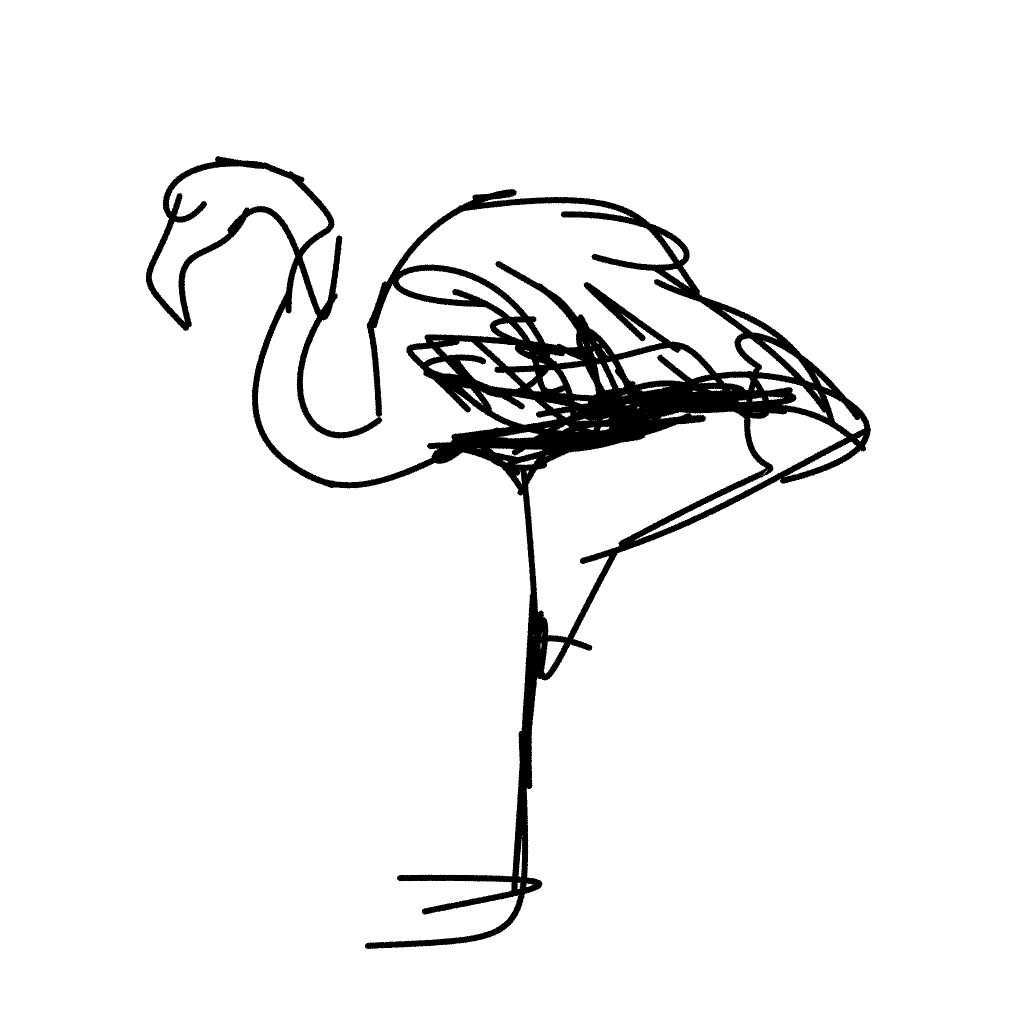} \\
       
    \end{tabular}
    }
    \caption{Examples of sketches generated by ControlSketch using different ControlNet models, alongside their respective conditions which influence the style and attributes of the final sketches.}
    \label{fig:deifferent_conditions}
\end{figure}

\begin{figure}
    \centering
    \includegraphics[width=0.6\linewidth]{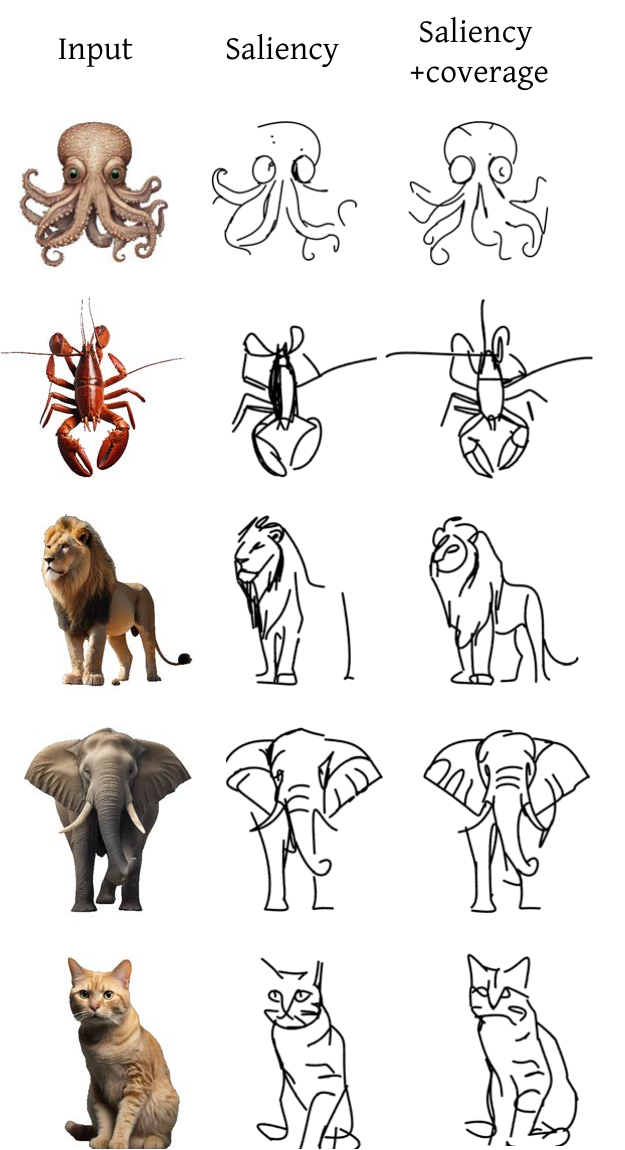}
    \caption{ Strokes initialization in the ControlSketch method. The "Saliency" column demonstrates the result when strokes are initialized based solely on the attention map (following common practive \cite{vinker2022clipasso}), often leading to an overemphasis on critical regions, such as the lion's head, at the expense of other important parts like the body. The "Saliency + coverage" column showcases our enhanced initialization method, which combines saliency with full object coverage, ensuring both essential details and global object representation are maintained, resulting in complete and recognizable sketches.}
    \label{fig:initalization_vis}
\end{figure}

\section{SwiftSketch}
Our implementation is built on the MDM codebase \cite{tevet2023human}.
Our model consists of 8 self- and cross-attention layers. It was trained with a batch size of 32, a learning rate of $5 \times 10^{-5}$, for 400,000 steps. The refinement network shares the same architecture as our diffusion model and is initialized with its final weights. The timestep condition is fixed at 0. We train the refinement network on the diffusion output sketches from the training dataset, using only the LPIPS loss between the network’s rendered output sketch and the target rendered sketch, as we found it resulting in more polished and visually improved final sketches. The refinement network was trained for 30,000 steps with a learning rate of $5 \times 10^{-6}$.

For training, We scaled the ground truth $(x, y)$ coordinates to the range [-2, 2]. Our experiments revealed that a scaling factor of 2 outperformed the standard value of 1.0 which is used in image generation tasks.
To extract input image features for our model, the image is processed using a pretrained CLIP ResNet model \cite{Radfordclip}, with features extracted from its fourth layer. These features are subsequently refined through three convolutional layers to capture additional spatial details. Each patch embedding is further refined using three linear layers, enhancing feature learning and aligning dimensions for compatibility with the model. The resulting feature representation is seamlessly integrated into the generation process via a cross-attention mechanism.

To encourage the diffusion model to focus on fine details, we adjust the noise scheduler to perturb the signal more subtly for small timesteps, by reducing the exponent in the standard cosine noise schedule proposed in \cite{Nichol2021ImprovedDD} from 2 to 0.4.
Our model $M_{\theta}$ was trained using classifier-free guidance so during inference, we enhance fidelity to the input image by extrapolating the following variants using s= 2.5:
\begin{equation}
    M_{\theta_s}(s^t, t, I) = M_{\theta}(s^t, t, \emptyset) + s \cdot \big( M_{\theta}(s^t, t, I) - M_{\theta}(s^t, t, \emptyset) \big)
\end{equation}.

Figure~\ref{fig:swiftsketch_random} showcases 100 random SwiftSketch samples across all categories in the ControlSketch dataset. The last three rows correspond to classes our model was trained on, while the remaining rows are unseen classes. Each sketch is generated in under a second. The results demonstrate that our model generalizes well to unseen categories, producing sketches with high fidelity to the input images. However, in some cases, high-level details are absent, and the sketch's category label can be difficult to identify. More examples for unseen classes are shown in Figure~\ref{fig:swiftskwtch_unseen1}, Figure~\ref{fig:swiftskwtch_unseen2} and Figure~\ref{fig:swiftskwtch_unseen3}

\section{Qualitative Comparison}
Figure~\ref{fig:comparison_train} and Figure~\ref{fig:comparison_test}  show more examples of qualitative comparison of seen and unseen categories. Input images are shown on the left. From left to right, the sketches are generated using  PhotoSketching \cite{Li2019PhotoSketchingIC}, Chan et al. \cite{Chan2022LearningTG} (in anime style), InstantStyle \cite{Wang2024InstantStyleFL}, and CLIPasso \cite{vinker2022clipasso}. On the right are the resulting sketches from our proposed methods, ControlSketch and SwiftSketch.

\begin{figure}
    \centering
    \includegraphics[width=1\linewidth]{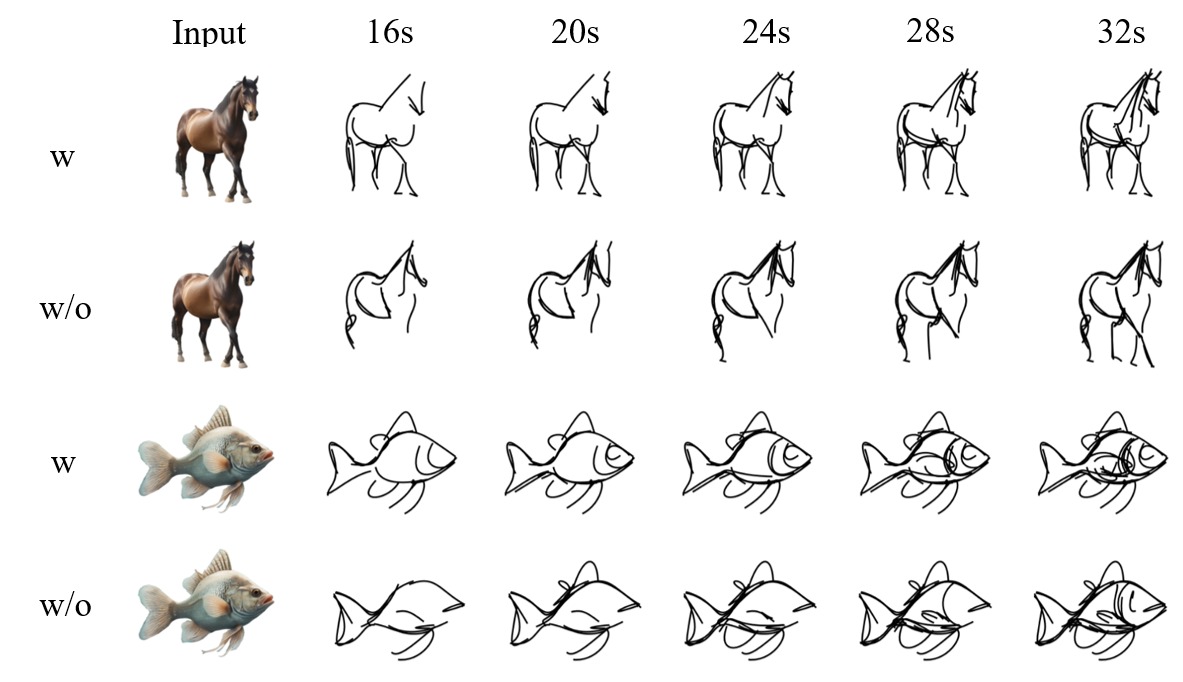}
    \caption{Stroke Order Visualization. SwiftSketch generated sketches are visualized progressively, with the stroke count shown on top. The first row for each example is with our sorting technique (w), while the second row omits it (w/o)}
    \label{fig:strokes_number_sort}
\end{figure}

\section{Quantitative Evaluation}
In this section, we present the details of the user study conducted to compare our new optimization method, ControlSketch, with the state-of-the-art optimization method for object sketching, CLIPasso.
We selected 24 distinct categories for the user study: 16 categories from our ControlSketch dataset, and 8 categories from the SketchyCOCO dataset. For each category, we randomly sampled one image. Participants were presented with the input image alongside two sketches—one generated by CLIPasso and the other by ControlSketch—displayed in random order. We asked participants two questions for each pair of sketches: 1. Which sketch more effectively depicts the input image? 2. Which sketch is of higher quality? Participants were required to choose one sketch for each question. A total of 40 individuals participated in the survey. The results are as follows: For the ControlSketch dataset, 87\% of participants chose ControlSketch for the first question, and 88\% for the second question. For the SketchyCOCO dataset—which is more challenging due to its low-resolution images and difficult lighting conditions—90\% chose ControlSketch for the first question, and 93\% for the second question.
These results highlight the significant advantages of ControlSketch over CLIPasso across diverse categories and datasets.

\section{Ablation}
Figure~\ref{fig:comparison_refine} presents a comparison of results with and without the refinement step in the SwiftSketch pipeline. As can be seen, the final output sketches generated by the denoising process of our diffusion model may still retain slight noise. Incorporating the refinement stage significantly enhances the quality and cleanliness of the sketches
Figure~\ref{fig:strokes_number_sort} illustrates the impact of the stroke sorting technique used for training. Early strokes effectively capture the object’s contour and key features, while later strokes refine the details. With sorting, the object is significantly more recognizable with fewer strokes compared to the case without sorting.

\begin{figure}
    \centering
    \includegraphics[width=1\linewidth]{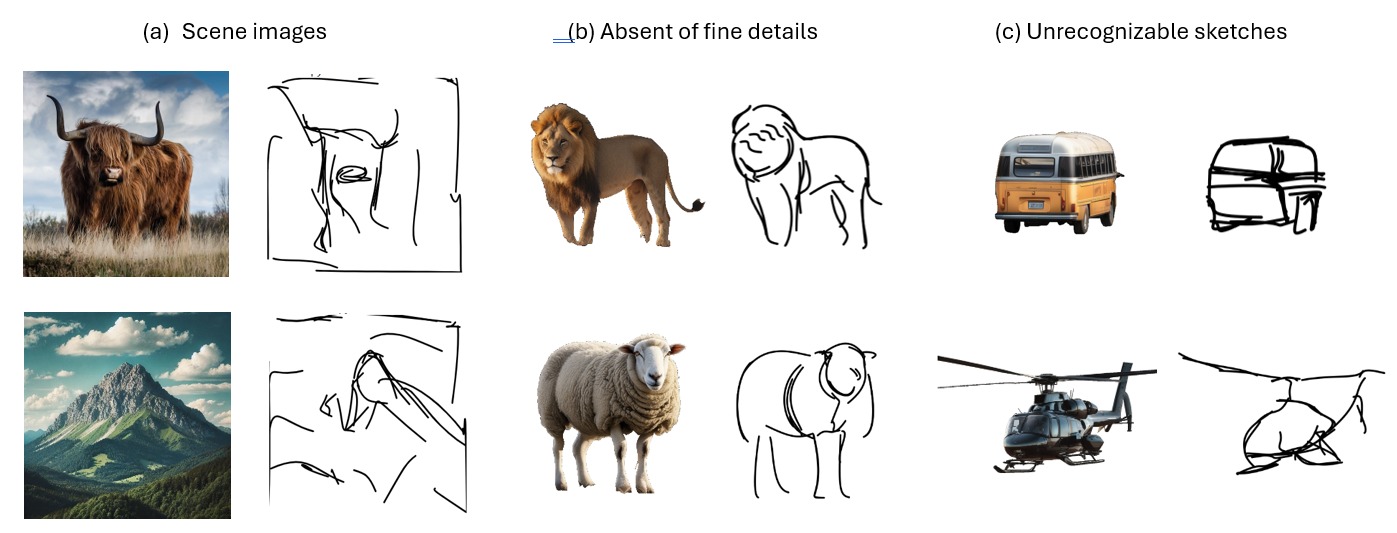}
    \caption{Limitations of SwiftSketch. (a) When trained solely on masked object images, SwiftSketch struggles to generate accurate sketches for complex scenes. As shown, it incorrectly assigns strokes to the image frame instead of capturing the scene's key elements. (b) During the refinement stage, fine details particularly facial features are often lost, resulting in oversimplified representations. (c) Sketches may appear unrecognizable. }
    \label{fig:limitations1}
\end{figure}

\section{Limitations}
SwiftSketch, which was trained only on masked object images, faces challenges in handling complex scenes. When provided with a scene image, as illustrated in Figure~\ref{fig:limitations1}(a), SwiftSketch struggles to generate accurate sketches, often misplacing strokes onto the image frame instead of capturing key elements of the scene. Another significant limitation is its tendency to omit fine details, particularly facial features, leading to oversimplified representations, as shown in Figure~\ref{fig:limitations1}(b). In some cases, sketches may appear unrecognizable, as shown in Figure~\ref{fig:limitations1}(c).

\begin{figure*}[t]
    \centering
    \setlength{\tabcolsep}{1pt}
    {\small
    \begin{tabular}{c}
         Seen Classes  \\
        \includegraphics[width=0.6\linewidth]{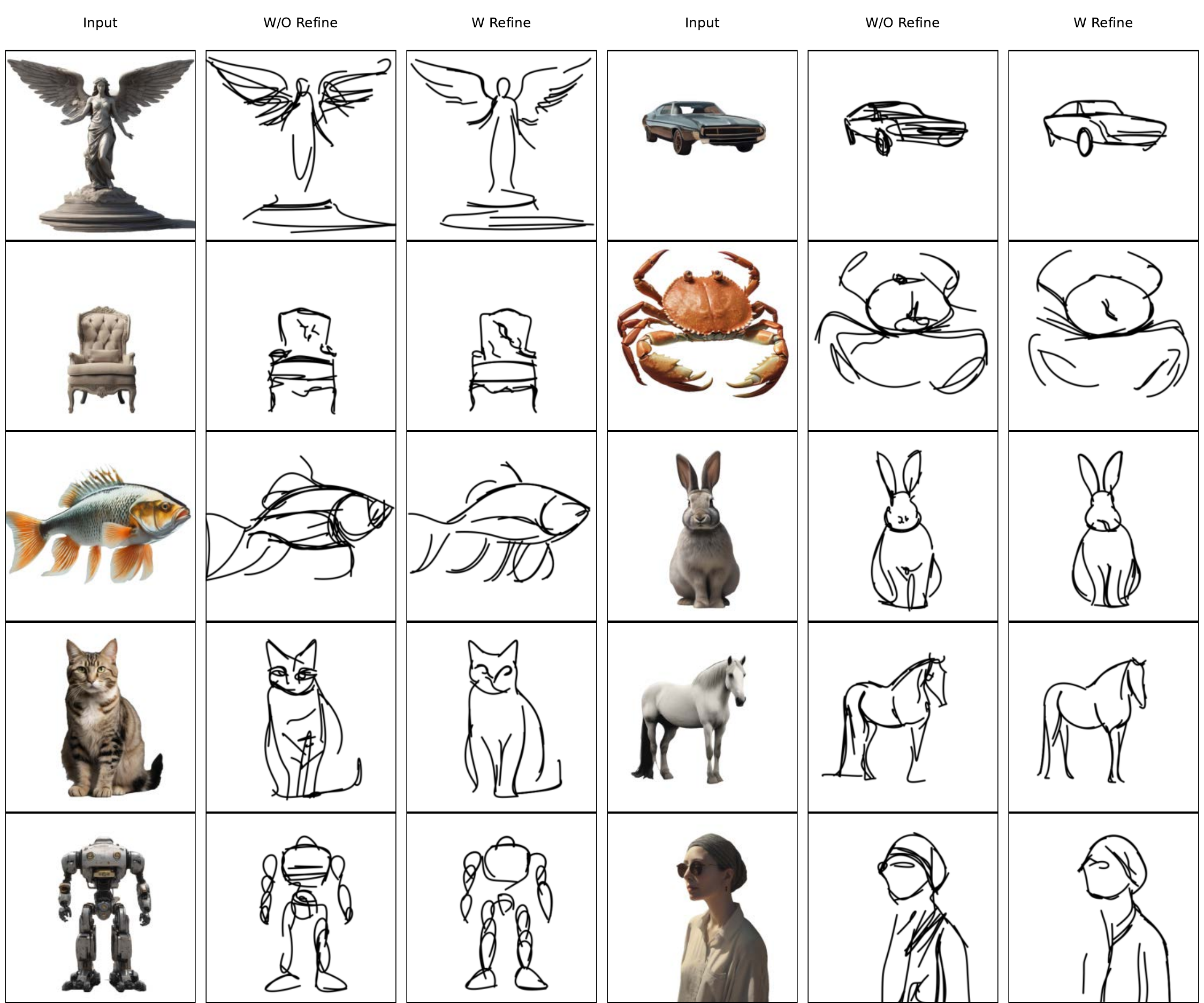}   \\
         Unseen Classes  \\
        \includegraphics[width=0.6\linewidth]{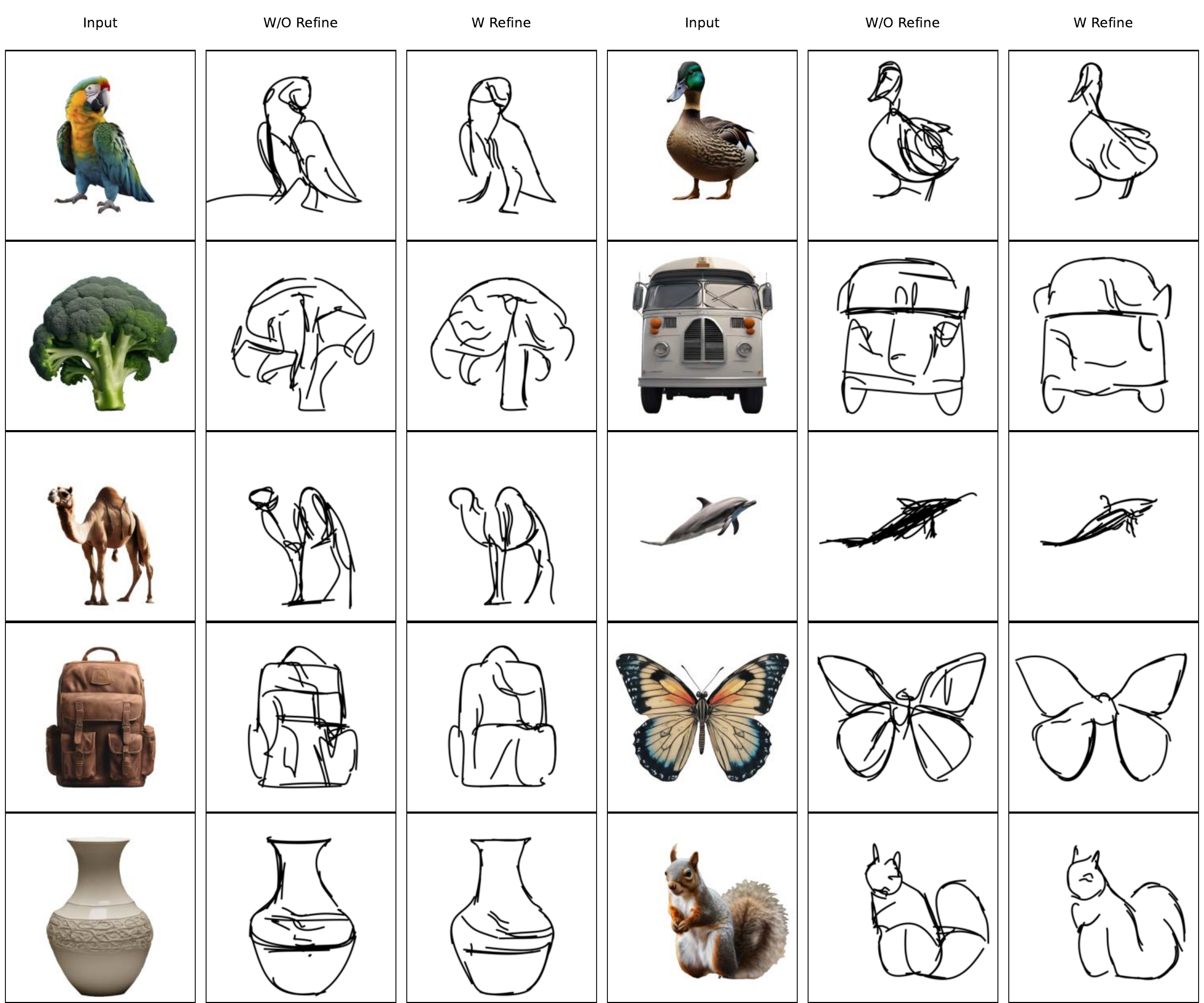}   \\
    
    \end{tabular}
    }
    \caption{Comparison of SwiftSketch sketches with (right) and without (left) the refinement step. This highlights the critical role of the refinement network in significantly improving the quality of the generated sketches and reducing noise}
    \label{fig:comparison_refine}
\end{figure*}

\begin{figure*}
    \centering
    \includegraphics[width=0.8\linewidth]{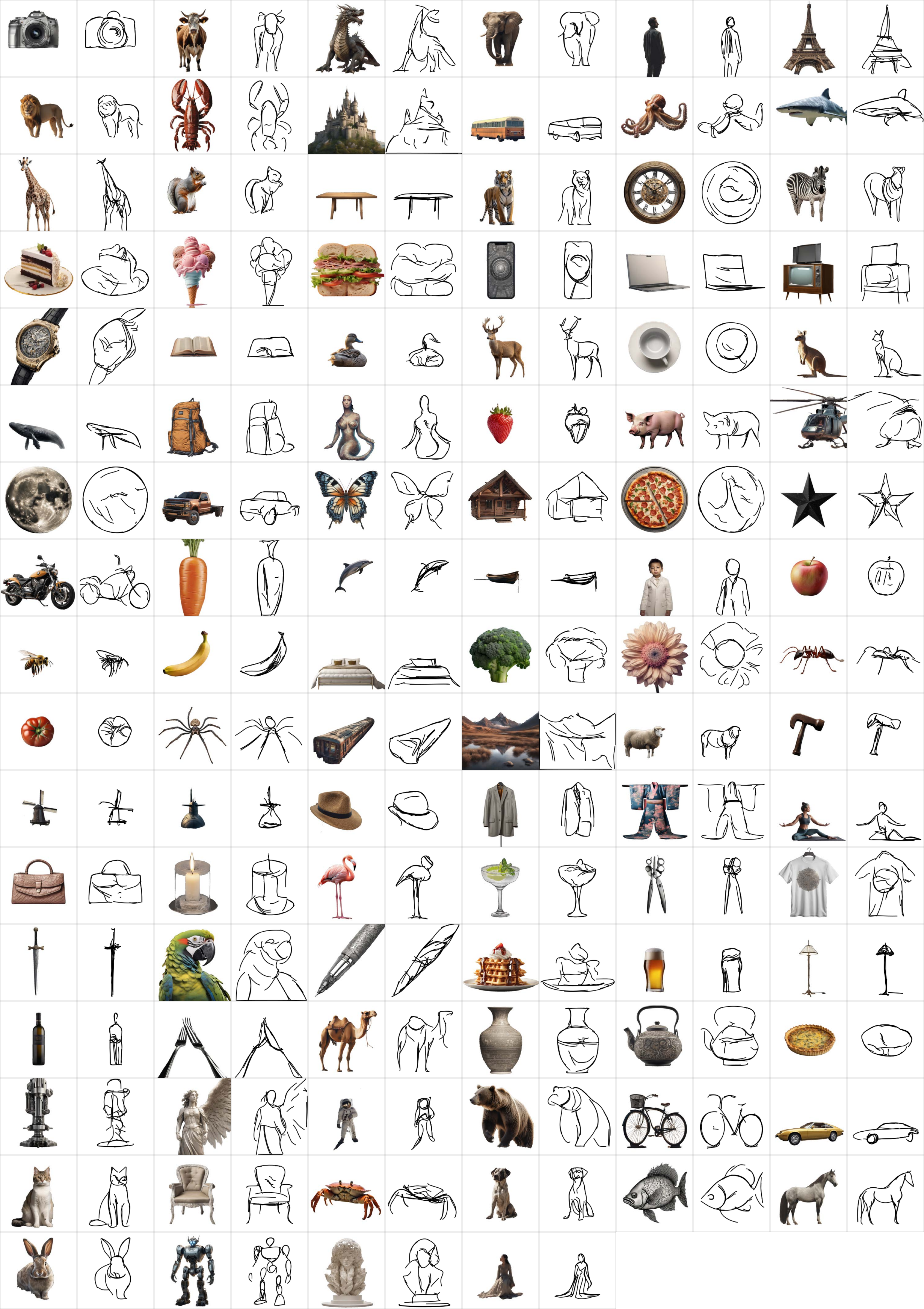}
    \caption{100 random sapmels of SwiftSketch sketches. The last three rows are seen classes, while the remaining rows are unseen classes}
    \label{fig:swiftsketch_random}
\end{figure*}

\begin{figure*}
    \centering
    \includegraphics[width=1\linewidth]{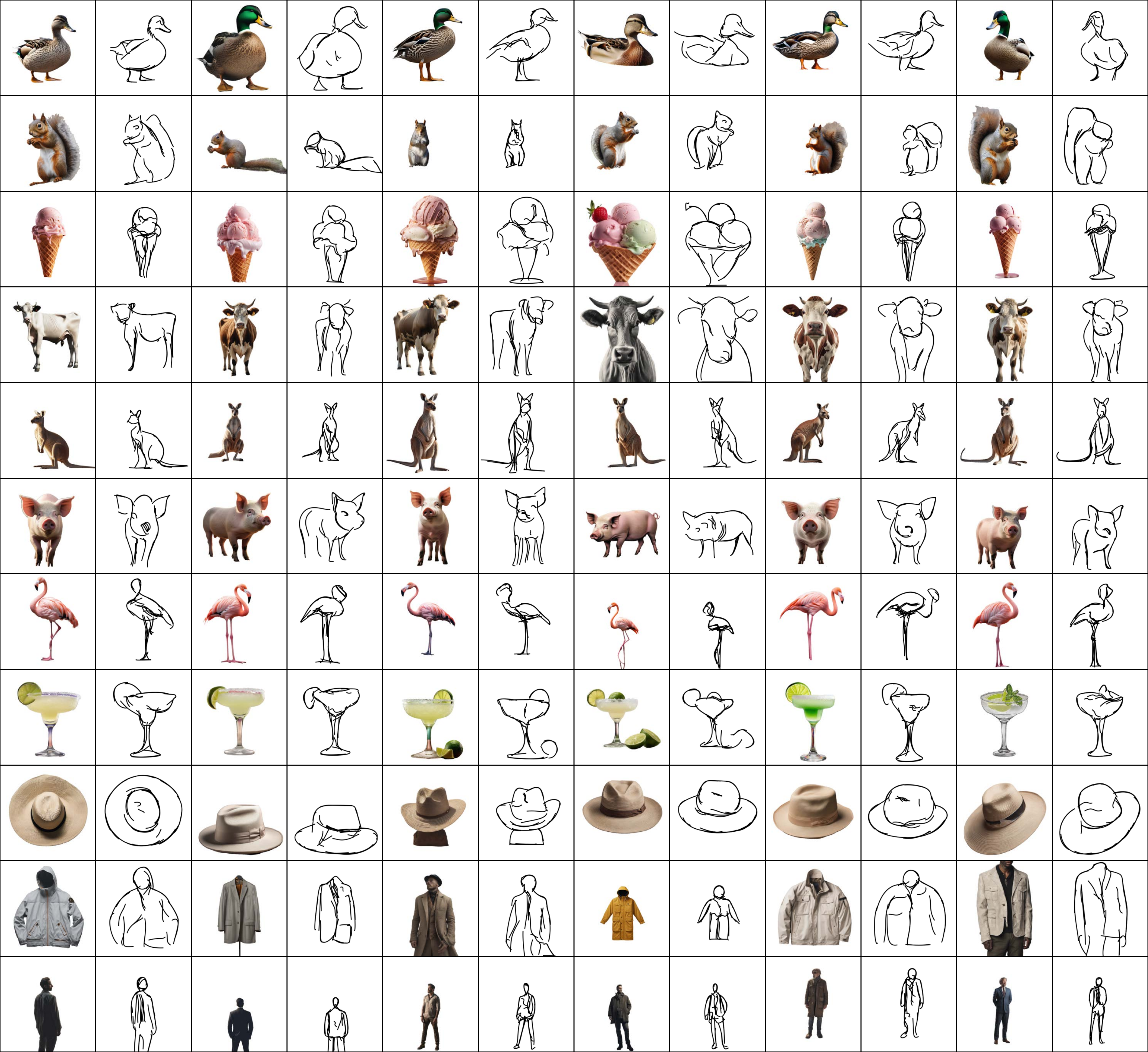}
    \caption{ Sketches generated by SwiftSketch for unseen categories.}
    \label{fig:swiftskwtch_unseen1}
\end{figure*}

\begin{figure*}
    \centering
    \includegraphics[width=1\linewidth]{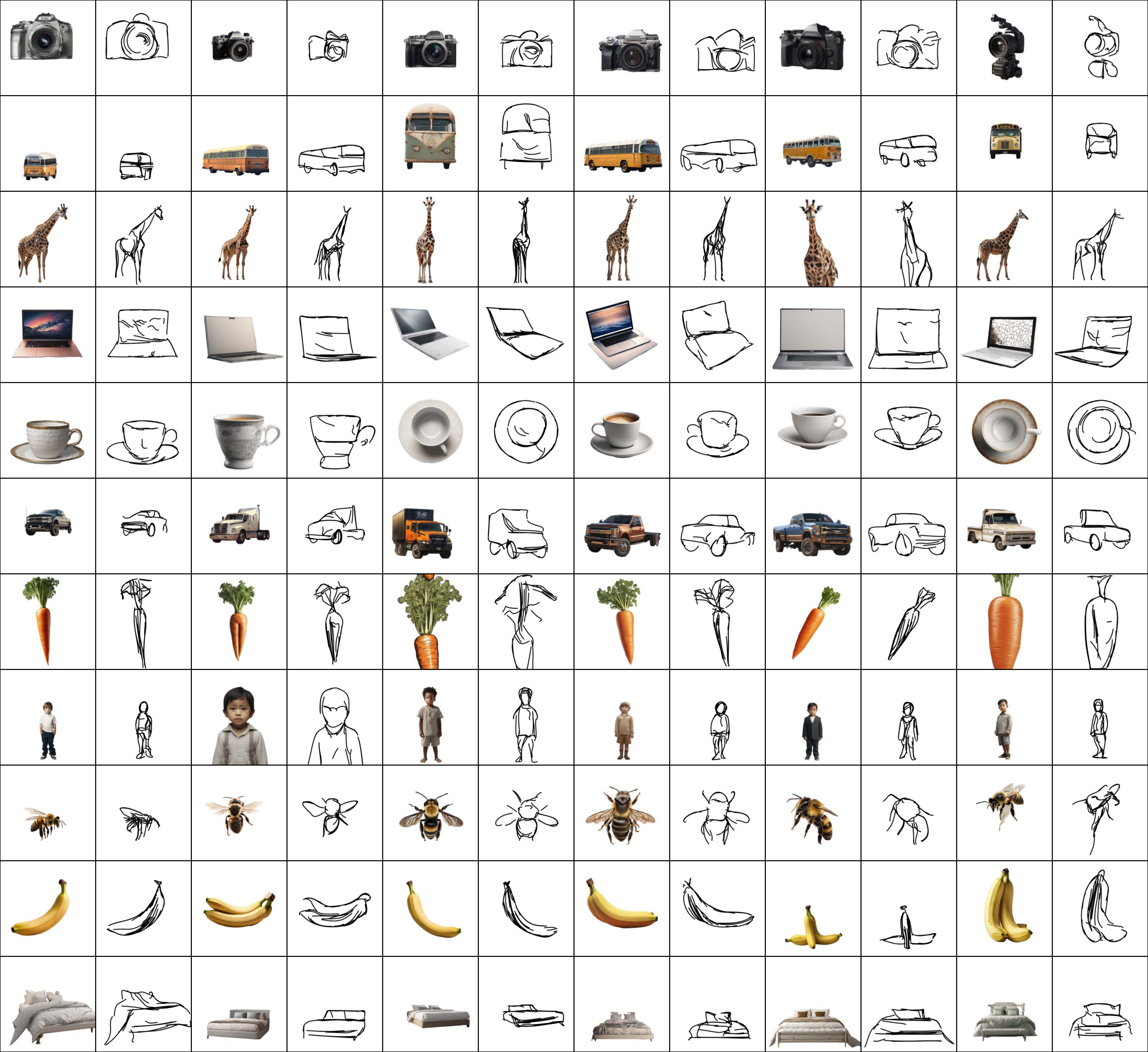}
    \caption{ Sketches generated by SwiftSketch for unseen categories.}
    \label{fig:swiftskwtch_unseen2}
\end{figure*}

\begin{figure*}
    \centering
    \includegraphics[width=1\linewidth]{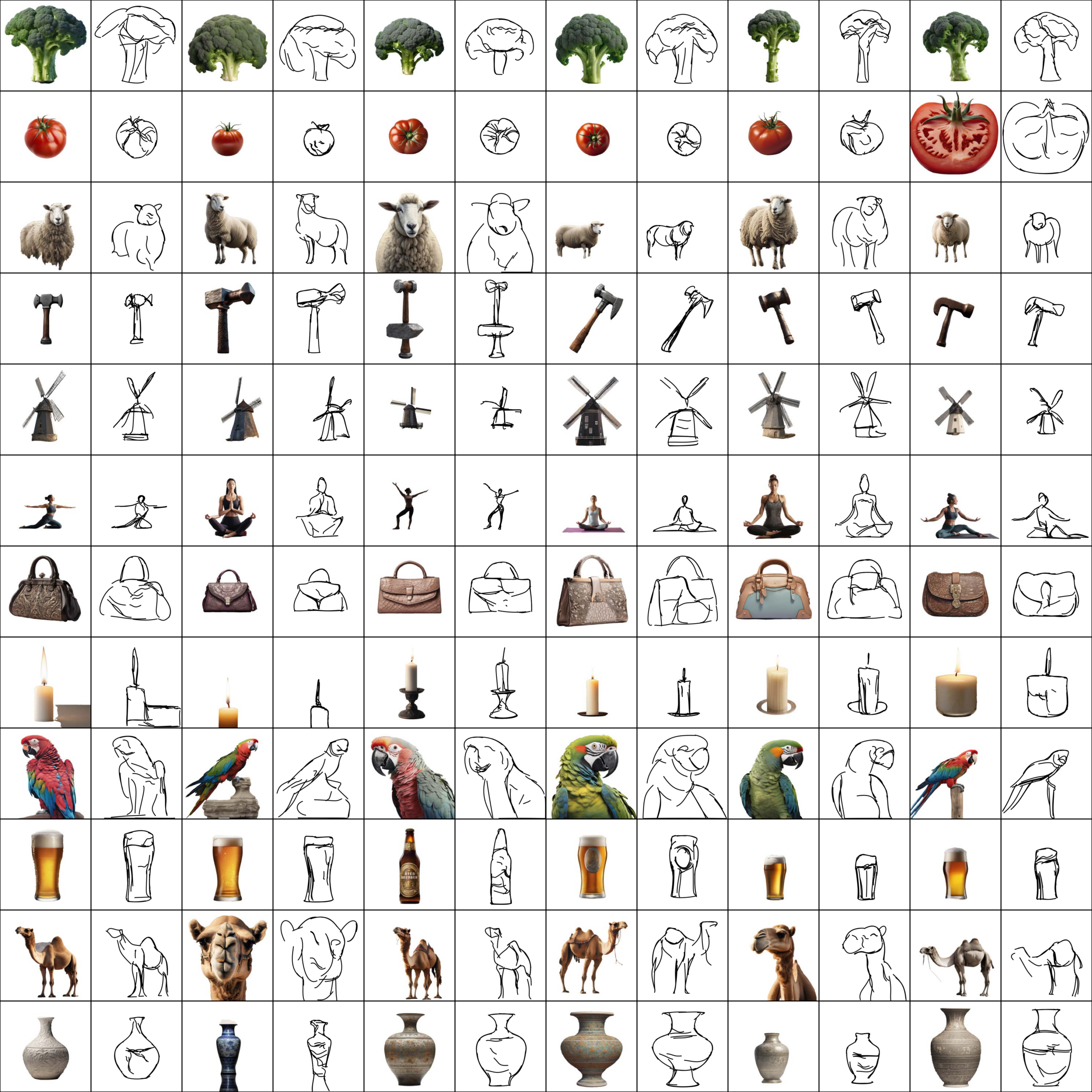}
    \caption{ Sketches generated by SwiftSketch for unseen categories.}
    \label{fig:swiftskwtch_unseen3}
\end{figure*}

\begin{figure*}
    \centering
    \includegraphics[trim=0cm 0.2cm 0 0cm,clip,width=0.8\linewidth]{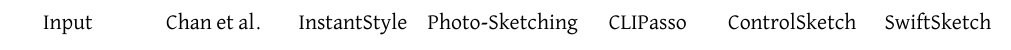}
    \includegraphics[trim=0cm 0cm 0 3.2cm,clip,width=0.8\linewidth]{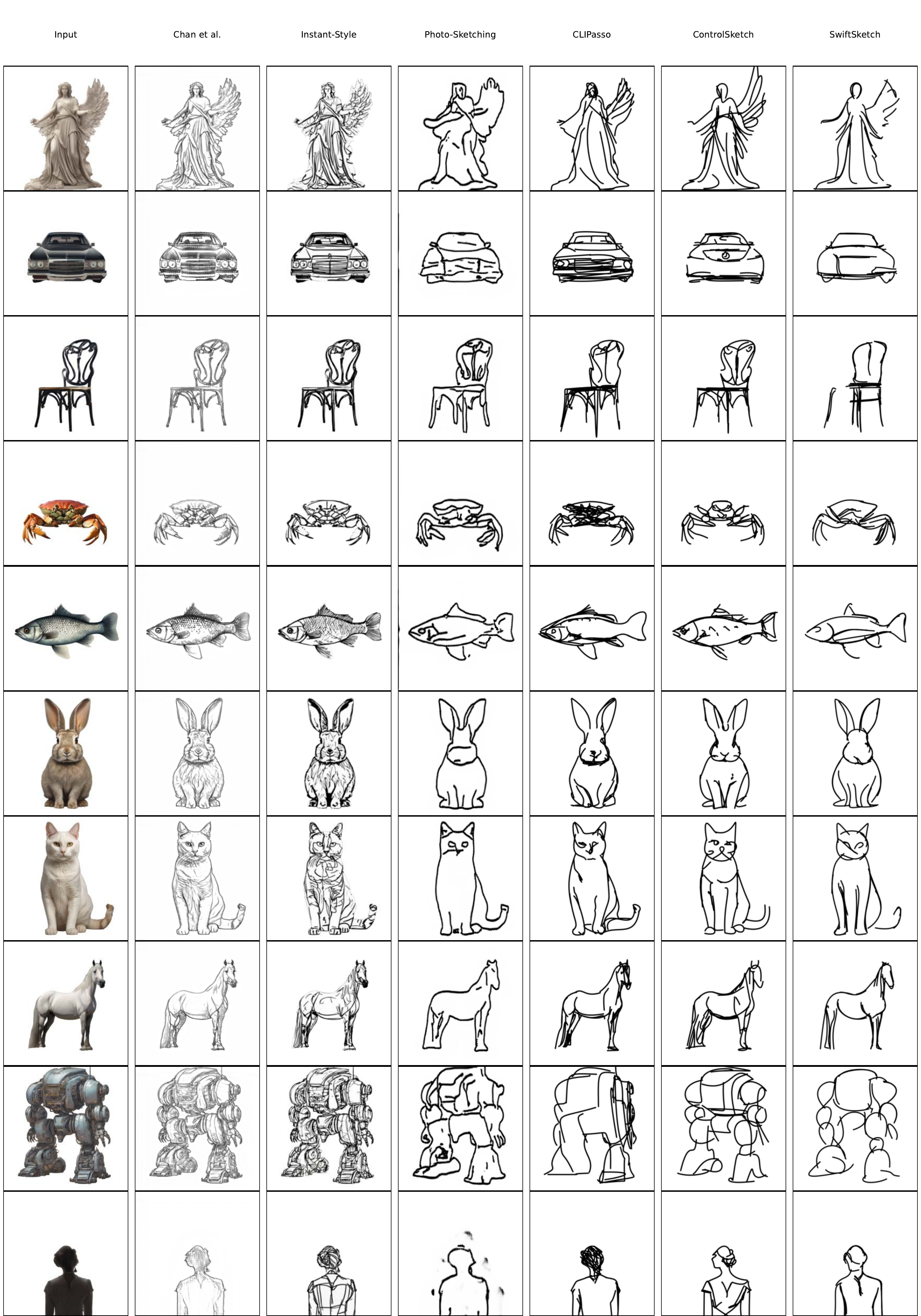}
    \caption{Qualitative comparison, seen categories}
    \label{fig:comparison_train}
\end{figure*}

\begin{figure*}
    \centering
    \includegraphics[trim=0cm 0.2cm 0 0cm,clip,width=0.8\linewidth]{figs_sup/comparison_images/titles.pdf}
    \includegraphics[trim=0cm 0cm 0 3.2cm,clip,width=0.8\linewidth]{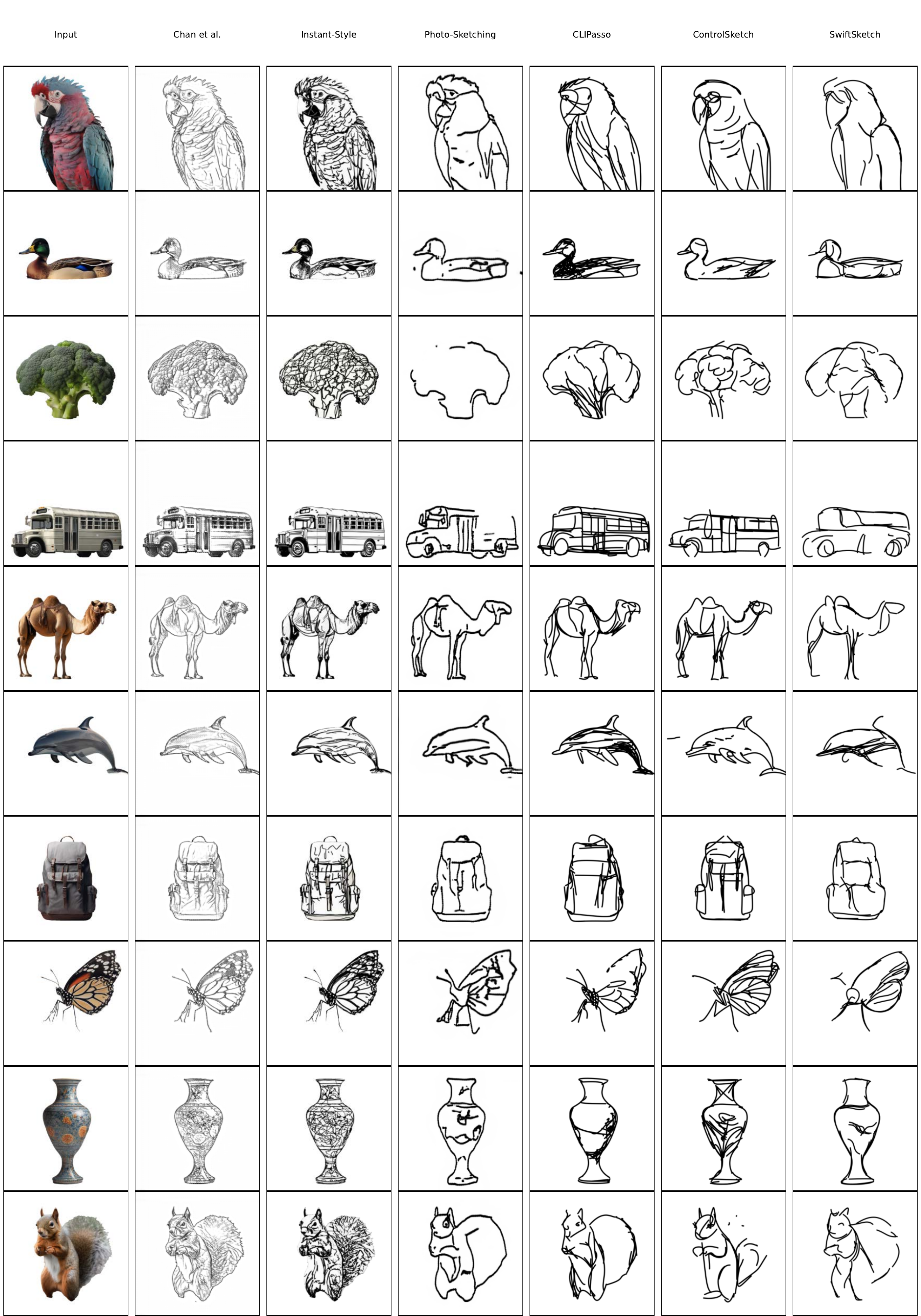}
    \caption{Qualitative comparison, unseen categories}
    \label{fig:comparison_test}
\end{figure*}

\begin{figure*}
    \centering
    \includegraphics[width=1\linewidth]{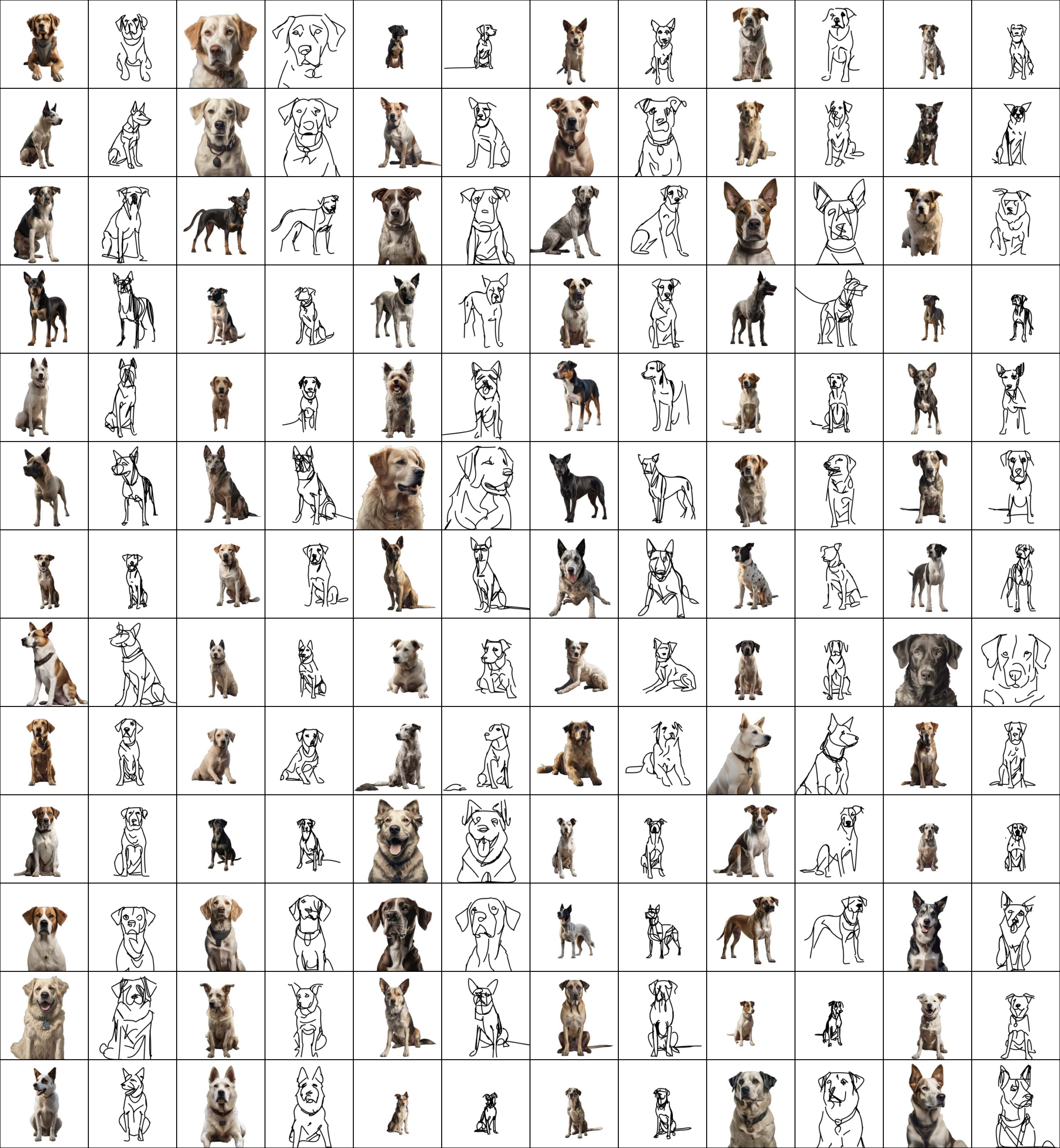}
    \caption{Dog - SwiftSketch training data examples }
    \label{fig:dog}
\end{figure*}

\begin{figure*}
    \centering
    \includegraphics[width=1\linewidth]{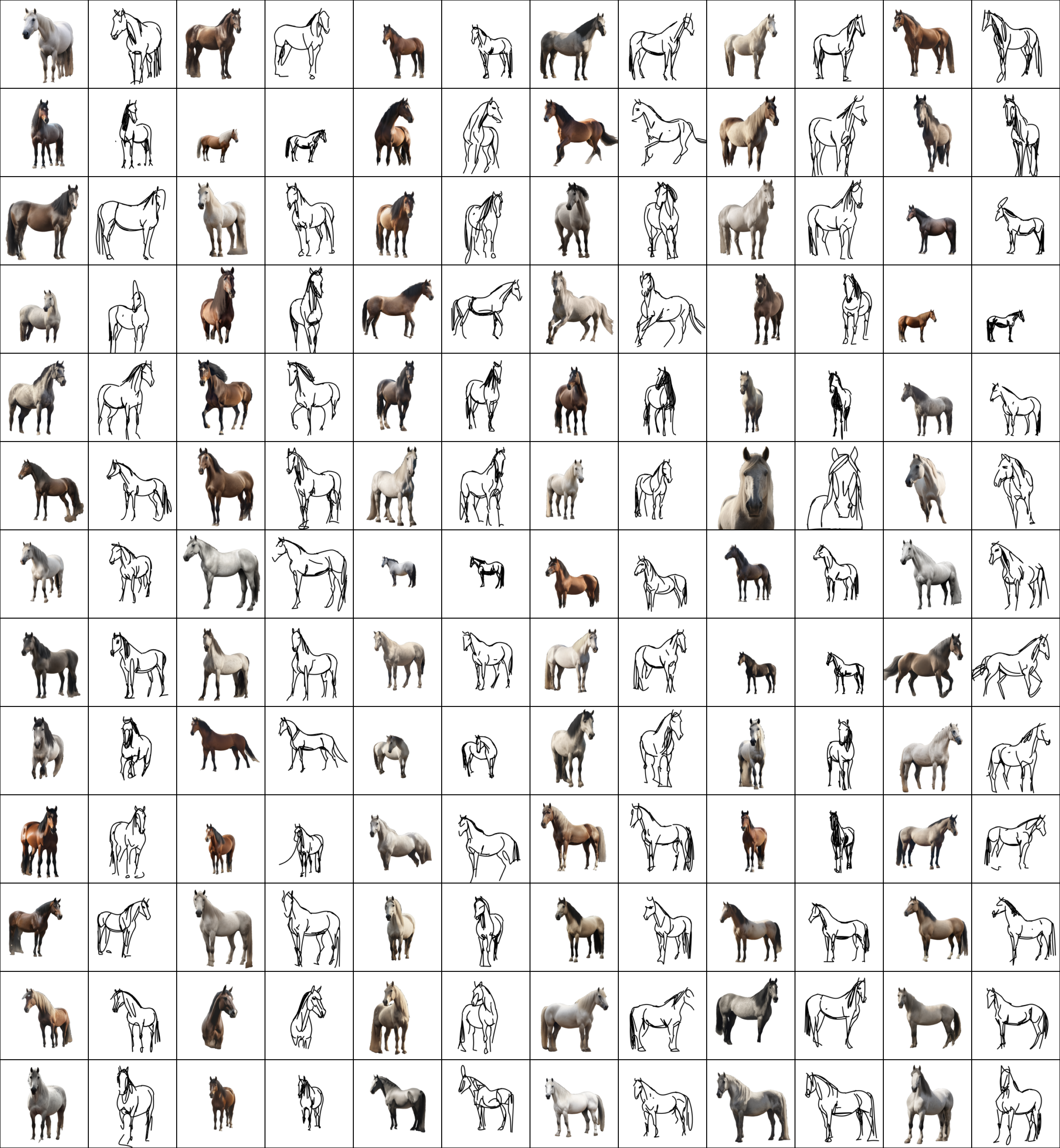}
    \caption{Horse - SwiftSketch training data examples }
    \label{fig:horse}
\end{figure*}

\begin{figure*}
    \centering
    \includegraphics[width=1\linewidth]{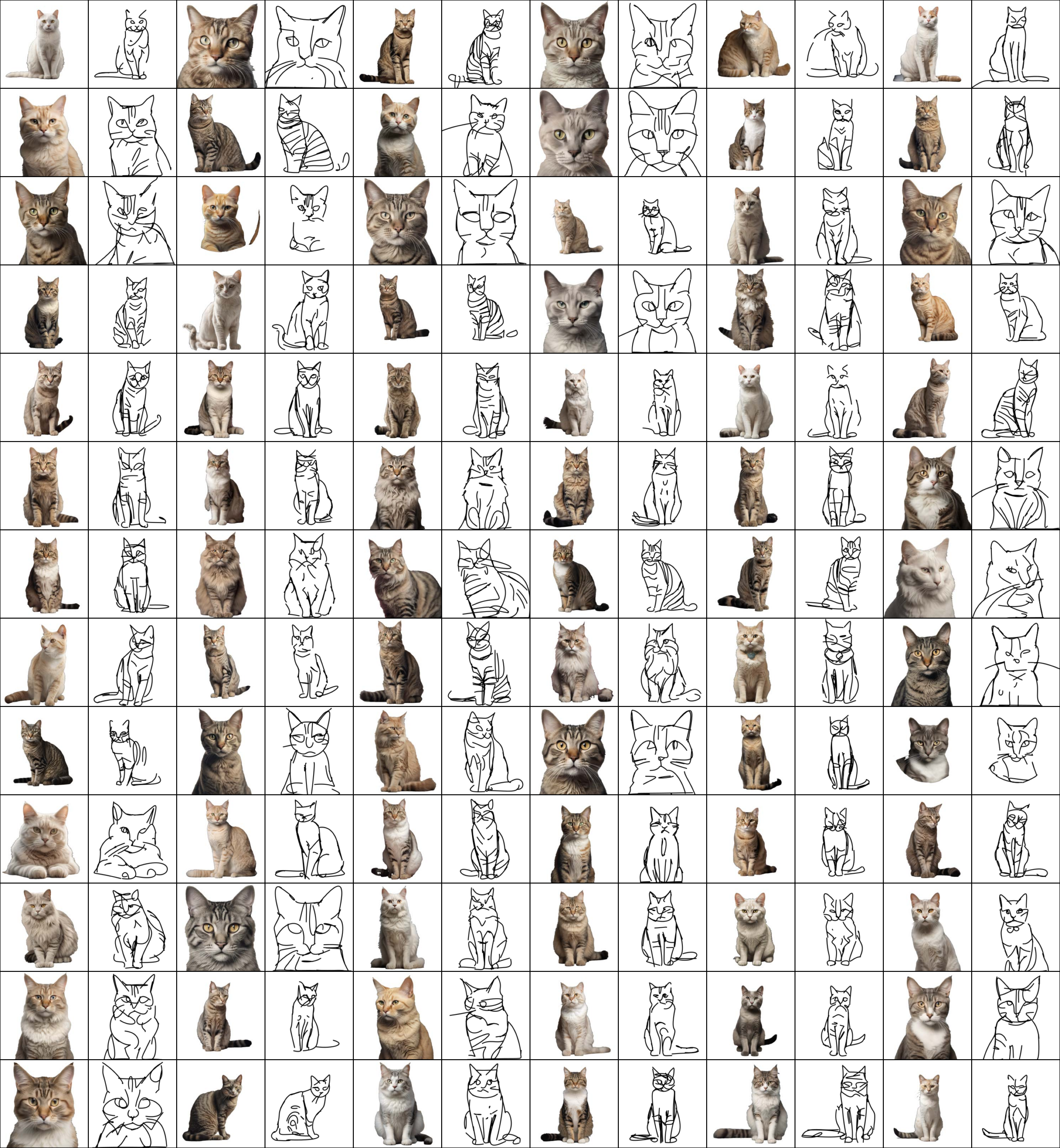}
    \caption{Cat - SwiftSketch training data examples}
    \label{fig:cat}
\end{figure*}

\begin{figure*}
    \centering
    \includegraphics[width=1\linewidth]{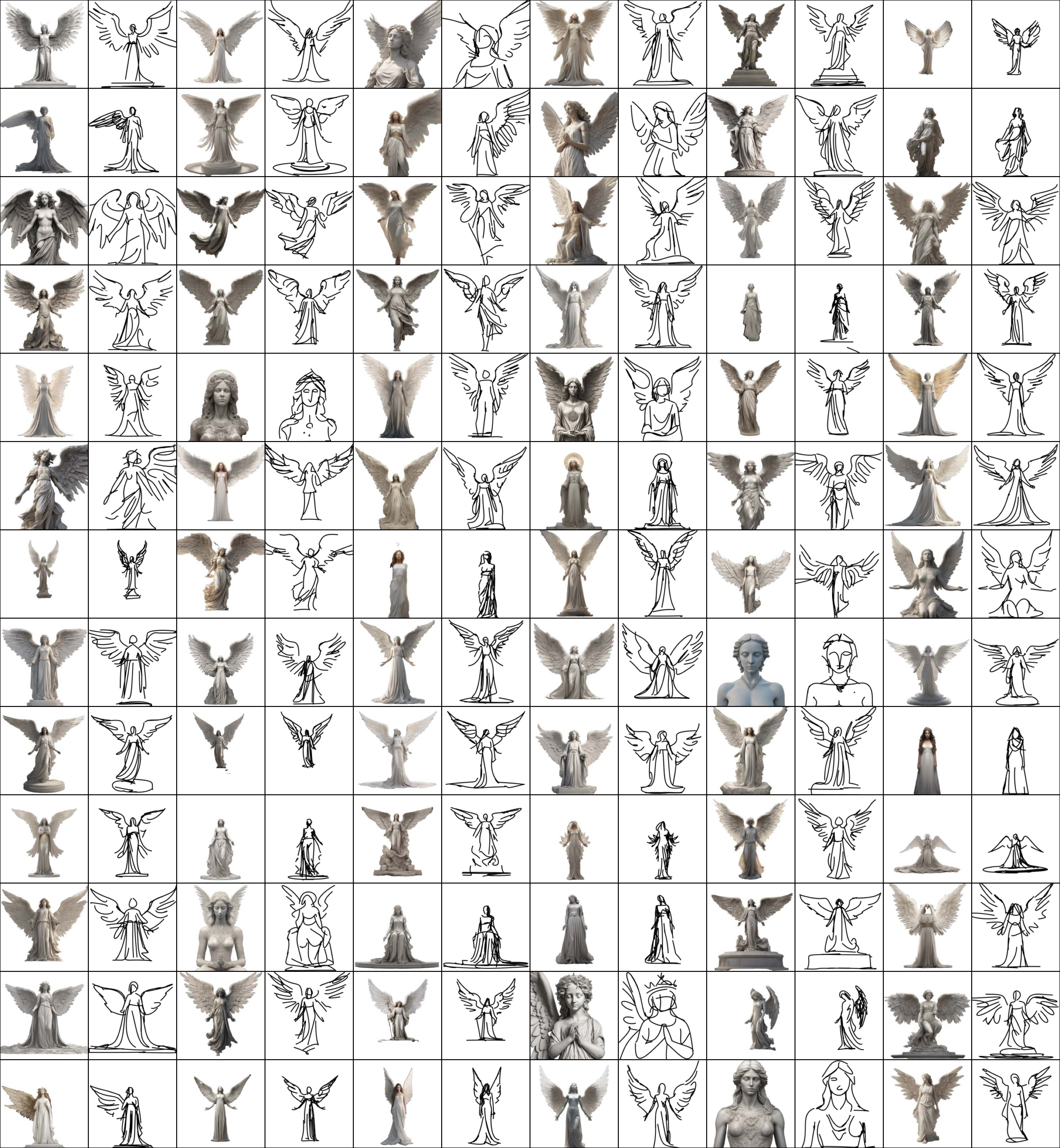}
    \caption{Angel - SwiftSketch training data examples}
    \label{fig:angle}
\end{figure*}

\begin{figure*}
    \centering
    \includegraphics[width=1\linewidth]{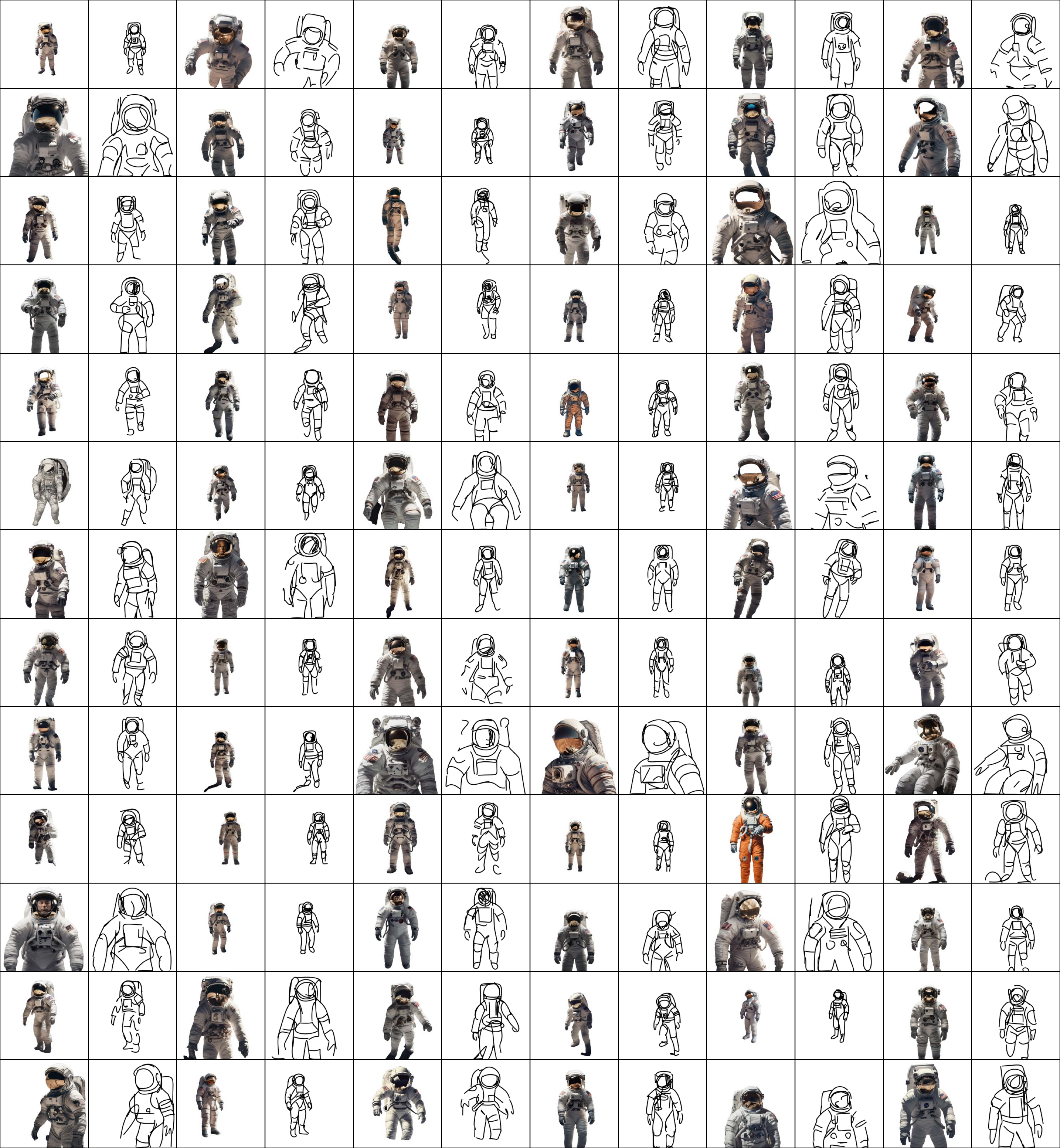}
    \caption{Astronaut - SwiftSketch training data examples}
    \label{fig:astronaut}
\end{figure*}

\begin{figure*}
    \centering
    \includegraphics[width=1\linewidth]{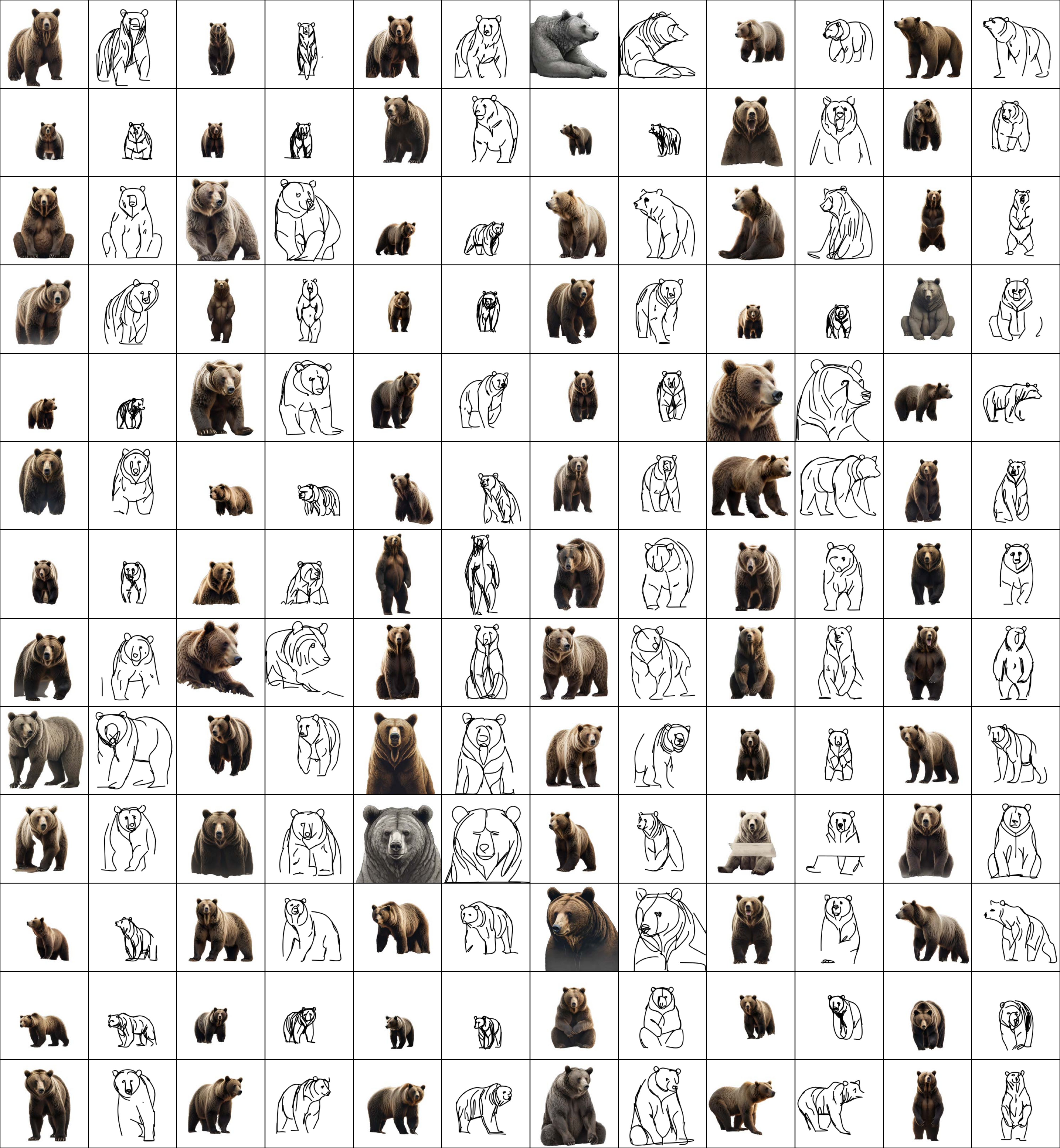}
    \caption{Bear - SwiftSketch training data examples}
    \label{fig:bear}
\end{figure*}

\begin{figure*}
    \centering
    \includegraphics[width=1\linewidth]{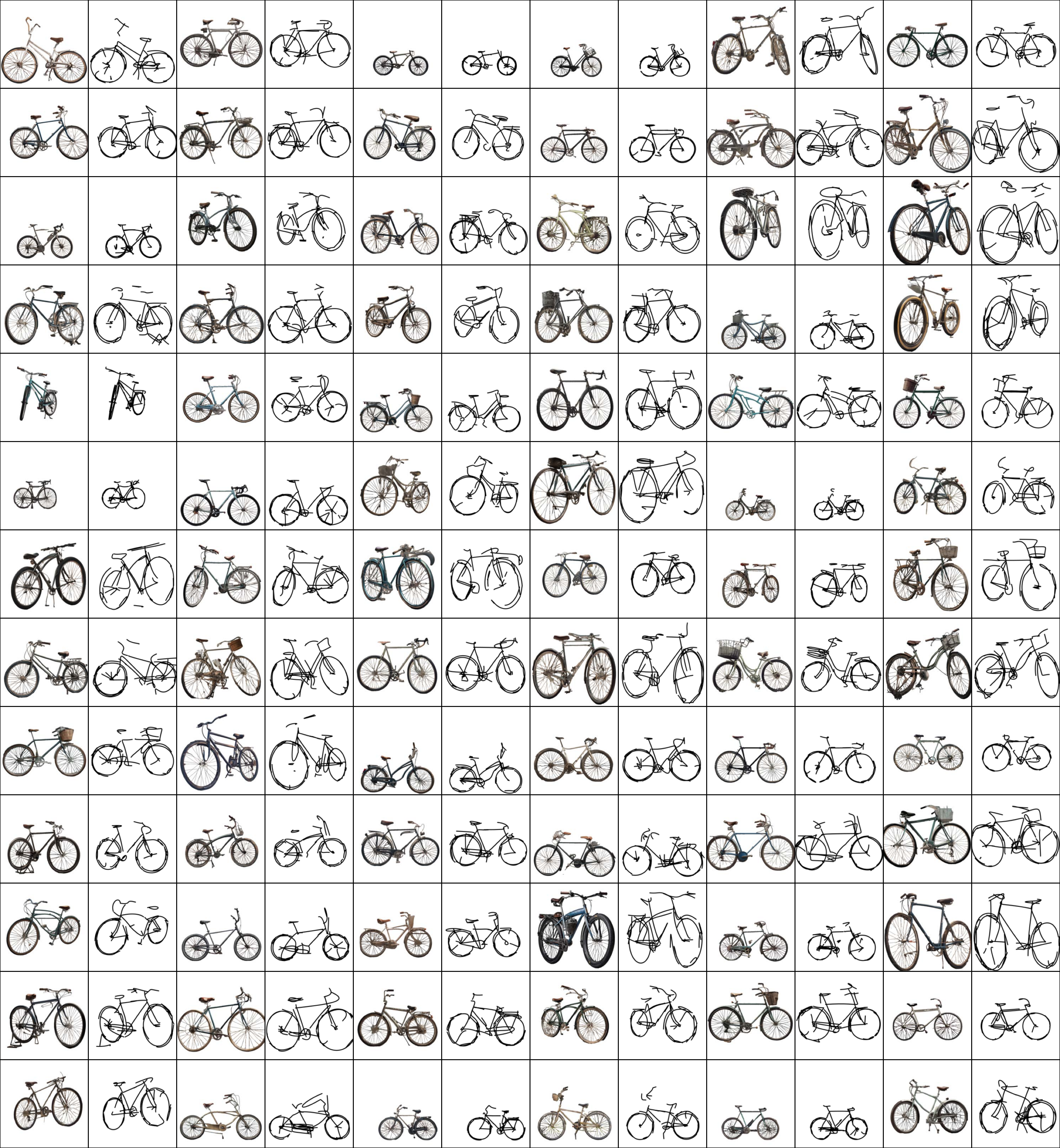}
    \caption{Bicycle - SwiftSketch training data examples}
    \label{fig:bicycle}
\end{figure*}

\begin{figure*}
    \centering
    \includegraphics[width=1\linewidth]{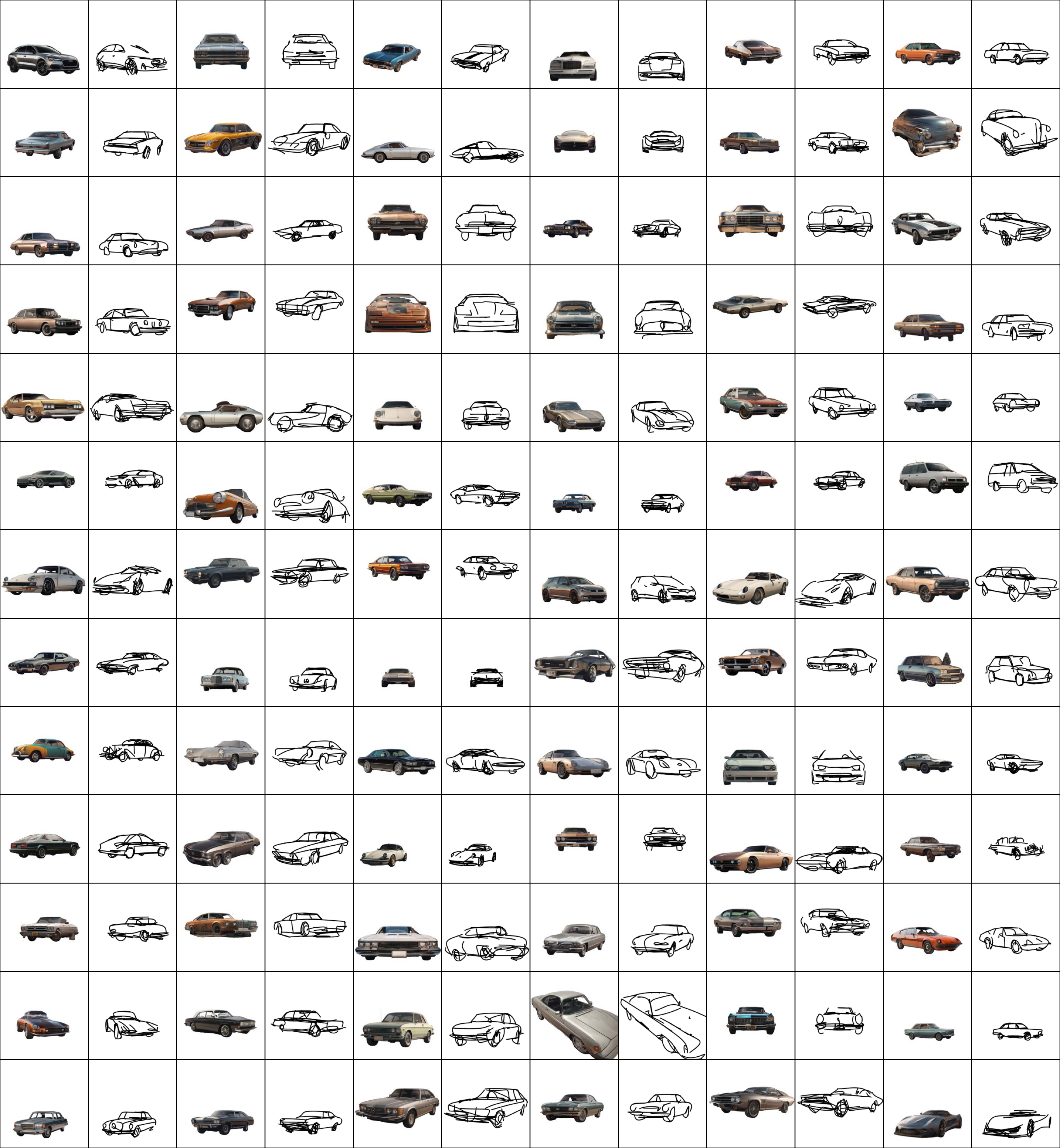}
    \caption{Car - SwiftSketch training data examples}
    \label{fig:car}
\end{figure*}

\begin{figure*}
    \centering
    \includegraphics[width=1\linewidth]{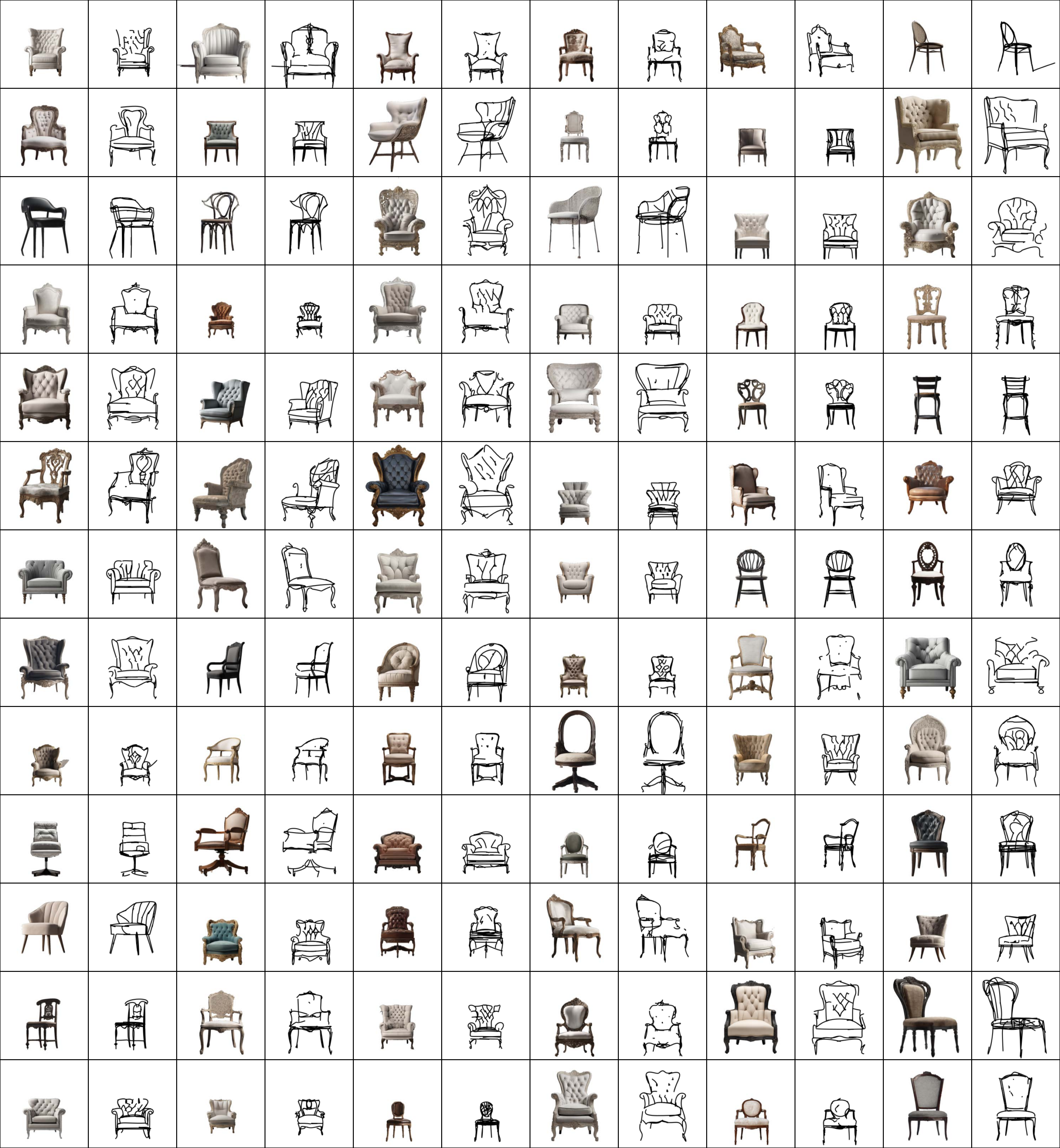}
    \caption{Chair - SwiftSketch training data examples }
    \label{fig:chair}
\end{figure*}

\begin{figure*}
    \centering
    \includegraphics[width=1\linewidth]{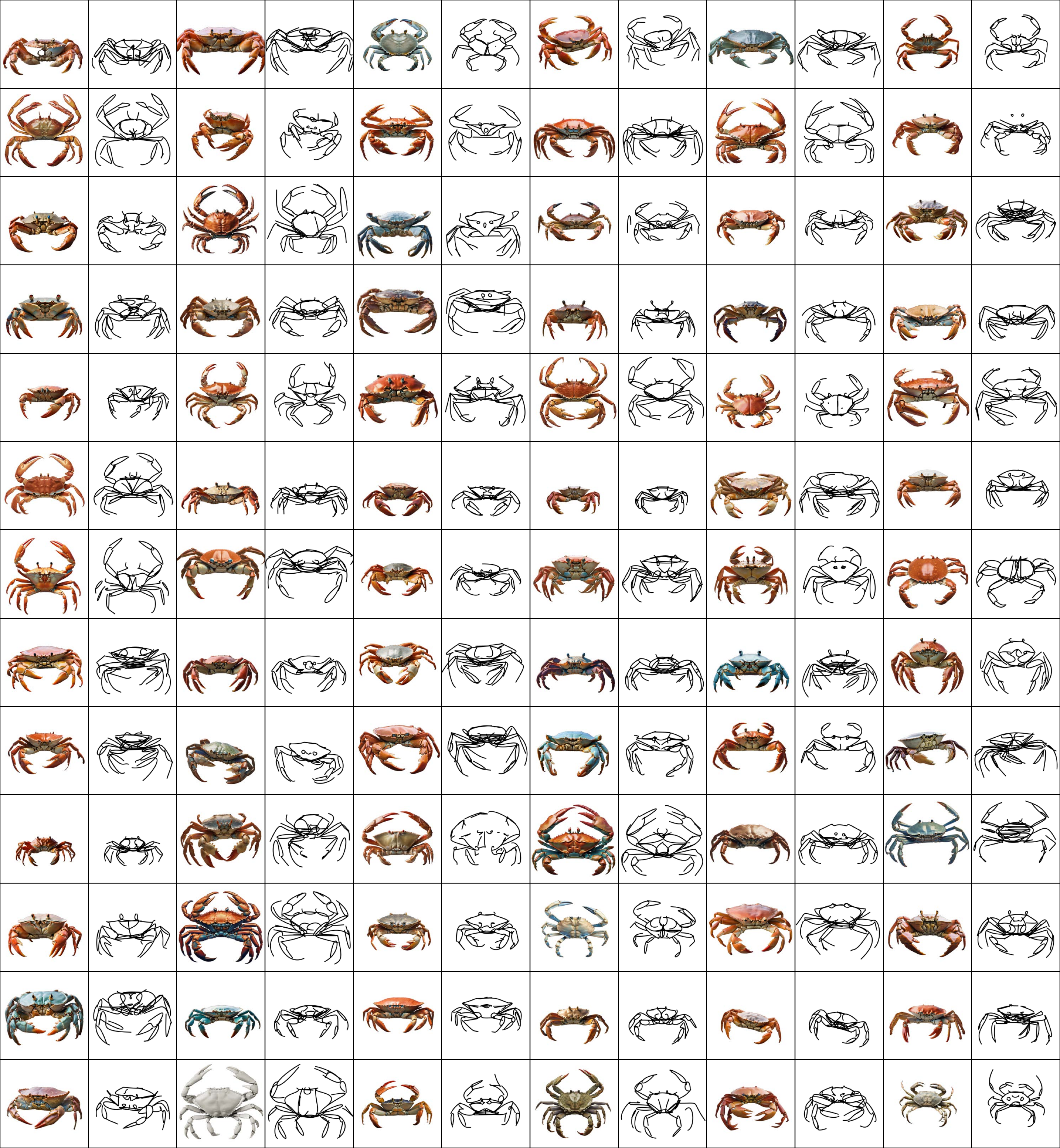}
    \caption{Crab - SwiftSketch training data examples}
    \label{fig:crab}
\end{figure*}

\begin{figure*}
    \centering
    \includegraphics[width=1\linewidth]{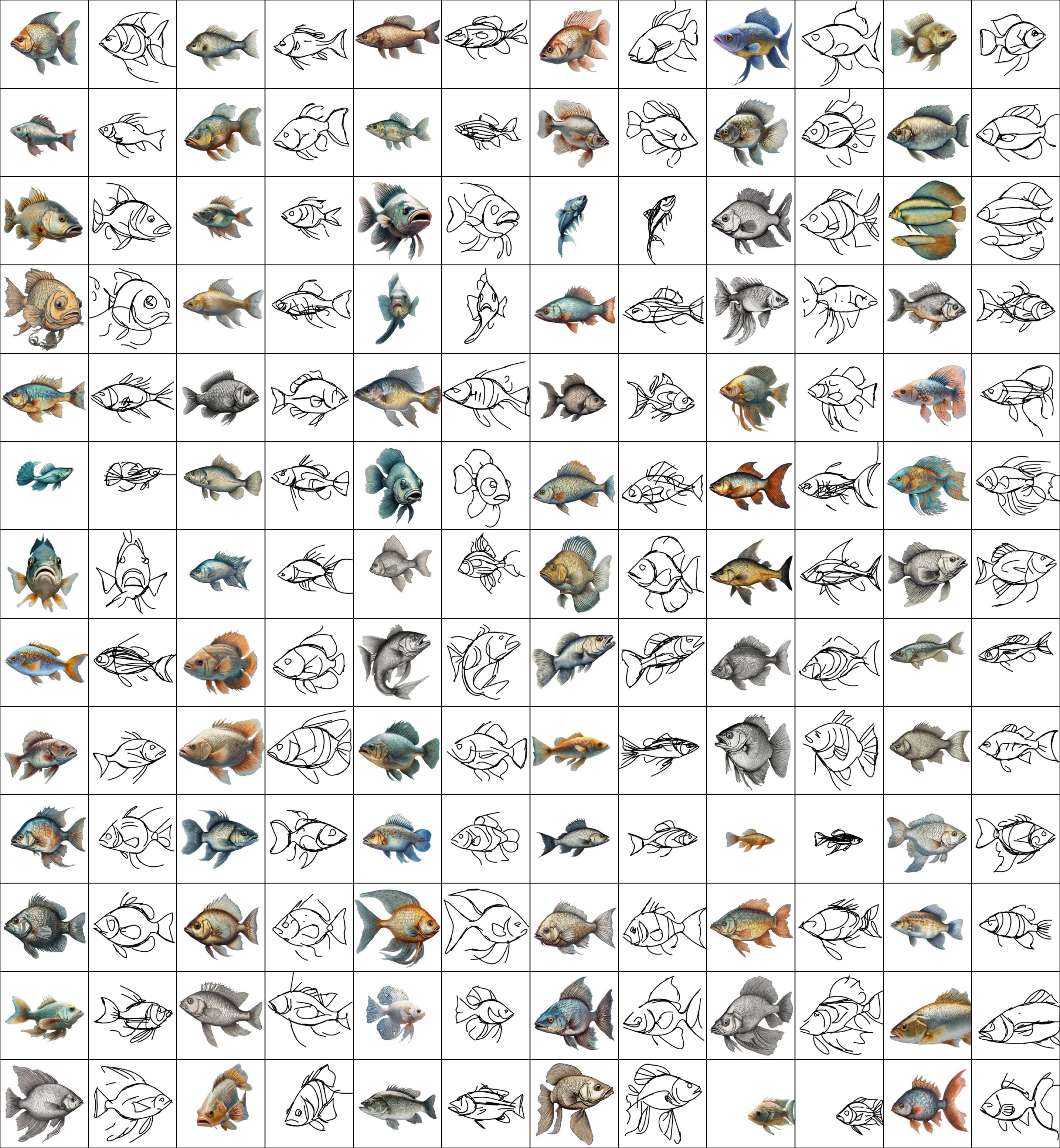}
    \caption{Fish - SwiftSketch training data examples}
    \label{fig:fish}
\end{figure*}

\begin{figure*}
    \centering
    \includegraphics[width=1\linewidth]{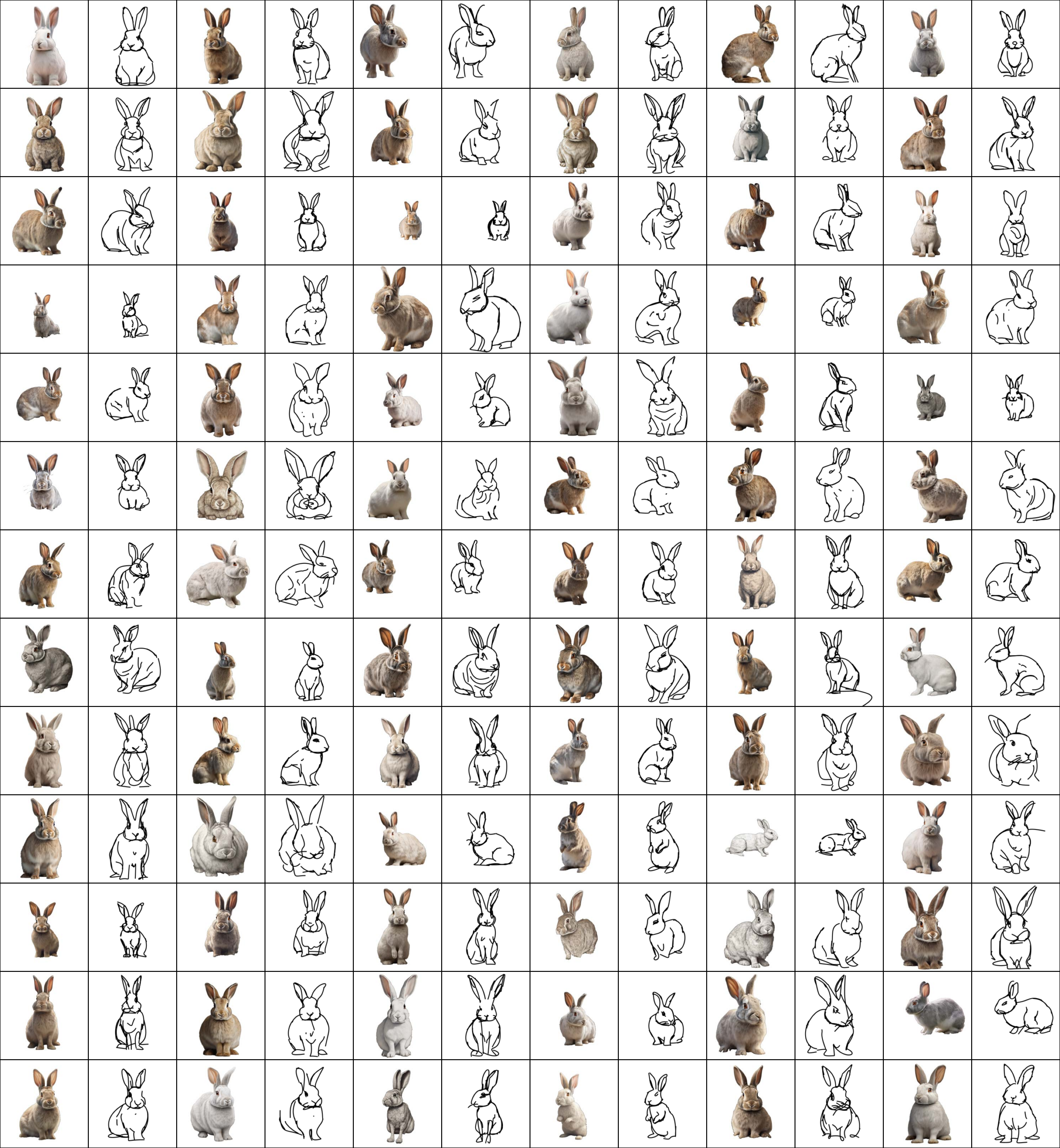}
    \caption{Rabbit - SwiftSketch training data examples}
    \label{fig:rabbit}
\end{figure*}

\begin{figure*}
    \centering
    \includegraphics[width=1\linewidth]{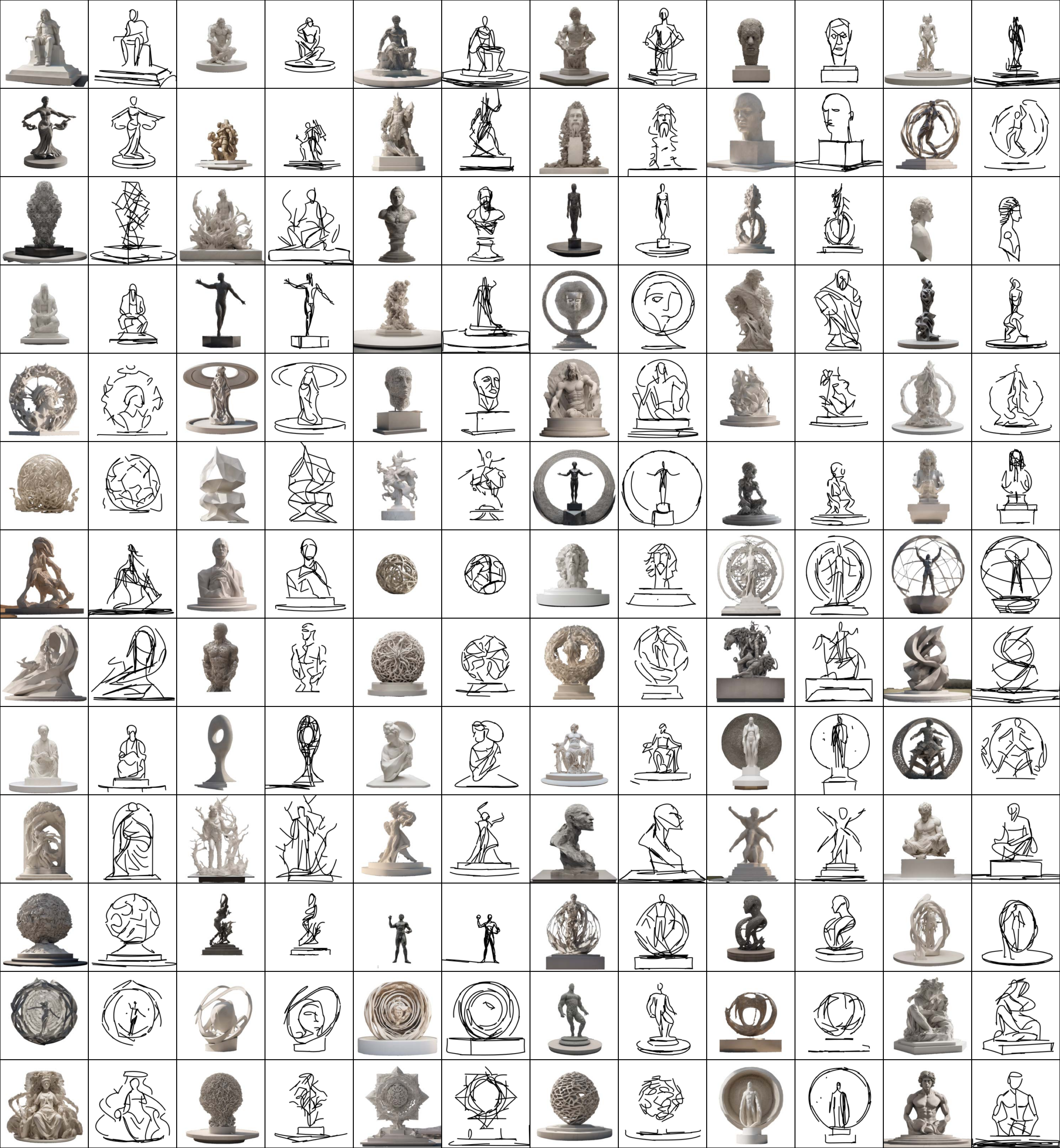}
    \caption{:Sculpture - SwiftSketch training data examples }
    \label{fig:Sculpture}
\end{figure*}

\begin{figure*}
    \centering
    \includegraphics[width=1\linewidth]{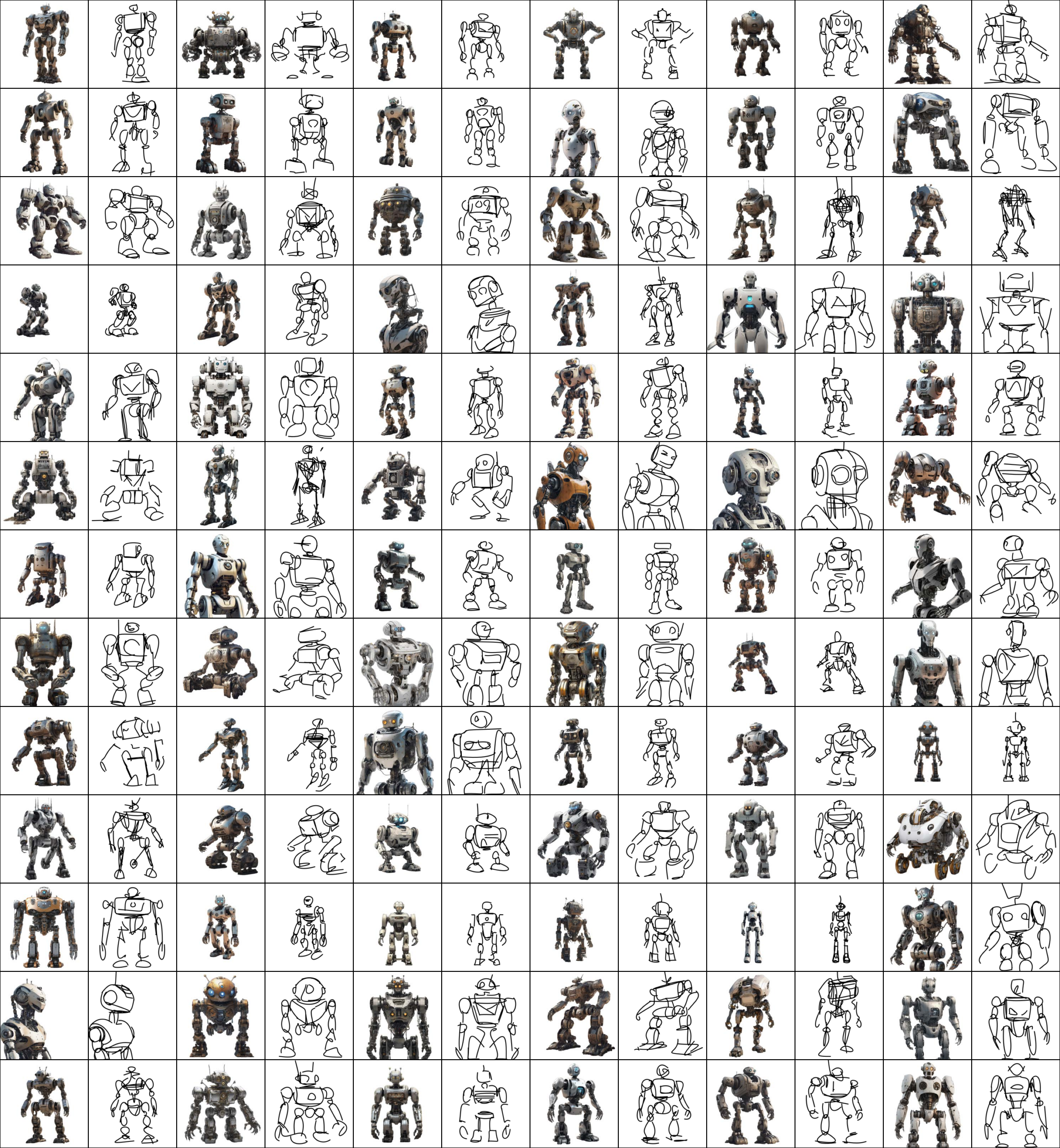}
    \caption{Robot - SwiftSketch training data examples }
    \label{fig:robot}
\end{figure*}

\begin{figure*}
    \centering
    \includegraphics[width=1\linewidth]{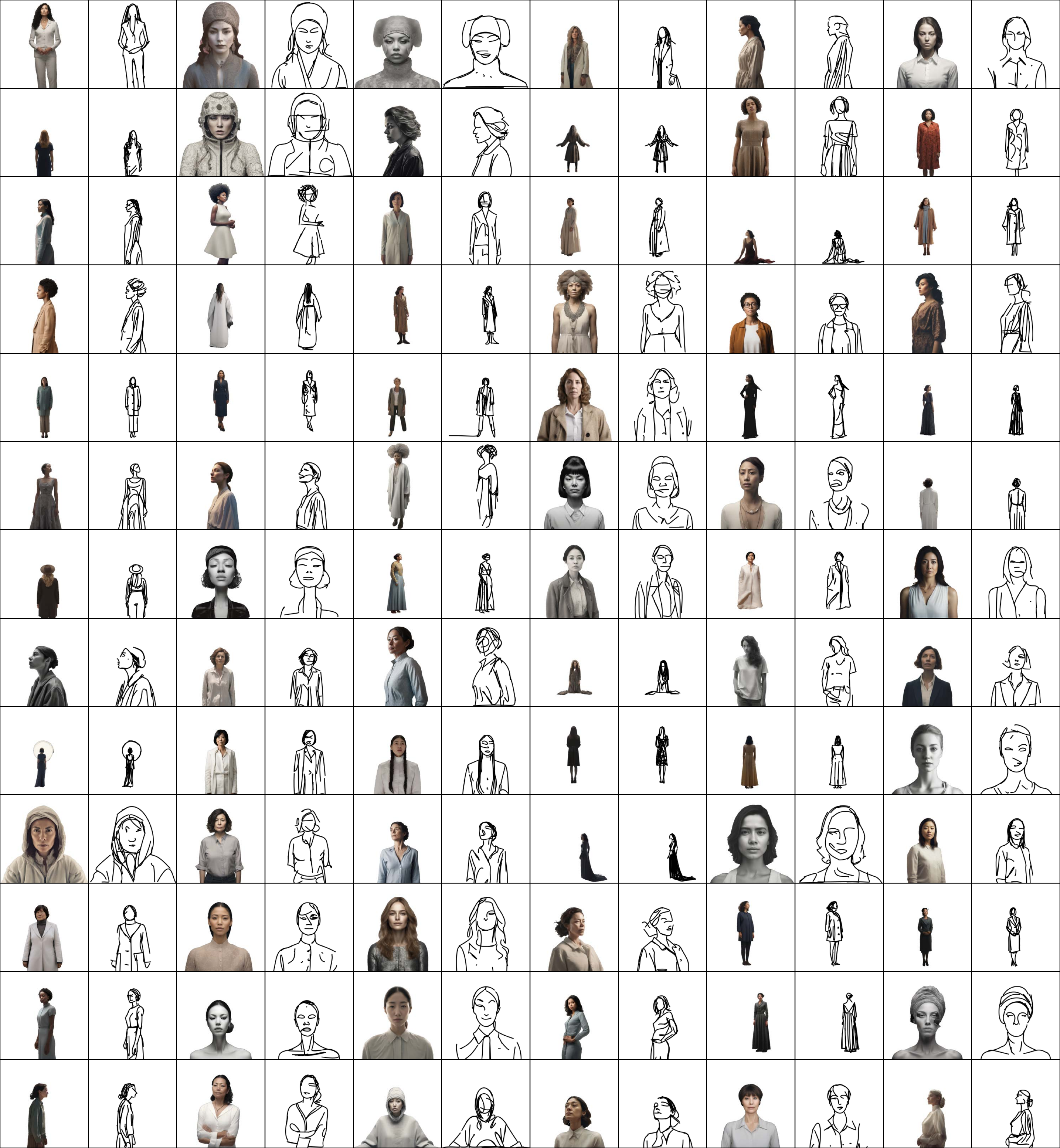}
    \caption{Woman - SwiftSketch training data examples }
    \label{fig:woman}
\end{figure*}

\end{document}